\documentclass[10pt,twocolumn,letterpaper]{article}

\usepackage[pagenumbers]{cvpr}
\usepackage{minitoc}

\usepackage[dvipsnames]{xcolor}

\usepackage{multirow,bigdelim}

\definecolor{cvprblue}{rgb}{0.21,0.49,0.74}
\usepackage[pagebackref,breaklinks,colorlinks,citecolor=cvprblue]{hyperref}

\newcommand{\ap}[1]{``#1''}

\def\eg{\emph{e.g}\onedot}

\def\ie{\emph{i.e}\onedot}

\def\etal{\emph{et al}\onedot}

\def\gtransform{\mathcal{T}}

\def\Bezier{B\'{e}zier\xspace}

\def\confName{CVPR}
\def\confYear{2024}

\title{\vspace{-0.6cm}Breathing Life Into Sketches Using Text-to-Video Priors\vspace{-0.3cm}}

\author{Rinon Gal$^{*,1,2}$ \hspace{0.05\linewidth} \and
Yael Vinker$^{*,1}$ \hspace{0.05\linewidth} \and
Yuval Alaluf$^{2}$ \hspace{0.05\linewidth} \and 
Amit Bermano$^{1}$ \and
\hspace{0.06\linewidth} Daniel Cohen-Or$^{1}$ \hspace{0.04\linewidth} \and
Ariel Shamir$^{3}$ \and
\hspace{0.04\linewidth} Gal Chechik$^{2}$ \hspace{0.06\linewidth}  \and \hspace{0.9\linewidth}
\and \hspace{0.1\linewidth}$^{1}$Tel-Aviv University \and $^{2}$NVIDIA \and $^{3}$Reichman University \and \hspace{0.05\linewidth} 
}

\begin{document}

\doparttoc %
\faketableofcontents %

\twocolumn[{%
\vspace{-1em}
\maketitle
\renewcommand\twocolumn[1][]{#1}%
\vspace{-2em}
\vspace{-0.15in}
\begin{center}

     \setlength{\abovecaptionskip}{0.15cm}
     \setlength{\belowcaptionskip}{0pt}
    \centering
    \includegraphics[width=0.97\linewidth]{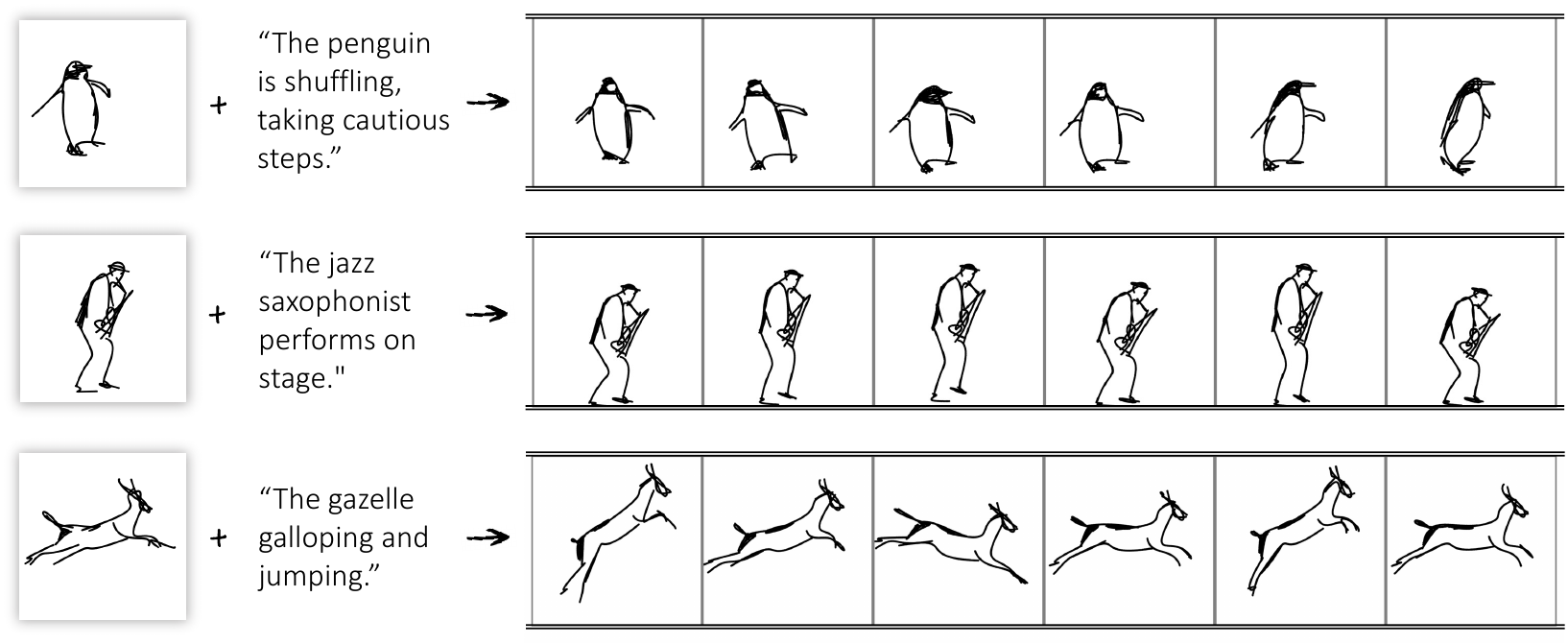}
    \captionsetup{type=figure}\caption{Given a still sketch in vector format and a text prompt describing a desired action, our method automatically animates the drawing with respect to the prompt. Please see the full animations in our project page: 
\url{https://livesketch.github.io/}}    
    \label{fig:teaser}
\end{center}
}]

\def\thefootnote{*}\footnotetext{Indicates Equal Contribution. Order determined by coin flip.}

\begin{abstract}
\vspace{-10pt}
A sketch is one of the most intuitive and versatile tools humans use to convey their ideas visually. An animated sketch opens another dimension to the expression of ideas and is widely used by designers for a variety of purposes.
Animating sketches is a laborious process, requiring extensive experience and professional design skills.
In this work, we present a method that automatically adds motion to a single-subject sketch (hence, ``breathing life into it''), merely by providing a text prompt indicating the desired motion.
The output is a short animation provided in vector representation, which can be easily edited.
Our method does not require extensive training, but instead leverages the motion prior of a large pretrained text-to-video diffusion model using a score-distillation loss to guide the placement of strokes. 
To promote natural and smooth motion and to better preserve the sketch's appearance, we model the learned motion through two components. The first governs small local deformations and the second controls global affine transformations.
Surprisingly, we find that even models that struggle to generate sketch videos on their own can still serve as a useful backbone for animating abstract representations. 
\vspace{-5pt}
\end{abstract}

\section{Introduction}
\label{sec:intro}
Sketches serve as a fundamental and intuitive tool for visual expression and communication \cite{Fan2023DrawingAA,aubert2014pleistocene,gombrich1995story}.
Sketches capture the essence of visual entities with a few strokes, allowing humans to communicate abstract visual ideas.
In this paper, we propose a method to \ap{breathe life} into a static sketch by generating semantically meaningful short videos from it.
Such animations can be useful for storytelling, illustrations, websites, presentations, and just for fun. 

Animating sketches using conventional tools (such as Adobe Animate and Toon Boom) is challenging even for experienced designers~\cite{su2018live}, requiring specific artistic expertise.
Hence, long-standing research efforts in computer graphics sought to develop automatic tools to simplify this process. 
However, these tools face multiple hurdles, such as a need to identify the semantic component of the sketch, or learning to create motion that appears natural. As such, existing methods commonly rely on user-annotated skeletal key points \cite{smith2023method,dvorovznak2018toonsynth} or user-provided reference motions that align with the sketch semantics \cite{su2018live,bregler2002turning,weng2019photo}.

In this work, we propose to bring a given static sketch to life, based on a textual prompt, without the need for any human annotations or explicit reference motions. We do so by leveraging a pretrained text-to-video diffusion model~\cite{khachatryan2023text2video}.
Several recent works propose using the prior of such models to bring life to a static \textit{image}~\cite{videocomposer2023,xing2023dynamicrafter,ni2023conditional}. 
However, sketches pose distinct challenges, which existing methods fail to tackle as they are not designed with this domain in mind. 
Our method takes the recent advancement in text-to-video models into this new realm, aiming to tackle the challenging task of sketch animation. For this purpose, we propose specific design choices considering the delicate characteristics of this abstract domain.

In line with prior sketch generation approaches \cite{vinker2022clipasso, clipascene}, we use a vector representation of sketches, defining a sketch as a set of strokes (cubic \Bezier curves) parameterized by their control points. 
Vector representations are popular among designers as they offer several advantages compared to pixel-based images. They are resolution-independent, \ie can be scaled without losing quality. 
Moreover, they are easily editable: one can modify the sketch's appearance by choosing different stroke styles or change its shape by dragging control points. Additionally, their sparsity promotes smooth motion while preventing pixelization and blurring.

To bring a static sketch to life, we train a network to modify the stroke parameters for each video frame with respect to a given text prompt. Our method is optimization-based and requires no data or fine-tuning of large models.
In addition, our method is general and can easily adapt to different text-to-video models, facilitating the use of future advancements in this field.

We train the network using a score-distillation sampling (SDS) loss \cite{poole2022dreamfusion}. This loss was designed to leverage pretrained text-to-\textit{image} diffusion models for the optimization of non-pixel representations (e.g., NeRFs~\cite{mildenhall2021nerf,metzer2022latent} or SVGs~\cite{jain2022vectorfusion,IluzVinker2023}) to meet given text-based constraints. 
We use this loss to extract motion priors from pretrained text-to-\textit{video} diffusion models \cite{ho2022imagen,videocomposer2023}. Importantly, this allows us to inherit the internet-scale knowledge embedded in such models, enabling animation for a wide range of subjects across multiple categories.

We seperate the object movement into two components: local motion and global motion. Local motion aims to capture isolated, local effects (a saxophone player bending their knee). Conversely, global motion affects the object shape as a whole and is modeled through a per-frame transformation matrix. 
It can thus capture rigid motion (a penguin hobbling across the frame), or coordinate effects (the same penguin growing in size as it approaches the camera).
We find that this separation is crucial in generating motion that is both locally smooth and globally significant while remaining faithful to the original characteristics of the subject.

We animate sketches from various domains and demonstrate the effectiveness of our approach in producing smooth and natural motion that conveys the intention of the control text while better preserving the shape and appearance of the input sketch. 

We compare our results with recent pixel-based approaches highlighting the advantage of vector-based animation in the sketch domain. Our work allows anyone to breath life into their sketch in a simple and intuitive manner.

\section{Previous Work}
\label{sec:prevwork}

\paragraph{Sketches}
Free-hand sketching is a valuable tool for expressing ideas, concepts, and actions~\cite{Fan2018CommonOR, Hertzmann2020WhyDL, Fan2023DrawingAA}.
Extensive research has been conducted on the automatic generation of sketches \cite{xu2020deep}. 
Some works utilize pixel representation \cite{li2019photosketching, song2018learning, human-like-sketches, xie2015holistically}, 
while others employ vector representation \cite{SketchRNN, Lin2020SketchBERTLS, Bhunia2020PixelorAC, Ribeiro2020SketchformerTR, Chen2017Sketchpix2seqAM, mo2021virtualsketching, Deformable_Stroke, bhunia2021doodleformer, mihai2021learning}. 
Several works propose a unified algorithm to produce sketches with a variety of styles \cite{chan2022learning, yi2020unpaired, liu2021neural} or at varying levels of abstraction \cite{Berger2013,Deep-Sketch-Abstraction,vinker2022clipasso,clipascene}. 
Traditional methods for sketch generation commonly rely on human-drawn sketch datasets. More recently, some works~\cite{CLIPDraw, vinker2022clipasso, clipascene} incorporated the prior of large pretrained language-vision models to eliminate the dependency on such datasets.
We also rely on such priors, and use a vector-based approach to depict our sketches, as it is a more natural representation for sketches and finds widespread use in character animation.

\vspace{-0.15cm}
\paragraph{Sketch-based animation}
A long-standing area of interest in computer graphics aims to develop intuitive tools for creating life-like animations from still inputs.
In character animation, motion is often represented as a temporal sequence of poses. These poses are commonly represented via user provided annotations, such as stick figures~\cite{davis2006sketching}, skeletons~\cite{Pan2011,ArtiSketch2013}, or partial bone lines~\cite{Oztireli13Differential}. 
An alternative line of work represents motion through user provided 2D paths~\cite{DavisKSketch2008,IgarashiPath98,Gleicher2001,thorne2004motion}, or through space-time curves~\cite{guay2015space}. However, these approaches still require some expertise and manual work to adjust different keyframes.
Some methods assist animation by interactively predicting what users will draw next~\cite{WangVideoTooning2004,Agarwala2004,Xing2015}. However, they still require manual sketching operations for each keyframe.

Rather than relying on user-created motion, some works propose to extract motion from real videos by statistical analysis of datasets \cite{Min2009}, or by applying dynamic deformations extracted from a driving video~\cite{su2018live}. 
Others turn to physically-based motion effects~\cite{DRACOKazi2014, XingEnergyBrushes2016}, or learn to synthesize animations of hand-drawn 2D characters using a set of images depicting the character in various poses~\cite{dvorovznak2018toonsynth,poursaeed2020neural,hinz2022charactergan}.

Drawing on 3D literature, some works aim to "wake up" a photo or a painting, extracting a textured human mesh from the image and moving it using pre-defined animations~\cite{Weng2018PhotoW3,levon2020texanmesh,Hornung2007}.
More recently, given a hand-drawn sketch of a human figure, Smith~\etal~\cite{smith2023method} construct a character rig onto which they re-target motion capture data. Their approach is similarly limited to human figures and a predefined set of movements. Moreover, it commonly requires direct human intervention to fix skeleton joint estimations. 

In contrast to these methods, our method requires only a single sketch and no skeletons or explicit references. Instead, it leverages the strong prior of text-to-video generative models and generalizes across a wide range of animations described by free-form prompts.

\vspace{-0.2cm}
\paragraph{Text-to-video generation}
Early works explored expanding the capabilities of recurrent neural networks~\cite{babaeizadeh2017stochastic,denton2018stochastic,castrejon2019improved}, GANs~\cite{kim2020tivgan,tian2021a,zhu2023motionvideogan,pan2017create, li2018video}, and auto-regressive transformers~\cite{weissenborn2019scaling, yan2021videogpt,wu2021godiva,wu2022nuwa} from image generation to video generation. However, these works primarily focused on generating videos within limited domains.

More recent research extends the capabilities of powerful text-to-image diffusion models to video generation by incorporating additional temporal attention modules into existing models or by temporally aligning an image decoder~\cite{singer2022make,blattmann2023videoldm,videofusion2023,wang2023lavie}. Commonly, such alignment is performed in a latent space~\cite{zhou2023magicvideo,li2023videogen,an2023latentshift,videofusion2023,esser2023structure,wang2023modelscope}.
Others train cascaded diffusion models~\cite{ho2022imagen}, or  learn to directly generate videos within a lower-dimensional 3D latent space~\cite{he2022lvdm}. 

We propose to extract the motion prior from such models and apply it to a vector sketch representation.

\vspace{-0.2cm}
\paragraph{Image-to-video generation}
A closely related research area is image-to-video generation, where the goal is to animate an input image. Make-It-Move~\cite{hu2022make} train an encoder-decoder architecture to generate video sequences conditioned on an input image and a driving text prompt. Latent Motion Diffusion~\cite{hu2023lamd} learn the optical flow between pairs of video frames and use a 3D-UNet-based diffusion model to generate the resulting video sequence. CoDi~\cite{tang2023anytoany} align multiple modalities (text, image, audio, and video) to a shared conditioning space and output space. ModelScope~\cite{wang2023modelscope} train a latent video diffusion model, conditioned on an image input. Others first caption an image, then use the caption to condition a text-to-video model~\cite{videofusion2023}. VideoCrafter~\cite{chen2023videocrafter1} train a model conditioned on both text and image, with a special focus preserving the content, structure, and style of this image. Gen-2~\cite{gen2runway} also operate in this domain, though their model's details are not public. 

While showing impressive results in the pixel domain, these methods struggle to generalize to sketches. Our method is designed for sketches, constraining the outputs to vector representations that better preserve both the domain, and the characteristics of the input sketch.

\setlength{\abovedisplayskip}{5pt}
\setlength{\belowdisplayskip}{5pt}

\section{Preliminaries}

\paragraph{Vector representation}
Vector graphics allow us to create visual images directly from geometric shapes such as points, lines, curves, and polygons. Unlike raster images (represented with pixels), vector representation is resolution-free, more compact, and easier to modify. This quality makes vector images the preferred choice for various design applications, such as logo design, prints, animation, CAD, typography, web design, infographics, and illustrations.
Scalable Vector Graphics (SVG) stands out as a popular vector image format due to its excellent support for interactivity and animation. 
We employ a differentiable rasterizer~\cite{diffvg} 
to convert a vector image into its pixel-based image. This lets us manipulate the vector content using raster-based loss functions, as described below.

\vspace{-0.15cm}
\paragraph{Score-Distillation Sampling}
The score-distillation sampling (SDS) loss, first proposed in Poole \etal~\cite{poole2022dreamfusion}, serves as a means for extracting a signal from a pretrained text-to-image diffusion model.

In their seminal work, Poole~\etal propose to first use a parametric image synthesis model (e.g., a NeRF~\cite{mildenhall2020nerf}) to generate an image $x$. This image is then noised to some intermediate diffusion time step $t$: 
\begin{equation}
    x_t = \alpha_t x + \sigma_t \epsilon,
\end{equation}
where $\alpha_t$, $\sigma_t$ are parameters dependant on the noising schedule of the pretrained diffusion model, and $\epsilon \in \mathbb{N}\left(0, 1\right)$ is a noise sample. 

The noised image is then passed through the diffusion model, conditioned on a text-prompt $c$ describing some desired scene. The diffusion model's output, $\epsilon_\theta(x_t,t,c)$, is a prediction of the noise added to the image. The deviation of this prediction from the true noise, $\epsilon$, can serve as a measure of the difference between the input image and one that better matches the prompt. This measure can then be used to approximate the gradients to the initial image synthesis model's parameters, $\phi$, that would better align its outputs with the prompt. Specifically, 
\begin{equation}\label{eq:sds_loss}
    \nabla_\phi \mathcal{L}_{SDS} = \left[ w(t)(\epsilon_\theta(x_t,t,y) - \epsilon) \frac{\partial x}{\partial \phi} \right] ,
\end{equation}
where $w(t)$ is a constant that depends on $\alpha_t$. This optimization process is repeated, with the parametric model converging toward outputs that match the conditioning prompt.

In our work, we use this approach to extract the motion prior learned by a text-to-video diffusion model.

\section{Method}
\label{sec:method}

Our method begins with two inputs: a user-provided static sketch in vector format, and a text prompt describing the desired motion. Our goal is to generate a short video, in the same vector format, which depicts the sketched subject acting in a manner consistent with the prompt.
We therefore define three objectives that our approach should strive to meet: (1) the output video should match the text prompt, (2) the characteristics of the original sketch should be preserved, and (3) the generated motion should appear natural and smooth.
Below, we outline the design choices we use to meet each of these objectives.

\subsection{Representation}
\begin{figure}
    \centering
     \setlength{\abovecaptionskip}{8pt}
     \setlength{\belowcaptionskip}{-8pt}
     \includegraphics[width=0.6\linewidth]{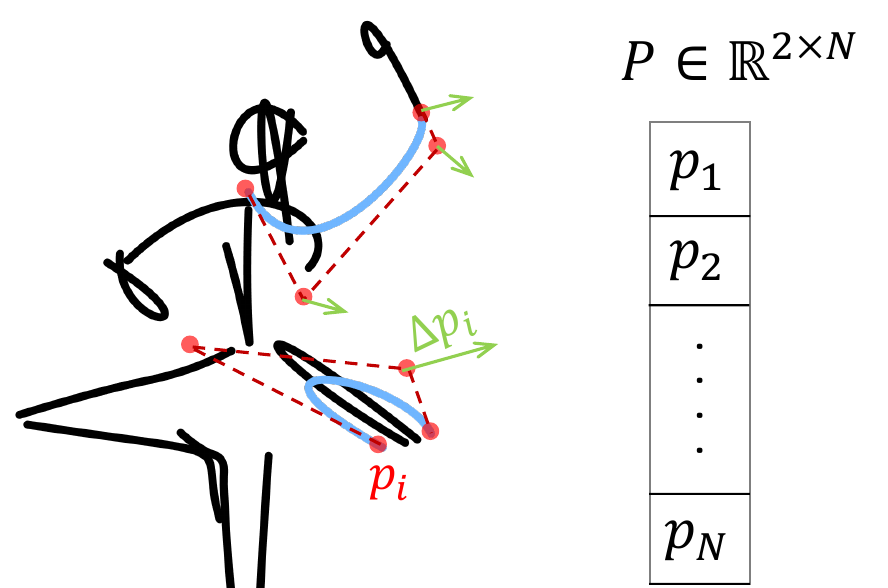}
    \caption{Data representation. Each curve (black or blue) is a cubic \Bezier curve with $4$ control points (red, shown for the blue curves). The total number of control points in the given sketch is denoted by $N$. For each frame and control point $p_i$, we learn a displacement $\Delta p_i$ (green).
    }
    \label{fig:data_rep}
\end{figure}

The input vector image is represented as a set of strokes placed over a white background, where each stroke is a two-dimensional \Bezier curve with four control points. Each control point is represented by its coordinates: $p = (x, y) \in \mathbb{R}^2$. We denote the set of control points in a single frame with $P=\{p_1, .. p_N\} \in \mathbb{R}^{N \times 2}$, where $N$ denotes the total number of points in the input sketch (see \Cref{fig:data_rep}). This number will remain fixed across all generated frames.
We define a video with $k$ frames as a sequence of $k$ such sets of control points, and denote it by $Z = \{P^j\}_{j=1}^k \in \mathbb{R}^{N \cdot k \times 2}$. 

Let $P^{init}$ denote the set of points in the initial sketch. We duplicate $P^{init}$ $k$ times to create the initial set of frames $Z^{init}$.
Our goal is to convert such a static sequence of frames into a sequence of frames animating the subject according to the motion described in the text prompt.
We formulate this task as learning a set of 2D displacements $\Delta Z = \{\Delta p_i^j\}_{i\in N}^{j\in k}$, indicating the displacement of each point $p_i^j$, for each frame $j$ (\cref{fig:data_rep}, in green).

\begin{figure}
    \centering

     \setlength{\abovecaptionskip}{8pt}
     \setlength{\belowcaptionskip}{-10pt}
     
    \includegraphics[width=0.99\linewidth]{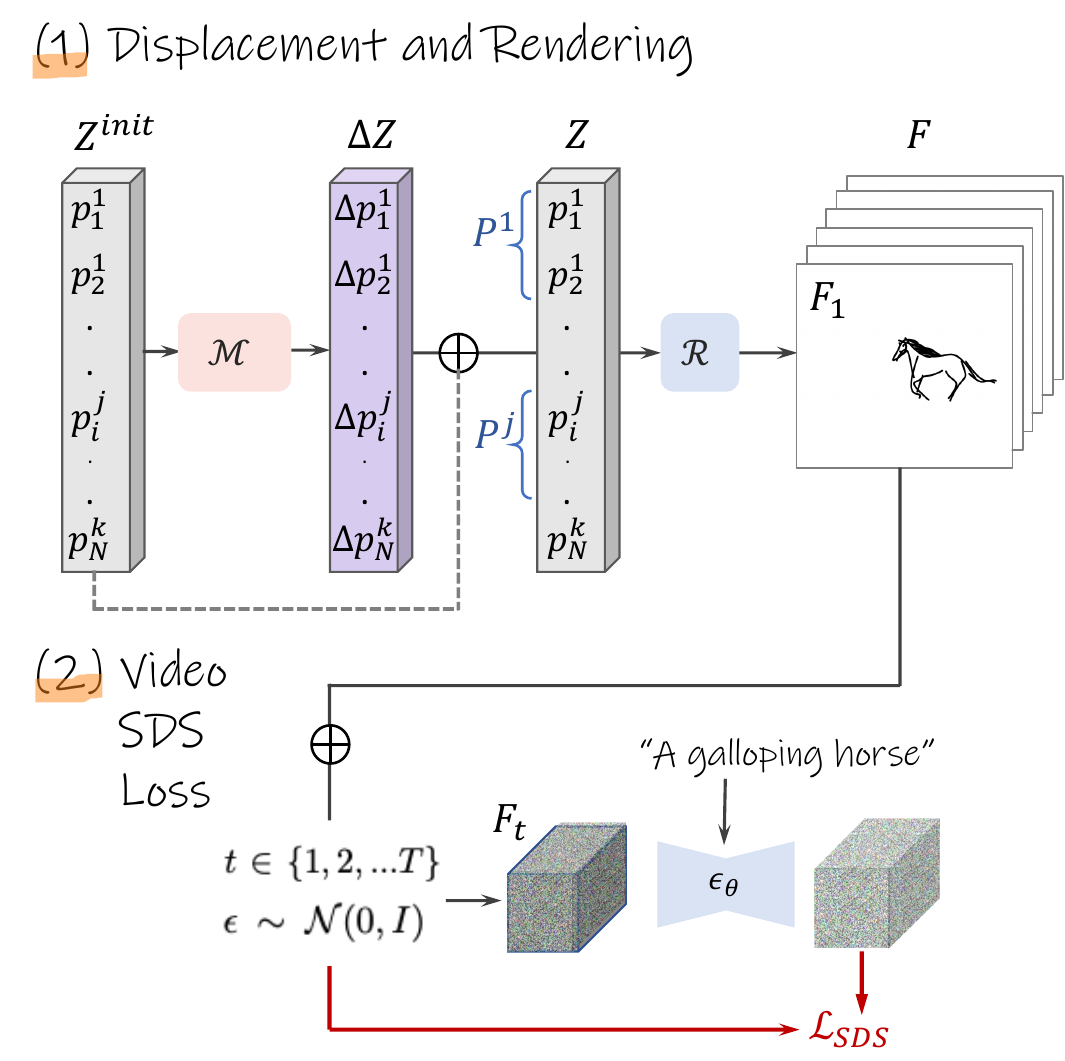}
    \caption{Text-driven optimization. At each training iteration: (1) We duplicate the initial control points across $k$ frames and sum them with their predicted offsets. We render each frame and concatenate them to create the output video. (2) We use the SDS loss to extract a signal from a pretrained text-to-video model, which is used to update $\mathcal{M}$, the model that predicts the offsets.}
    \label{fig:pipe_part1}
\end{figure}

\begin{figure*}
    \centering
     \setlength{\abovecaptionskip}{8pt}
     \setlength{\belowcaptionskip}{-6pt}
    \includegraphics[width=0.89\linewidth]{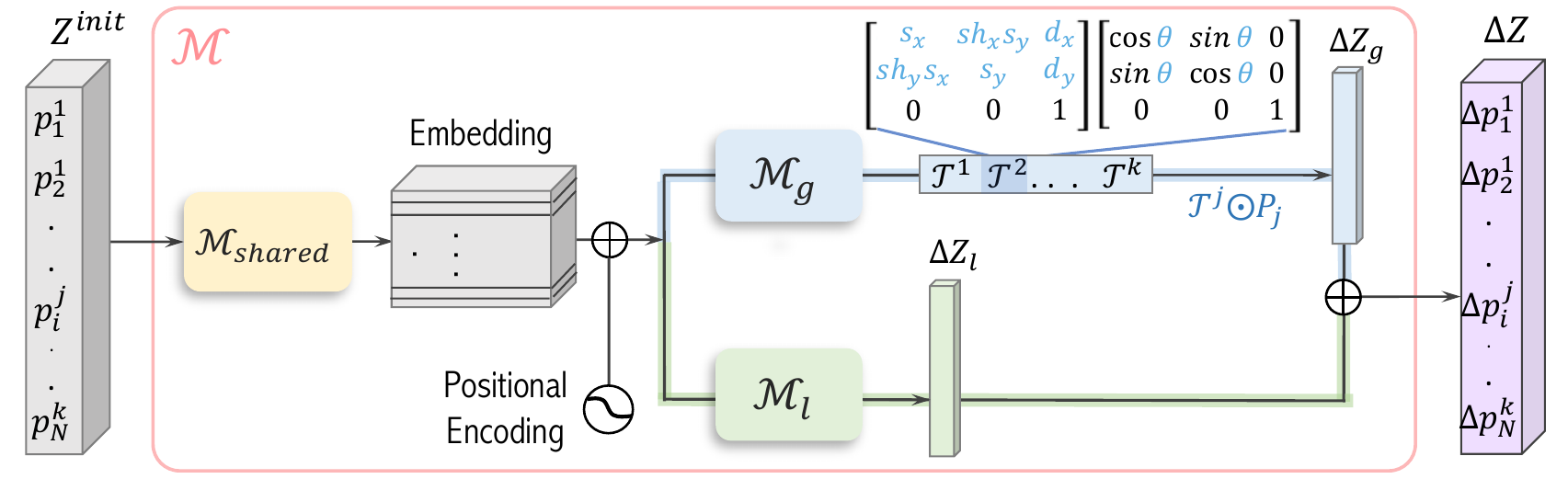}
    \caption{Network architecture. The input to the network is the initial set of control points $Z^{init}$ (left, gray), and the output is the set of displacements $\Delta Z$. The network consists of three parts. First, each control point $p_i^j$ is projected with $M_{shared}$ into a latent representation and summed with a positional encoding. These point features are passed to two different branches to predict global and local motion. The local motion predictor $M_l$ (green) is a simple MLP that predicts an offset for each point ($\Delta Z_l$), representing unconstrained local motion. The global motion predictor $M_g$ predicts a per-frame transformation matrix $\gtransform^j$ which applies scaling, shear, rotation, and translation. $\gtransform^j$ is then applied to the points $P_j$ in the corresponding frame to produce $\Delta Z_g$. $\Delta Z$ is given by the sum: $\Delta Z =\Delta Z_g +\Delta Z_l$. }
    \label{fig:architecture}
\end{figure*}

\subsection{Text-Driven Optimization}

\label{subsec:optimize_loss}

We begin by addressing our first objective: creating an output animation that aligns with the text prompt. We model the animation using a ``neural displacement field'' (\cref{sec:neural_field}), a small network $\mathcal{M}$ that receives as input the initial point set $Z^{init}$ and predicts their displacements $\mathcal{M}(Z^{init}) = \Delta Z$. To train this network, we distill the motion prior encapsulated in a pretrained text-to-video diffusion model \cite{videocomposer2023}, using the SDS loss of \cref{eq:sds_loss}.

At each training iteration (illustrated in \cref{fig:pipe_part1}), we add the predicted displacement vector $\Delta Z$ (marked in purple) to the initial set of points $Z^{init}$ to form the sequence $Z$. 
We then use a differentiable rasterizer $\mathcal{R}$~\cite{diffvg}, to transfer each set of per-frame points $P^j$ to its corresponding frame in pixel space, denoted as $F^j = \mathcal{R}(P^j)$. The animated sketch is then defined by the concatenation of the rasterized frames, $F = \{F^1, .. F^k\} \in \mathbb{R}^{h\times w \times k}$. 

Next, we sample a diffusion timestep $t$ and noise $\epsilon \sim \mathcal{N}\left(0,1\right)$. We use these to add noise to the rasterized video according to the diffusion schedule, creating $F_t$. This noisy video is then denoised using the pretrained text-to-video diffusion model $\epsilon_{\theta}$, where the diffusion model is conditioned on a prompt describing an animated scene (e.g., ``a galloping horse''). Finally, we use \cref{eq:sds_loss} to update the parameters of $\mathcal{M}$ and repeat the process iteratively.

The SDS loss thus guides $\mathcal{M}$ to learn displacements whose corresponding rasterized animation aligns with the desired text prompt. The extent of this alignment, and hence the intensity of the motion, is determined by optimization hyperparameters such as the diffusion guidance scale and learning rates. 
However, we find that increasing these parameters typically leads to artifacts such as jitter and shape-deformations, compromising both the fidelity of the original sketch and the fluidity of natural motion (see \cref{sec:ablation}).
As such, SDS alone fails to address our additional goals: (2) preserving the input sketch characteristics, and (3) creating natural motion. Instead, we tackle these goals through the design of our displacement field, $\mathcal{M}$.

\subsection{Neural Displacement Field}\label{sec:neural_field}

We approach the network design with the intent of producing smoother motion with reduced shape deformations. We hypothesize that the artifacts observed with the unconstrained SDS optimization approach can be attributed in part to two mechanisms: (1) The SDS loss can be minimized by deforming the generated shape into one that better aligns with the text-to-video model's semantic prior (e.g., prompting for a scuttling crab may lead to undesired changes in the shape of the crab itself). (2) Smooth motion requires small displacements at the local scale, and the network struggles to reconcile these with the large changes required for global translations. We propose to tackle both of these challenges by modeling our motion through two components: An unconstrained local-motion path, which models small deformations, and a global path which models affine transformations applied uniformly to an entire frame. This split will allow the network to separately model motion along both scales while restricting semantic changes in the path that controls greater scale movement. Below we outline the specific network design choices, as well as the parametrization that allows us to achieve this split.

\vspace{-0.2cm}
\paragraph{Shared backbone}
Recall that our network, illustrated in \cref{fig:architecture}, aims to map the initial control point set $Z^{init}$ to their per-frame displacements $\mathcal{M}(Z^{init}) = \Delta Z$. Our first step is to create a shared feature backbone which will feed the separate motion paths. This component is built of an embedding step, where the coordinates of each control point are projected using a shared matrix $\mathcal{M}_{shared}$, and then summed with a positional encoding that depends on the frame index, and on the order of the point in the sketch. These point features are then fed into two parallel prediction paths: local, and global (\cref{fig:architecture}, green and blue paths, respectively).

\vspace{-0.2cm}
\paragraph{Local path} The local path is parameterized by $\mathcal{M}_l$, a small MLP that takes the shared features and maps them to an offset $\Delta Z_l$ for every control point in $Z^{init}$. Here, the goal is to allow the network to learn unconstrained motion on its own to best match the given prompt. Indeed, in \cref{sec:results} we show that an unconstrained branch is crucial for the model to create meaningful motion. On the other hand, using this path to create displacements on the scale needed for global changes requires stronger SDS guidance or larger learning rates, leading to jitter and unwanted deformations at the local level. Hence, we delegate these changes to the global motion path. 
We note that similar behavior can be observed when directly optimizing the control points (\ie without a network, following \cite{jain2022vectorfusion,IluzVinker2023}, see \cref{sec:ablation}).

\begin{figure*}[!t]
    \centering
     \setlength{\abovecaptionskip}{5pt}
     \setlength{\belowcaptionskip}{-6pt}
     \includegraphics[width=0.95\linewidth]{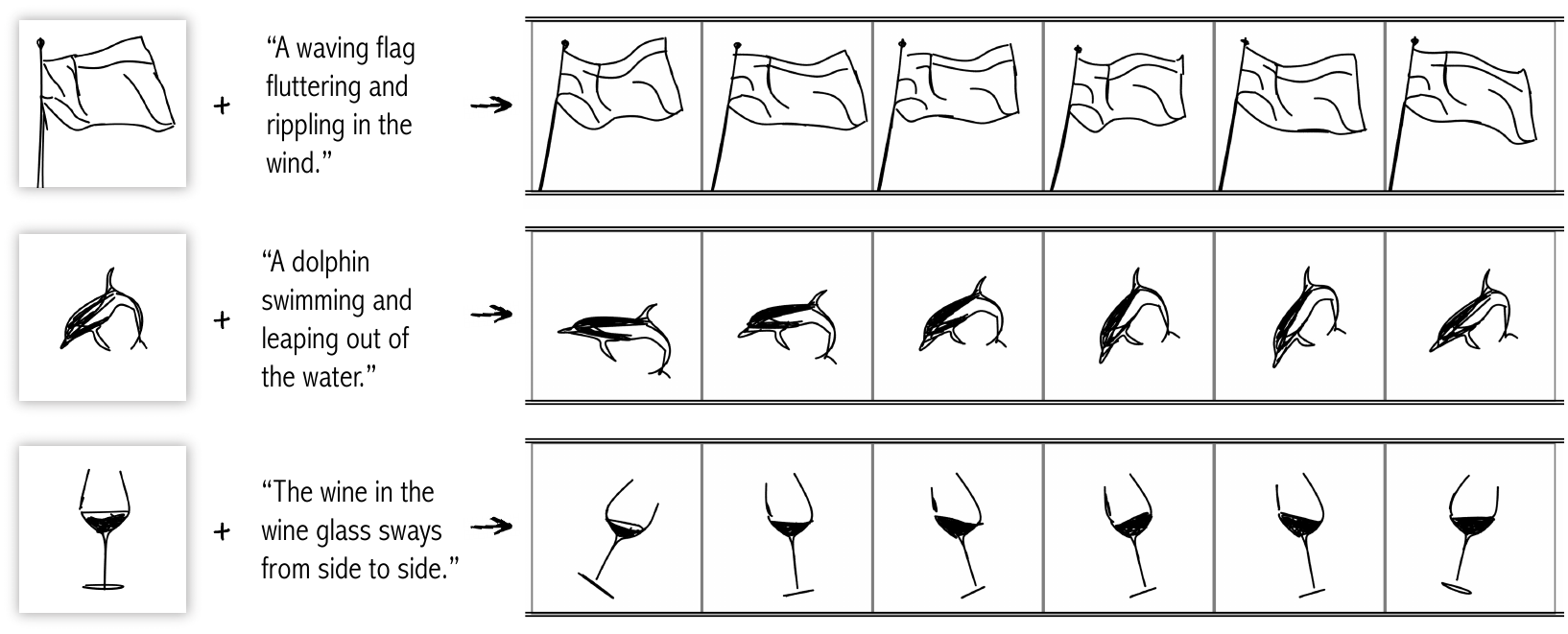}
    \caption{Qualitative results. Our model converts an initial sketch and a driving prompt describing some desired motion into a short video depicting the sketch moving according to the prompt. See the supplementary for the full videos and additional results.}
    \label{fig:ours_qualitative}
\end{figure*}

\vspace{-0.2cm}
\paragraph{Global path}
The goal of the global displacement prediction branch is to allow the model to capture meaningful global movements such as center-of-mass translation, rotation, or scaling, while maintaining the object's original shape. This path consists of a neural network, $\mathcal{M}_g$, that predicts a single global transformation matrix for each frame $P^j$. The matrix is then used to transform all control points of that frame, ensuring that the shape remains coherent. Specifically, we model the global motion as the sequential application of scaling, shear, rotation, and translation. These are parameterized using their standard affine matrix form (\cref{fig:architecture}), which contains two parameters each for scale, shear, and translation, and one for rotation.  %
Denoting the successive application of these transforms for frame $j$ by $\gtransform^j$, the global branch displacement for each point in this frame is then given by: $\Delta p^j_{i,global} = \gtransform^j \odot p^{init}_i - p^{init}_i$.

We further extend the user's control over the individual components of the generated motion by adding a scaling parameter for each type of transformation: $\lambda_t, \lambda_r, \lambda_s$ and $\lambda_{sh}$ for translation, rotation, scale, and shear, respectively. For example, let $(d_x^j, d_y^j)$ denote the network's predicted translation parameters. We re-scale them as: $(d_x^j, d_y^j) \rightarrow (\lambda_{t}d_x^j, \lambda_{t}d_y^j)$. This allows us to attenuate specific aspects of motion that are undesired. For example, we can keep a subject roughly stationary by setting $\lambda_t = 0$. By modeling global changes through constrained transformations, applied uniformly to the entire frame, we limit the model's ability to create arbitrary deformations while preserving its ability to create large translations or coordinated effects.

\mbox{} \\
Our final predicted displacements $\Delta Z$ are simply the sum of the two branches: $\Delta Z_l + \Delta Z_g$. The strength of these two terms (governed by the learning rates and guidance scales used to optimize each branch) will affect a tradeoff between our first goal (text-to-video alignment), and the other two goals (preserving the shape of the original sketch and creating smooth and natural motion). As such, a user can use this tradeoff to gain additional control over the generated video. For instance, prioritizing the preservation of sketch appearance by using a low learning rate for the local path, while affording greater freedom to global motion. We further demonstrate this tradeoff in the supplementary.

\subsection{Training Details}

We alternate between optimizing the local path and optimizing the global path. The shared backbone is optimized in both cases. Unless otherwise noted, we set the SDS guidance scale to $30$ for the local path and $40$ for the global path. We use Adam~\cite{kingma2014adam} with a learning rate of $1\mathrm{e}{-4}$ for the local path and a learning rate of $5\mathrm{e}{-3}$ for the global path. We find it useful to apply augmentations (random crops and perspective transformations) to the rendered videos during training. We further set $\lambda_t = 1.0, \lambda_r = 1\mathrm{e}{-2}, \lambda_s = 5\mathrm{e}{-2}, \lambda_{sh} = 1\mathrm{e}{-1}$. 
For our diffusion backbone, we use ModelScope text-to-video~\cite{wang2023modelscope}, but observe similar results with other backbones (see the supplementary file).

We optimize the networks for $1,000$ steps, taking roughly $30$ minutes per video on a single A100 GPU. In practice, the model often converges after $500$ steps ($15$ minutes). For additional training details, see the supplementary.

\section{Results}
\label{sec:results}
\begin{figure}
    \centering
     \includegraphics[width=1\linewidth]{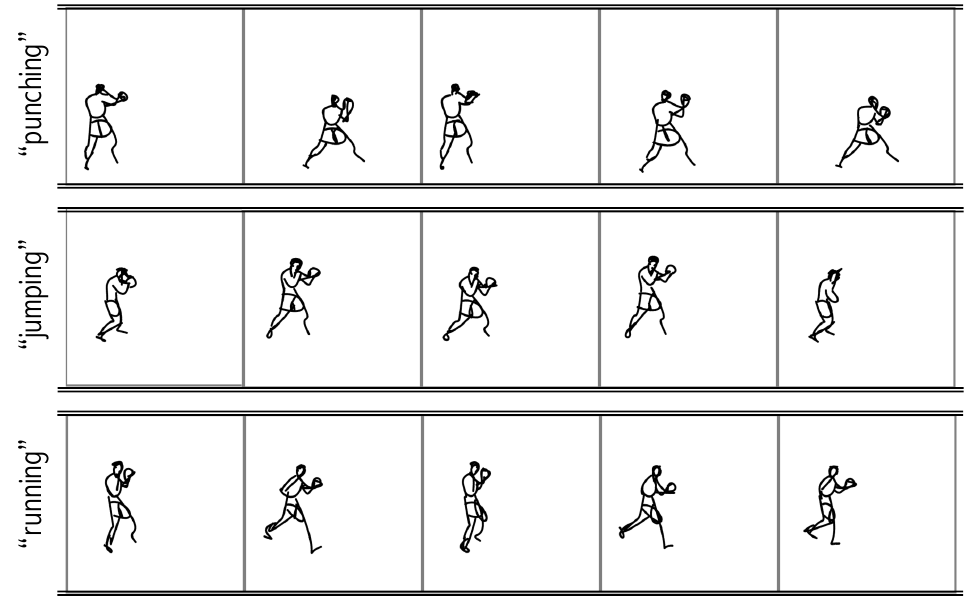}
    \caption{Our method can be used to animate the same sketch according to different prompts. These are typically restricted to actions that the portrayed subject would naturally perform. See the supplementary videos for more examples.}
    \vspace{-0.2cm}
    \label{fig:boxer_prompts}
\end{figure}

We begin by showcasing our method's ability to animate a diverse set of sketches, following an array of text prompts (see \cref{fig:ours_qualitative} and supplementary videos). Our method can capture the delicate swaying of a dolphin in the water, follow a ballerina's dance routine, or mimic the gentle motion of wine swirling in a glass. Notably, it can apply these motions to sketches without any common skeleton or an explicit notation of parts. Moreover, our approach can animate the same sketch using different prompts (see \cref{fig:boxer_prompts}), extending the freedom and diversity of text-to-video models to the more abstract sketch domain. 
Additional examples and full videos can be found in the supplementary materials.

\begin{figure}
    \centering
     \setlength{\belowcaptionskip}{-6pt}
    \includegraphics[width=1\linewidth]{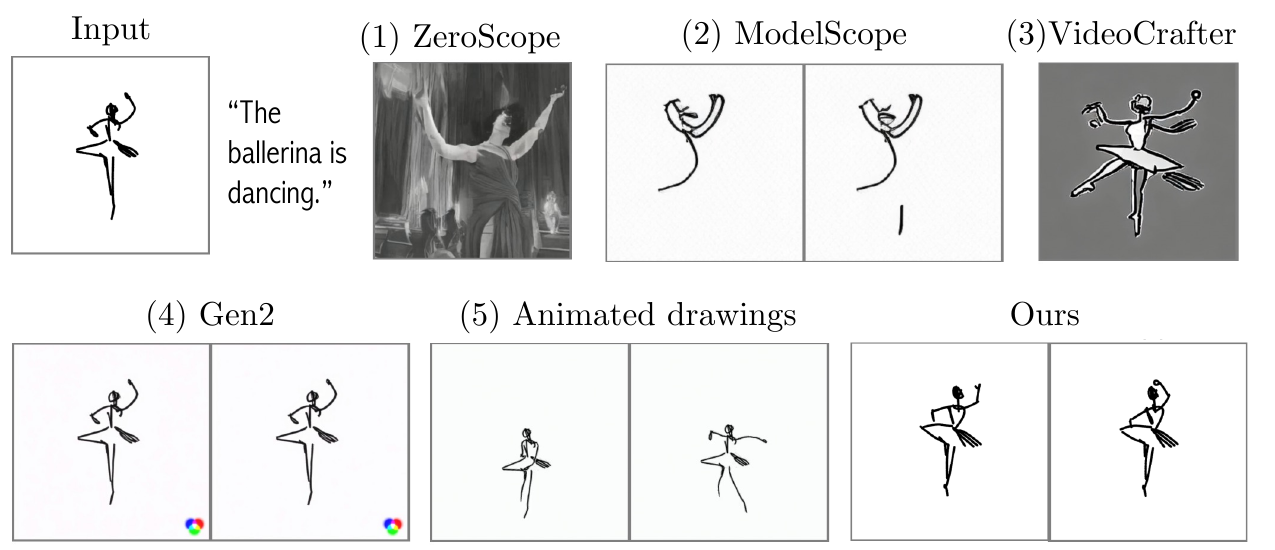}
    \caption{Qualitative comparisons. Image-to-video models suffer from artifacts and struggle to preserve the sketch shape (or even remain in a sketch domain). Animated drawings relies on skeletons and pre-captured reference motions. Hence, it cannot generalize to new domains. See the supplementary videos for more examples. }
    \label{fig:image-to-video}
\end{figure}

\subsection{Comparisons}\label{sec:comparison}
As no prior art directly tackles the reference-free sketch animation task, we explore two alternative approaches: pixel-based image-to-video approaches, and skeleton-based methods that build on pre-defined motions.

In the pixel-based scenario, we compare our method with four models: (1) ZeroScope image-to-video~\cite{videofusion2023} which automatically captions the image~\cite{Yu2022CoCaCC} and uses the caption to prompt a text-to-video model. (2) ModelScope~\cite{wang2023modelscope} image-to-video, which is directly conditioned on the image. (3) VideoCrafter~\cite{chen2023videocrafter1} which is conditioned on both the image and the given text prompt, and (4) Gen-2~\cite{gen2runway}, a commercial web-based tool, conditioned on both image and text. 

The results are shown in \cref{fig:image-to-video}. We select representative frames from the output videos. The full videos are available in the supplementary material.

The results of ZeroScope and VideoCrafter show significant artifacts, and commonly fail to even produce a sketch. ModelScope fare batter, but struggle to preserve the shape of the sketch. Gen-2 either struggle to animate the sketch, or transforms it into a real image, depending on the input parameters (see the supplementary videos).

We further compare our approach with a skeleton and reference-based method~\cite{smith2023method} (\cref{fig:image-to-video}, Animated Drawings). This method accounts for the sketch-based nature of our data and can better preserve its shape. However, it requires per-sketch manual annotations and is restricted to a pre-determined set of human motions. Hence, it struggles to animate subjects which cannot be matched to a human skeleton, or whose motion does not align with the presets (see supplementary). In contrast, our method inherits the diversity of the text-to-video model and generalizes to multiple target classes without annotations or explicit references.

We additionally evaluate our method quantitatively. We compare with open methods that require no human intervention and can be evaluated at scale (ZeroScope~\cite{videofusion2023}, ModelScope~\cite{wang2023modelscope}, and Videocrafter~\cite{chen2023videocrafter1}).
We follow \citep{clipascene} and collect sketches spanning three categories: humans, animals, and objects. We asked ChatGPT to randomly select ten instances per category and suggest prompts describing their typical motion.
We used CLIPasso~\cite{vinker2022clipasso} to generate a sketch for each subject. We applied our method and the alternative methods to these sketches and prompts, resulting in 30 animations per method (videos in the supplementary).

Following pixel-based methods~\cite{esser2023structure,chen2023videocrafter1}, we use CLIP~\cite{radford2021learning} to measure the \ap{sketch-to-video} consistency, defined as the average cosine similarity between the video's frames and the input sketch used to produce it. 

We further evaluate the alignment between the generated videos and their corresponding prompts (\ap{text-to-video alignment}). We use X-CLIP~\cite{XCLIP}, a model that extends CLIP to video recognition. Here, we compare to the only baseline which is jointly guided by both image and text~\cite{chen2023videocrafter1}. 

All results are provided in \cref{tab:all_quant}a. 
Our method outperforms the baselines on sketch-to-video consistency. In particular, it achieves significant gains over ModelScope whose text-to-video model serves as our prior. Moreover, our approach better aligns with the prompted motion, despite the use of a weaker text-to-video model as a backbone. These results, and in particular the ModelScope scores, demonstrate the importance of the vector representation which assists us in successfully extracting a motion prior without the low quality and artifacts introduced when trying to create sketches in the pixel domain.

\begin{figure}
    \centering
    \setlength{\abovecaptionskip}{5pt}
     \setlength{\belowcaptionskip}{-8pt}
    \includegraphics[width=0.98\linewidth]{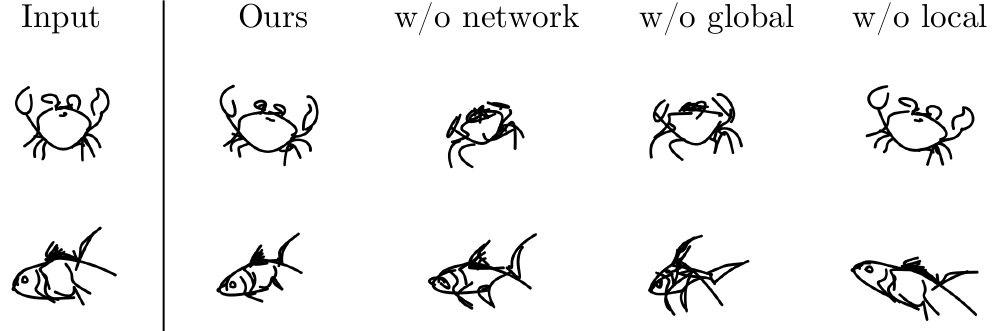}
    \caption{Qualitative ablation. Removing the neural network or the global path leads to shape deviations or jittery motion due to the need for higher learning rates (see supplementary videos). Modeling only global movement improves shape consistency, but fails to create realistic motion.}
    \label{fig:ablation_qualitative}
\end{figure}

\begin{table*}[hbt]\setlength{\tabcolsep}{3pt}
\vspace{-4pt}
\setlength{\abovecaptionskip}{6.5pt}
\footnotesize
\centering 
\begin{tabular}{c c c}

\begin{tabular}{lcc} 
    \toprule
    \multirow{2}{*}{Method} & Sketch-to-video & Text-to-Video  \\
     & consistency $\left(\uparrow\right)$ & alignment$\left(\uparrow\right)$ \\
    \midrule 
    ZeroScope & $0.754 \pm 0.009$ & - \\
    ModelScope & $0.779 \pm 0.009$ & - \\
    VideoCrafter & $0.876 \pm 0.007$ & $0.124 \pm 0.005$ \\
    Ours & $\mathbf{0.965} \pm 0.003$ & $\mathbf{0.142} \pm 0.005$ \\
    \bottomrule \\
    \multicolumn{3}{c}{(a) Comparisons to pixel-based approaches}
\end{tabular} &

\begin{tabular}{lcc} 
    \toprule
    \multirow{2}{*}{Setup} 
     & Sketch-to-video & Text-to-Video \\
     & consistency $\left(\uparrow\right)$ & alignment$\left(\uparrow\right)$ \\
    \midrule 
    Full & $0.965 \pm 0.003$ & $0.142 \pm 0.005$ \\
    No Net & $0.926 \pm 0.007$ & $0.142 \pm 0.005$  \\
    No Glob. & $0.936 \pm 0.006$ & $0.140 \pm 0.005$  \\
    No Local & $0.970 \pm 0.002$ & $0.140 \pm 0.004$ \\
    \bottomrule \\
    \multicolumn{3}{c}{(b) Ablation results}
\end{tabular} &

\begin{tabular}{c} 
\includegraphics[width=0.34\linewidth, height=3cm]{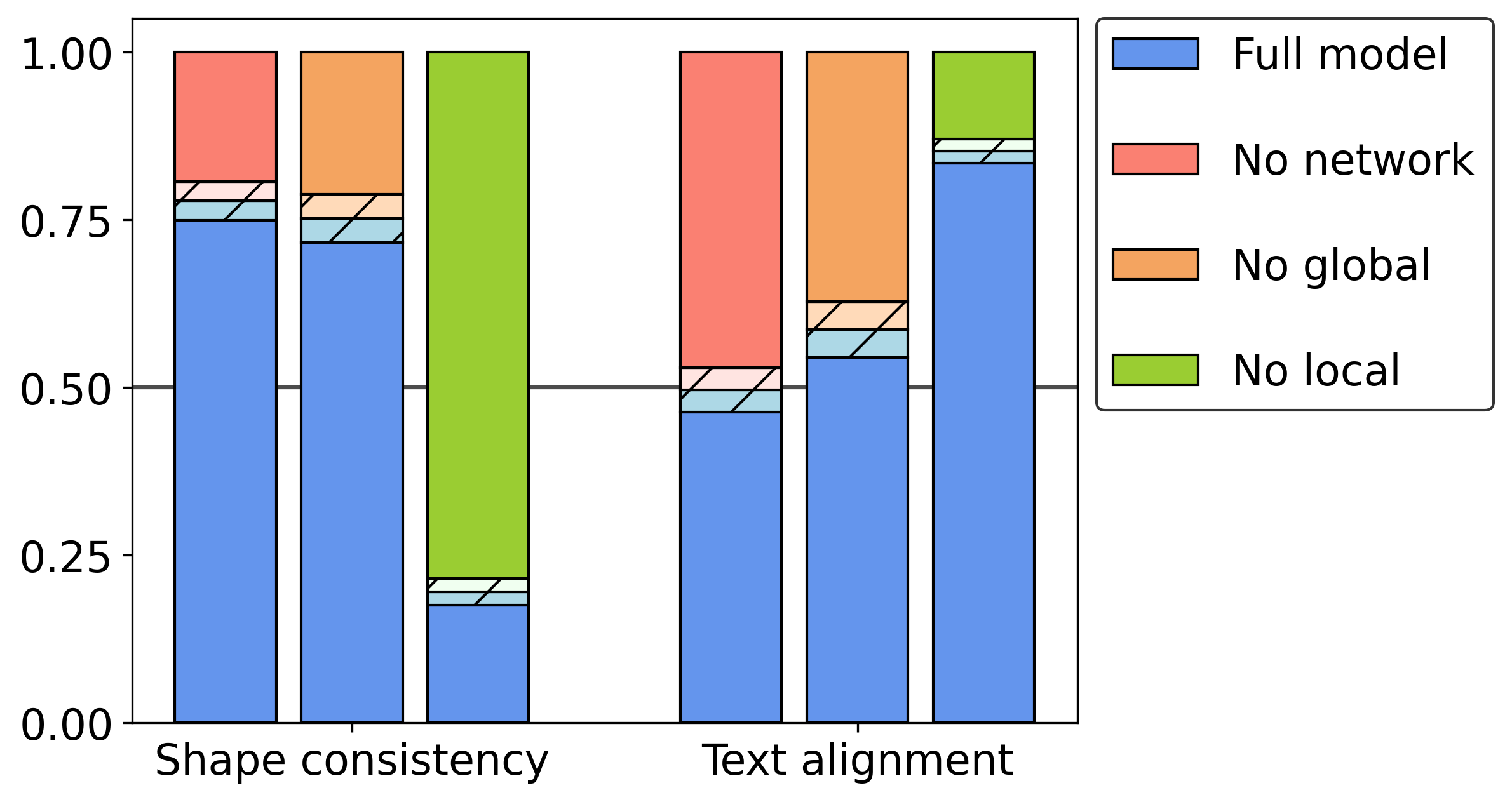} \\
   (c) User study
\end{tabular}
\end{tabular}
\caption{Quantitative metrics. \textbf{(a)} CLIP-based consistency and text-video alignment comparisons to open-source image-to-video baselines. \textbf{(b)} The same CLIP-metrics used for an ablation study. \textbf{(c)} User study results. We pit our full model against each ablation setup. The blue bar indicates the percent of responders that preferred our full model over each baseline. Dashed area is one standard error. }\label{tab:all_quant}
\vspace{-8pt}
\end{table*}

\subsection{Ablation Study}\label{sec:ablation}

We further validate our suggested components through an ablation study. In particular, we evaluate the effect of using the neural prior in place of direct coordinate optimization and the effect of the global-local separation. 

Qualitative results are shown in \cref{fig:ablation_qualitative} (the corresponding videos are provided in the supplementary materials). As can be observed, removing the neural network can lead to increased jitter and harms shape preservation. Removing the global path leads to diminished movement across the frame and less coherent shape transformations. In contrast, removing the local path leads to unrealistic wobbling while keeping the original sketch almost unchanged.

In \cref{tab:all_quant}b, we show quantitative results, following the same protocol as in \cref{sec:comparison}. The sketch-to-video consistency results align with the qualitative observations. However, we observe that the metric for text-to-video alignment \cite{XCLIP} is not sensitive enough to gauge the difference between our ablation setups (standard errors are larger than the gaps). 

We additionally conduct a user study, based on a two-alternative forced-choice setup.
Each user is shown two videos (one output from the full method, and one from a random ablation setup) and asked to select: (1) the video that better preserves the appearance of the initial sketch, and (2) the video that better matches the motion outlined in the prompt. We collected responses from 31 participants over $30$ pairs. The results are provided in \cref{tab:all_quant}c. 

Users considered the full method's text-to-video alignment to be on-par or better than all ablation setups. When considering sketch-to-video consistency, our method is preferred over both setups that create reasonable motion (no network and no global). Removing the local path leads to higher consistency with respect to the original frame, largely because the sketch remains almost unchanged. Our full method allows for more expressive motion, while still showing remarkable preservation of the input sketch. 

In the supplementary materials, we provide further analysis on the effects of our hyperparameter choices, and highlight an emergent trade-off between shape preservation and the quality of generated motion.

\vspace{-3pt}
\section{Limitations}
\begin{figure}
    \centering
    \setlength{\abovecaptionskip}{7pt}
     \setlength{\belowcaptionskip}{-7pt}
    \includegraphics[width=1\linewidth]{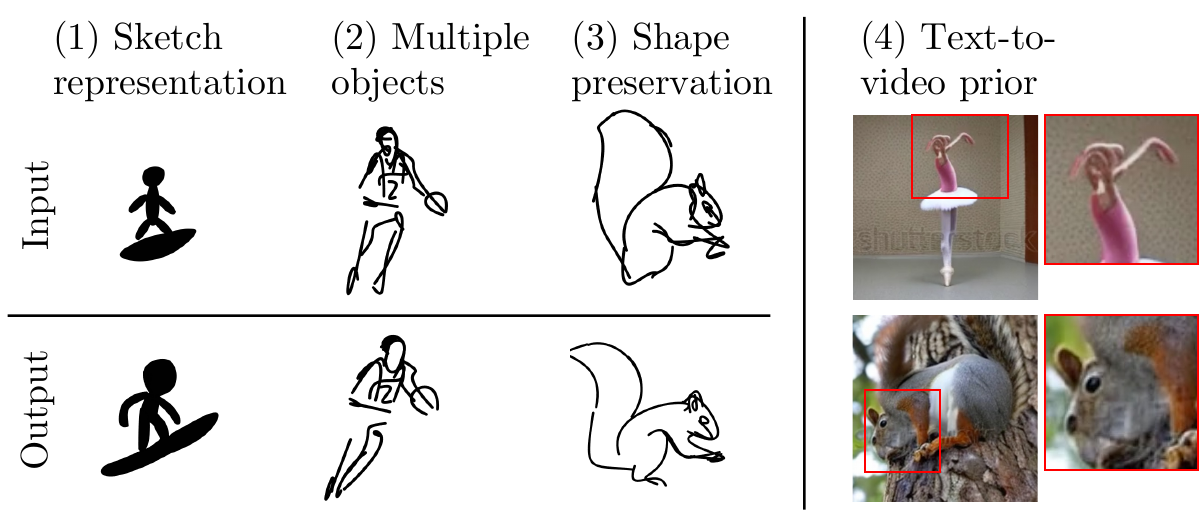}
    \caption{Method limitations. The method may struggle with certain sketch representations, fail to tackle multiple objects or complex scenes, or create undesired shape changes. Moreover, it is restricted to motions which the text-to-video prior can create.}
    \label{fig:limitations}
\end{figure}

While our work enables sketch-animation across various classes and prompts, it comes with limitations.
First, we build upon the sketch representation from \cite{vinker2022clipasso}. However, sketches can be represented in many forms with different types of curves and primitive shapes. Using our method with other sketch representations could result in performance degradation. For instance, in \cref{fig:limitations}(1) the surfer's scale has significantly changed. Addressing the diversity of vector sketches requires further development.
Second, our method assumes a single-subject input sketch (a common scenario in character animation techniques). When applied to scene sketches or sketches with multiple objects, we observe reduced result quality due to the design constraints. For example in \cref{fig:limitations}(2), the basketball cannot be separated from the player, contrary to the natural motion of dribbling.

Third, our method faces a trade-off between motion quality and sketch fidelity, and a diligent balance should be achieved between the two. In \cref{fig:limitations}(3), the animated squirrel's appearance differs from the input sketch. This trade-off is further discussed in the supplementary material. 
Potential improvement lies in adopting a mesh-based representation with an approximate rigidity loss~\cite{igarashi2005arapshape}, or by trying to enforce consistency in the diffusion feature space~\cite{tokenflow2023} .

Finally, our approach inherits the limitations of text-to-video priors. Such models are trained on large-scale data, but may be unaware of specific motions, or portray strong biases. For example, as demonstrated in \cref{fig:limitations}, the model we utilize tend to produce significant artefacts when used for text-to-video generation. However, our method is agnostic to the backbone model and hence could likely by used with newer, improved models as they become available, or with personalized models~\cite{gal2022image} that were augmented with new, unobserved motions.

\section{Conclusions}

We presented a technique to breath life into a given static sketch, following a text prompt. Our method builds on the motion prior captured by powerful text-to-video models. We show that even though these models struggle with generating sketches directly, they can still comprehend such abstract representations in a semantically meaningful way, creating smooth and appealing motions. 
We hope that our work will facilitate further research to provide intuitive and practical tools for sketch animation that incorporate recent advances in text-based video generation.

\section{Acknowledgements}
We thank Oren Katzir and Guy Tevet for providing feedback on early versions of this manuscript. 
This work was partially supported by BSF (grant 2020280) and ISF (grants 2492/20 and 3441/21). 

{
    \small
    \bibliographystyle{ieeenat_fullname}
    \bibliography{main}

\begin{thebibliography}{99}
\providecommand{\natexlab}[1]{#1}
\providecommand{\url}[1]{\texttt{#1}}
\expandafter\ifx\csname urlstyle\endcsname\relax
  \providecommand{\doi}[1]{doi: #1}\else
  \providecommand{\doi}{doi: \begingroup \urlstyle{rm}\Url}\fi

\bibitem[Agarwala et~al.(2004)Agarwala, Hertzmann, Salesin, and Seitz]{Agarwala2004}
Aseem Agarwala, Aaron Hertzmann, David Salesin, and Steven Seitz.
\newblock Keyframe-based tracking of rotoscoping and animation.
\newblock \emph{ACM Trans. Graph.}, 23:\penalty0 584--591, 2004.

\bibitem[An et~al.(2023)An, Zhang, Yang, Gupta, Huang, Luo, and Yin]{an2023latentshift}
Jie An, Songyang Zhang, Harry Yang, Sonal Gupta, Jia-Bin Huang, Jiebo Luo, and Xi Yin.
\newblock Latent-shift: Latent diffusion with temporal shift for efficient text-to-video generation.
\newblock \emph{arXiv preprint arXiv:2304.08477}, 2023.

\bibitem[Aubert et~al.(2014)Aubert, Brumm, Ramli, Sutikna, Saptomo, Hakim, Morwood, van~den Bergh, Kinsley, and Dosseto]{aubert2014pleistocene}
Maxime Aubert, Adam Brumm, Muhammad Ramli, Thomas Sutikna, E~Wahyu Saptomo, Budianto Hakim, Michael~J Morwood, Gerrit~D van~den Bergh, Leslie Kinsley, and Anthony Dosseto.
\newblock Pleistocene cave art from sulawesi, indonesia.
\newblock \emph{Nature}, 514\penalty0 (7521):\penalty0 223--227, 2014.

\bibitem[Babaeizadeh et~al.(2017)Babaeizadeh, Finn, Erhan, Campbell, and Levine]{babaeizadeh2017stochastic}
Mohammad Babaeizadeh, Chelsea Finn, Dumitru Erhan, Roy~H Campbell, and Sergey Levine.
\newblock Stochastic variational video prediction.
\newblock \emph{arXiv preprint arXiv:1710.11252}, 2017.

\bibitem[Berger et~al.(2013)Berger, Shamir, Mahler, Carter, and Hodgins]{Berger2013}
Itamar Berger, Ariel Shamir, Moshe Mahler, Elizabeth Carter, and Jessica Hodgins.
\newblock Style and abstraction in portrait sketching.
\newblock \emph{ACM Trans. Graph.}, 32\penalty0 (4), 2013.

\bibitem[Bhunia et~al.(2020)Bhunia, Das, Muhammad, Yang, Hospedales, Xiang, Gryaditskaya, and Song]{Bhunia2020PixelorAC}
Ayan~Kumar Bhunia, Ayan Das, Umar~Riaz Muhammad, Yongxin Yang, Timothy~M. Hospedales, Tao Xiang, Yulia Gryaditskaya, and Yi-Zhe Song.
\newblock Pixelor: a competitive sketching ai agent. so you think you can sketch?
\newblock \emph{ACM Trans. Graph.}, 39:\penalty0 166:1--166:15, 2020.

\bibitem[Bhunia et~al.(2022)Bhunia, Khan, Cholakkal, Anwer, Khan, Laaksonen, and Felsberg]{bhunia2021doodleformer}
Ankan~Kumar Bhunia, Salman Khan, Hisham Cholakkal, Rao~Muhammad Anwer, Fahad~Shahbaz Khan, Jorma Laaksonen, and Michael Felsberg.
\newblock Doodleformer: Creative sketch drawing with transformers.
\newblock \emph{ECCV}, 2022.

\bibitem[Blattmann et~al.(2023)Blattmann, Rombach, Ling, Dockhorn, Kim, Fidler, and Kreis]{blattmann2023videoldm}
Andreas Blattmann, Robin Rombach, Huan Ling, Tim Dockhorn, Seung~Wook Kim, Sanja Fidler, and Karsten Kreis.
\newblock Align your latents: High-resolution video synthesis with latent diffusion models.
\newblock In \emph{IEEE Conference on Computer Vision and Pattern Recognition ({CVPR})}, 2023.

\bibitem[Bregler et~al.(2002)Bregler, Loeb, Chuang, and Deshpande]{bregler2002turning}
Christoph Bregler, Lorie Loeb, Erika Chuang, and Hrishi Deshpande.
\newblock Turning to the masters: Motion capturing cartoons.
\newblock \emph{ACM Transactions on Graphics (TOG)}, 21\penalty0 (3):\penalty0 399--407, 2002.

\bibitem[Castrejon et~al.(2019)Castrejon, Ballas, and Courville]{castrejon2019improved}
Lluis Castrejon, Nicolas Ballas, and Aaron Courville.
\newblock Improved conditional vrnns for video prediction.
\newblock In \emph{Proceedings of the IEEE/CVF international conference on computer vision}, pages 7608--7617, 2019.

\bibitem[Chan et~al.(2022)Chan, Durand, and Isola]{chan2022learning}
Caroline Chan, Fr{\'e}do Durand, and Phillip Isola.
\newblock Learning to generate line drawings that convey geometry and semantics.
\newblock In \emph{Proceedings of the IEEE/CVF Conference on Computer Vision and Pattern Recognition}, pages 7915--7925, 2022.

\bibitem[Chen et~al.(2023)Chen, Xia, He, Zhang, Cun, Yang, Xing, Liu, Chen, Wang, Weng, and Shan]{chen2023videocrafter1}
Haoxin Chen, Menghan Xia, Yingqing He, Yong Zhang, Xiaodong Cun, Shaoshu Yang, Jinbo Xing, Yaofang Liu, Qifeng Chen, Xintao Wang, Chao Weng, and Ying Shan.
\newblock Videocrafter1: Open diffusion models for high-quality video generation, 2023.

\bibitem[Chen et~al.(2017)Chen, Tu, Yi, and Xu]{Chen2017Sketchpix2seqAM}
Yajing Chen, Shikui Tu, Yuqi Yi, and Lei Xu.
\newblock Sketch-pix2seq: a model to generate sketches of multiple categories.
\newblock \emph{ArXiv}, abs/1709.04121, 2017.

\bibitem[Davis et~al.(2006)Davis, Agrawala, Chuang, Popovi{\'c}, and Salesin]{davis2006sketching}
James Davis, Maneesh Agrawala, Erika Chuang, Zoran Popovi{\'c}, and David Salesin.
\newblock A sketching interface for articulated figure animation.
\newblock In \emph{Acm siggraph 2006 courses}, pages 15--es. 2006.

\bibitem[Davis et~al.(2008)Davis, Colwell, and Landay]{DavisKSketch2008}
Richard~C. Davis, Brien Colwell, and James~A. Landay.
\newblock K-sketch: A 'kinetic' sketch pad for novice animators.
\newblock In \emph{Proceedings of the SIGCHI Conference on Human Factors in Computing Systems}, page 413–422, New York, NY, USA, 2008. Association for Computing Machinery.

\bibitem[Denton and Fergus(2018)]{denton2018stochastic}
Emily Denton and Rob Fergus.
\newblock Stochastic video generation with a learned prior.
\newblock In \emph{International conference on machine learning}, pages 1174--1183. PMLR, 2018.

\bibitem[Dvoro{\v{z}}n{\'a}k et~al.(2018)Dvoro{\v{z}}n{\'a}k, Li, Kim, and S{\`y}kora]{dvorovznak2018toonsynth}
Marek Dvoro{\v{z}}n{\'a}k, Wilmot Li, Vladimir~G Kim, and Daniel S{\`y}kora.
\newblock Toonsynth: example-based synthesis of hand-colored cartoon animations.
\newblock \emph{ACM Transactions on Graphics (TOG)}, 37\penalty0 (4):\penalty0 1--11, 2018.

\bibitem[Eitz et~al.(2012)Eitz, Hays, and Alexa]{eitz2012hdhso}
Mathias Eitz, James Hays, and Marc Alexa.
\newblock How do humans sketch objects?
\newblock \emph{ACM Trans. Graph. (Proc. SIGGRAPH)}, 31\penalty0 (4):\penalty0 44:1--44:10, 2012.

\bibitem[Esser et~al.(2023)Esser, Chiu, Atighehchian, Granskog, and Germanidis]{esser2023structure}
Patrick Esser, Johnathan Chiu, Parmida Atighehchian, Jonathan Granskog, and Anastasis Germanidis.
\newblock Structure and content-guided video synthesis with diffusion models.
\newblock \emph{arXiv preprint arXiv:2302.03011}, 2023.

\bibitem[Fan et~al.(2023)Fan, Bainbridge, Chamberlain, and Wammes]{Fan2023DrawingAA}
Judy Fan, Wilma~A. Bainbridge, Rebecca Chamberlain, and Jeffrey~D. Wammes.
\newblock Drawing as a versatile cognitive tool.
\newblock \emph{Nature Reviews Psychology}, 2:\penalty0 556 -- 568, 2023.

\bibitem[Fan et~al.(2018)Fan, Yamins, and Turk-Browne]{Fan2018CommonOR}
Judith~E. Fan, Daniel L.~K. Yamins, and Nicholas~B. Turk-Browne.
\newblock Common object representations for visual production and recognition.
\newblock \emph{Cognitive science}, 42 8:\penalty0 2670--2698, 2018.

\bibitem[Frans et~al.(2021)Frans, Soros, and Witkowski]{CLIPDraw}
Kevin Frans, Lisa~B. Soros, and Olaf Witkowski.
\newblock Clipdraw: Exploring text-to-drawing synthesis through language-image encoders.
\newblock \emph{CoRR}, abs/2106.14843, 2021.

\bibitem[Gal et~al.(2022)Gal, Alaluf, Atzmon, Patashnik, Bermano, Chechik, and Cohen-or]{gal2022image}
Rinon Gal, Yuval Alaluf, Yuval Atzmon, Or Patashnik, Amit~Haim Bermano, Gal Chechik, and Daniel Cohen-or.
\newblock An image is worth one word: Personalizing text-to-image generation using textual inversion.
\newblock In \emph{The Eleventh International Conference on Learning Representations}, 2022.

\bibitem[Geyer et~al.(2023)Geyer, Bar-Tal, Bagon, and Dekel]{tokenflow2023}
Michal Geyer, Omer Bar-Tal, Shai Bagon, and Tali Dekel.
\newblock Tokenflow: Consistent diffusion features for consistent video editing.
\newblock \emph{arXiv preprint arxiv:2307.10373}, 2023.

\bibitem[Gleicher(2001)]{Gleicher2001}
Michael Gleicher.
\newblock Motion path editing.
\newblock In \emph{Proceedings of the 2001 Symposium on Interactive 3D Graphics}, page 195–202, New York, NY, USA, 2001. Association for Computing Machinery.

\bibitem[Gombrich(1995)]{gombrich1995story}
Ernst~Hans Gombrich.
\newblock \emph{The story of art}.
\newblock Phaidon London, 1995.

\bibitem[Guay et~al.(2015)Guay, Ronfard, Gleicher, and Cani]{guay2015space}
Martin Guay, R{\'e}mi Ronfard, Michael Gleicher, and Marie-Paule Cani.
\newblock Space-time sketching of character animation.
\newblock \emph{ACM Transactions on Graphics (ToG)}, 34\penalty0 (4):\penalty0 1--10, 2015.

\bibitem[Ha and Eck(2017)]{SketchRNN}
David Ha and Douglas Eck.
\newblock A neural representation of sketch drawings.
\newblock \emph{CoRR}, abs/1704.03477, 2017.

\bibitem[He et~al.(2022)He, Yang, Zhang, Shan, and Chen]{he2022lvdm}
Yingqing He, Tianyu Yang, Yong Zhang, Ying Shan, and Qifeng Chen.
\newblock Latent video diffusion models for high-fidelity long video generation.
\newblock 2022.

\bibitem[Hertzmann(2020)]{Hertzmann2020WhyDL}
Aaron Hertzmann.
\newblock Why do line drawings work? a realism hypothesis.
\newblock \emph{Perception}, 49:\penalty0 439 -- 451, 2020.

\bibitem[Hinz et~al.(2022)Hinz, Fisher, Wang, Shechtman, and Wermter]{hinz2022charactergan}
Tobias Hinz, Matthew Fisher, Oliver Wang, Eli Shechtman, and Stefan Wermter.
\newblock Charactergan: Few-shot keypoint character animation and reposing.
\newblock In \emph{Proceedings of the IEEE/CVF Winter Conference on Applications of Computer Vision}, pages 1988--1997, 2022.

\bibitem[Ho et~al.(2022)Ho, Chan, Saharia, Whang, Gao, Gritsenko, Kingma, Poole, Norouzi, Fleet, et~al.]{ho2022imagen}
Jonathan Ho, William Chan, Chitwan Saharia, Jay Whang, Ruiqi Gao, Alexey Gritsenko, Diederik~P Kingma, Ben Poole, Mohammad Norouzi, David~J Fleet, et~al.
\newblock Imagen video: High definition video generation with diffusion models.
\newblock \emph{arXiv preprint arXiv:2210.02303}, 2022.

\bibitem[Hornung et~al.(2007)Hornung, Dekkers, and Kobbelt]{Hornung2007}
Alexander Hornung, Ellen Dekkers, and Leif Kobbelt.
\newblock Character animation from 2d pictures and 3d motion data.
\newblock \emph{ACM Trans. Graph.}, 26\penalty0 (1):\penalty0 1–es, 2007.

\bibitem[Hu et~al.(2022)Hu, Luo, and Chen]{hu2022make}
Yaosi Hu, Chong Luo, and Zhenzhong Chen.
\newblock Make it move: controllable image-to-video generation with text descriptions.
\newblock In \emph{Proceedings of the IEEE/CVF Conference on Computer Vision and Pattern Recognition}, pages 18219--18228, 2022.

\bibitem[Hu et~al.(2023)Hu, Chen, and Luo]{hu2023lamd}
Yaosi Hu, Zhenzhong Chen, and Chong Luo.
\newblock Lamd: Latent motion diffusion for video generation, 2023.

\bibitem[Igarashi et~al.(1998)Igarashi, Kadobayashi, Mase, and Tanaka]{IgarashiPath98}
Takeo Igarashi, Rieko Kadobayashi, Kenji Mase, and Hidehiko Tanaka.
\newblock Path drawing for 3d walkthrough.
\newblock In \emph{Proceedings of the 11th Annual ACM Symposium on User Interface Software and Technology}, page 173–174, New York, NY, USA, 1998. Association for Computing Machinery.

\bibitem[Igarashi et~al.(2005)Igarashi, Moscovich, and Hughes]{igarashi2005arapshape}
Takeo Igarashi, Tomer Moscovich, and John~F. Hughes.
\newblock As-rigid-as-possible shape manipulation.
\newblock \emph{ACM Trans. Graph.}, 24\penalty0 (3):\penalty0 1134–1141, 2005.

\bibitem[Iluz et~al.(2023)Iluz, Vinker, Hertz, Berio, Cohen-Or, and Shamir]{IluzVinker2023}
Shir Iluz, Yael Vinker, Amir Hertz, Daniel Berio, Daniel Cohen-Or, and Ariel Shamir.
\newblock Word-as-image for semantic typography.
\newblock \emph{ACM Trans. Graph.}, 42\penalty0 (4), 2023.

\bibitem[Jain et~al.(2022)Jain, Xie, and Abbeel]{jain2022vectorfusion}
Ajay Jain, Amber Xie, and Pieter Abbeel.
\newblock Vectorfusion: Text-to-svg by abstracting pixel-based diffusion models.
\newblock \emph{arXiv}, 2022.

\bibitem[Kampelm{\"{u}}hler and Pinz(2020)]{human-like-sketches}
Moritz Kampelm{\"{u}}hler and Axel Pinz.
\newblock Synthesizing human-like sketches from natural images using a conditional convolutional decoder.
\newblock \emph{CoRR}, abs/2003.07101, 2020.

\bibitem[Kazi et~al.(2014)Kazi, Chevalier, Grossman, Zhao, and Fitzmaurice]{DRACOKazi2014}
Rubaiat Kazi, Fanny Chevalier, Tovi Grossman, Shengdong Zhao, and George Fitzmaurice.
\newblock Draco: Bringing life to illustrations with kinetic textures.
\newblock \emph{Conference on Human Factors in Computing Systems - Proceedings}, 2014.

\bibitem[Khachatryan(2020)]{levon2020texanmesh}
Levon Khachatryan.
\newblock Tex-an mesh: Textured and animatable human body mesh reconstruction from a single image.
\newblock https://github.com/lev1khachatryan/Tex-An\_Mesh, 2020.

\bibitem[Khachatryan et~al.(2023)Khachatryan, Movsisyan, Tadevosyan, Henschel, Wang, Navasardyan, and Shi]{khachatryan2023text2video}
Levon Khachatryan, Andranik Movsisyan, Vahram Tadevosyan, Roberto Henschel, Zhangyang Wang, Shant Navasardyan, and Humphrey Shi.
\newblock Text2video-zero: Text-to-image diffusion models are zero-shot video generators.
\newblock \emph{arXiv preprint arXiv:2303.13439}, 2023.

\bibitem[Kim et~al.(2020)Kim, Joo, and Kim]{kim2020tivgan}
Doyeon Kim, Donggyu Joo, and Junmo Kim.
\newblock Tivgan: Text to image to video generation with step-by-step evolutionary generator.
\newblock \emph{IEEE Access}, 8:\penalty0 153113--153122, 2020.

\bibitem[Kingma and Ba(2014)]{kingma2014adam}
Diederik~P Kingma and Jimmy Ba.
\newblock Adam: A method for stochastic optimization.
\newblock \emph{arXiv preprint arXiv:1412.6980}, 2014.

\bibitem[Levi and Gotsman(2013)]{ArtiSketch2013}
Zohar Levi and Craig Gotsman.
\newblock {ArtiSketch: A System for Articulated Sketch Modeling}.
\newblock \emph{Computer Graphics Forum}, 2013.

\bibitem[Li et~al.(2019)Li, Lin, Mech, Yumer, and Ramanan]{li2019photosketching}
Mengtian Li, Zhe Lin, Radomir Mech, Ersin Yumer, and Deva Ramanan.
\newblock Photo-sketching: Inferring contour drawings from images, 2019.

\bibitem[Li et~al.(2020)Li, Luk\'{a}\v{c}, Micha\"{e}l, and Ragan-Kelley]{diffvg}
Tzu-Mao Li, Michal Luk\'{a}\v{c}, Gharbi Micha\"{e}l, and Jonathan Ragan-Kelley.
\newblock Differentiable vector graphics rasterization for editing and learning.
\newblock \emph{ACM Trans. Graph. (Proc. SIGGRAPH Asia)}, 39\penalty0 (6):\penalty0 193:1--193:15, 2020.

\bibitem[Li et~al.(2023)Li, Chu, Wu, Yuan, Liu, Zhang, Li, Feng, Ding, and Wang]{li2023videogen}
Xin Li, Wenqing Chu, Ye Wu, Weihang Yuan, Fanglong Liu, Qi Zhang, Fu Li, Haocheng Feng, Errui Ding, and Jingdong Wang.
\newblock Videogen: A reference-guided latent diffusion approach for high definition text-to-video generation.
\newblock \emph{arXiv preprint arXiv:2309.00398}, 2023.

\bibitem[Li et~al.(2015)Li, Song, Hospedales, and Gong]{Deformable_Stroke}
Yi Li, Yi{-}Zhe Song, Timothy~M. Hospedales, and Shaogang Gong.
\newblock Free-hand sketch synthesis with deformable stroke models.
\newblock \emph{CoRR}, abs/1510.02644, 2015.

\bibitem[Li et~al.(2018)Li, Min, Shen, Carlson, and Carin]{li2018video}
Yitong Li, Martin Min, Dinghan Shen, David Carlson, and Lawrence Carin.
\newblock Video generation from text.
\newblock In \emph{Proceedings of the AAAI conference on artificial intelligence}, 2018.

\bibitem[Lin et~al.(2020)Lin, Fu, Jiang, and Xue]{Lin2020SketchBERTLS}
Hangyu Lin, Yanwei Fu, Yu-Gang Jiang, and X. Xue.
\newblock Sketch-bert: Learning sketch bidirectional encoder representation from transformers by self-supervised learning of sketch gestalt.
\newblock \emph{2020 IEEE/CVF Conference on Computer Vision and Pattern Recognition (CVPR)}, pages 6757--6766, 2020.

\bibitem[Liu et~al.(2021)Liu, Fisher, Hertzmann, and Kalogerakis]{liu2021neural}
Difan Liu, Matthew Fisher, Aaron Hertzmann, and Evangelos Kalogerakis.
\newblock Neural strokes: Stylized line drawing of 3d shapes.
\newblock In \emph{Proceedings of the IEEE/CVF International Conference on Computer Vision}, pages 14204--14213, 2021.

\bibitem[Luo et~al.(2023)Luo, Chen, Zhang, Huang, Wang, Shen, Zhao, Zhou, and Tan]{videofusion2023}
Zhengxiong Luo, Dayou Chen, Yingya Zhang, Yan Huang, Liang Wang, Yujun Shen, Deli Zhao, Jingren Zhou, and Tieniu Tan.
\newblock Videofusion: Decomposed diffusion models for high-quality video generation.
\newblock In \emph{Proceedings of the IEEE/CVF Conference on Computer Vision and Pattern Recognition}, 2023.

\bibitem[Metzer et~al.(2022)Metzer, Richardson, Patashnik, Giryes, and Cohen-Or]{metzer2022latent}
Gal Metzer, Elad Richardson, Or Patashnik, Raja Giryes, and Daniel Cohen-Or.
\newblock Latent-nerf for shape-guided generation of 3d shapes and textures.
\newblock \emph{arXiv preprint arXiv:2211.07600}, 2022.

\bibitem[Mihai and Hare(2021)]{mihai2021learning}
Daniela Mihai and Jonathon Hare.
\newblock Learning to draw: Emergent communication through sketching.
\newblock \emph{Advances in Neural Information Processing Systems}, 34:\penalty0 7153--7166, 2021.

\bibitem[Mildenhall et~al.(2020)Mildenhall, Srinivasan, Tancik, Barron, Ramamoorthi, and Ng]{mildenhall2020nerf}
Ben Mildenhall, Pratul~P Srinivasan, Matthew Tancik, Jonathan~T Barron, Ravi Ramamoorthi, and Ren Ng.
\newblock Nerf: Representing scenes as neural radiance fields for view synthesis.
\newblock In \emph{European Conference on Computer Vision}, pages 405--421. Springer, 2020.

\bibitem[Mildenhall et~al.(2021)Mildenhall, Srinivasan, Tancik, Barron, Ramamoorthi, and Ng]{mildenhall2021nerf}
Ben Mildenhall, Pratul~P Srinivasan, Matthew Tancik, Jonathan~T Barron, Ravi Ramamoorthi, and Ren Ng.
\newblock Nerf: Representing scenes as neural radiance fields for view synthesis.
\newblock \emph{Communications of the ACM}, 65\penalty0 (1):\penalty0 99--106, 2021.

\bibitem[Min et~al.(2009)Min, Chen, and Chai]{Min2009}
Jianyuan Min, Yen-Lin Chen, and Jinxiang Chai.
\newblock Interactive generation of human animation with deformable motion models.
\newblock \emph{ACM Trans. Graph.}, 29\penalty0 (1), 2009.

\bibitem[Mo et~al.(2021)Mo, Simo-Serra, Gao, Zou, and Wang]{mo2021virtualsketching}
Haoran Mo, Edgar Simo-Serra, Chengying Gao, Changqing Zou, and Ruomei Wang.
\newblock General virtual sketching framework for vector line art.
\newblock \emph{ACM Transactions on Graphics (Proceedings of ACM SIGGRAPH 2021)}, 40\penalty0 (4):\penalty0 51:1--51:14, 2021.

\bibitem[Muhammad et~al.(2018)Muhammad, Yang, Song, Xiang, and Hospedales]{Deep-Sketch-Abstraction}
Umar~Riaz Muhammad, Yongxin Yang, Yi{-}Zhe Song, Tao Xiang, and Timothy~M. Hospedales.
\newblock Learning deep sketch abstraction.
\newblock \emph{CoRR}, abs/1804.04804, 2018.

\bibitem[Ni et~al.(2022)Ni, Peng, Chen, Zhang, Meng, Fu, Xiang, and Ling]{XCLIP}
Bolin Ni, Houwen Peng, Minghao Chen, Songyang Zhang, Gaofeng Meng, Jianlong Fu, Shiming Xiang, and Haibin Ling.
\newblock Expanding language-image pretrained models for general video recognition.
\newblock 2022.

\bibitem[Ni et~al.(2023)Ni, Shi, Li, Huang, and Min]{ni2023conditional}
Haomiao Ni, Changhao Shi, Kai Li, Sharon~X. Huang, and Martin~Renqiang Min.
\newblock Conditional image-to-video generation with latent flow diffusion models, 2023.

\bibitem[\"{O}ztireli et~al.(2013)\"{O}ztireli, Baran, Popa, Dalstein, Sumner, and Gross]{Oztireli13Differential}
A.~Cengiz \"{O}ztireli, Ilya Baran, Tiberiu Popa, Boris Dalstein, Robert~W. Sumner, and Markus Gross.
\newblock Differential blending for expressive sketch-based posing.
\newblock In \emph{Proceedings of the 2013 ACM SIGGRAPH/Eurographics Symposium on Computer Animation}, New York, NY, USA, 2013. ACM.

\bibitem[Pan and Zhang(2011)]{Pan2011}
Junjun Pan and Jian~J. Zhang.
\newblock \emph{Sketch-Based Skeleton-Driven 2D Animation and Motion Capture}, pages 164--181.
\newblock Springer Berlin Heidelberg, Berlin, Heidelberg, 2011.

\bibitem[Pan et~al.(2017)Pan, Qiu, Yao, Li, and Mei]{pan2017create}
Yingwei Pan, Zhaofan Qiu, Ting Yao, Houqiang Li, and Tao Mei.
\newblock To create what you tell: Generating videos from captions.
\newblock In \emph{Proceedings of the 25th ACM international conference on Multimedia}, pages 1789--1798, 2017.

\bibitem[Poole et~al.(2022)Poole, Jain, Barron, and Mildenhall]{poole2022dreamfusion}
Ben Poole, Ajay Jain, Jonathan~T Barron, and Ben Mildenhall.
\newblock Dreamfusion: Text-to-3d using 2d diffusion.
\newblock In \emph{The Eleventh International Conference on Learning Representations}, 2022.

\bibitem[Poursaeed et~al.(2020)Poursaeed, Kim, Shechtman, Saito, and Belongie]{poursaeed2020neural}
Omid Poursaeed, Vladimir Kim, Eli Shechtman, Jun Saito, and Serge Belongie.
\newblock Neural puppet: Generative layered cartoon characters.
\newblock In \emph{Proceedings of the IEEE/CVF Winter Conference on Applications of Computer Vision}, pages 3346--3356, 2020.

\bibitem[Radford et~al.(2021)Radford, Kim, Hallacy, Ramesh, Goh, Agarwal, Sastry, Askell, Mishkin, Clark, et~al.]{radford2021learning}
Alec Radford, Jong~Wook Kim, Chris Hallacy, Aditya Ramesh, Gabriel Goh, Sandhini Agarwal, Girish Sastry, Amanda Askell, Pamela Mishkin, Jack Clark, et~al.
\newblock Learning transferable visual models from natural language supervision.
\newblock In \emph{International conference on machine learning}, pages 8748--8763. PMLR, 2021.

\bibitem[Ribeiro et~al.(2020)Ribeiro, Bui, Collomosse, and Ponti]{Ribeiro2020SketchformerTR}
Leo Sampaio~Ferraz Ribeiro, Tu Bui, John~P. Collomosse, and Moacir~Antonelli Ponti.
\newblock Sketchformer: Transformer-based representation for sketched structure.
\newblock \emph{2020 IEEE/CVF Conference on Computer Vision and Pattern Recognition (CVPR)}, pages 14141--14150, 2020.

\bibitem[Runway(2023)]{gen2runway}
Runway.
\newblock Gen-2: Text driven video generation.
\newblock https://research.runwayml.com/gen2, 2023.

\bibitem[Singer et~al.(2022)Singer, Polyak, Hayes, Yin, An, Zhang, Hu, Yang, Ashual, Gafni, et~al.]{singer2022make}
Uriel Singer, Adam Polyak, Thomas Hayes, Xi Yin, Jie An, Songyang Zhang, Qiyuan Hu, Harry Yang, Oron Ashual, Oran Gafni, et~al.
\newblock Make-a-video: Text-to-video generation without text-video data.
\newblock \emph{arXiv preprint arXiv:2209.14792}, 2022.

\bibitem[Smith et~al.(2023)Smith, Zheng, Li, Jain, and Hodgins]{smith2023method}
Harrison~Jesse Smith, Qingyuan Zheng, Yifei Li, Somya Jain, and Jessica~K Hodgins.
\newblock A method for animating children’s drawings of the human figure.
\newblock \emph{ACM Transactions on Graphics}, 42\penalty0 (3):\penalty0 1--15, 2023.

\bibitem[Song et~al.(2018)Song, Pang, Song, Xiang, and Hospedales]{song2018learning}
Jifei Song, Kaiyue Pang, Yi-Zhe Song, Tao Xiang, and Timothy Hospedales.
\newblock Learning to sketch with shortcut cycle consistency, 2018.

\bibitem[Su et~al.(2018)Su, Bai, Fu, Tai, and Wang]{su2018live}
Qingkun Su, Xue Bai, Hongbo Fu, Chiew-Lan Tai, and Jue Wang.
\newblock Live sketch: Video-driven dynamic deformation of static drawings.
\newblock In \emph{Proceedings of the 2018 CHI Conference on Human Factors in Computing Systems}, pages 1--12, 2018.

\bibitem[Tang et~al.(2023)Tang, Yang, Zhu, Zeng, and Bansal]{tang2023anytoany}
Zineng Tang, Ziyi Yang, Chenguang Zhu, Michael Zeng, and Mohit Bansal.
\newblock Any-to-any generation via composable diffusion.
\newblock 2023.

\bibitem[Thorne et~al.(2004)Thorne, Burke, and Van De~Panne]{thorne2004motion}
Matthew Thorne, David Burke, and Michiel Van De~Panne.
\newblock Motion doodles: an interface for sketching character motion.
\newblock \emph{ACM Transactions on Graphics (ToG)}, 23\penalty0 (3):\penalty0 424--431, 2004.

\bibitem[Tian et~al.(2021)Tian, Ren, Chai, Olszewski, Peng, Metaxas, and Tulyakov]{tian2021a}
Yu Tian, Jian Ren, Menglei Chai, Kyle Olszewski, Xi Peng, Dimitris~N. Metaxas, and Sergey Tulyakov.
\newblock A good image generator is what you need for high-resolution video synthesis.
\newblock In \emph{International Conference on Learning Representations}, 2021.

\bibitem[Vinker et~al.(2022{\natexlab{a}})Vinker, Alaluf, Cohen-Or, and Shamir]{clipascene}
Yael Vinker, Yuval Alaluf, Daniel Cohen-Or, and Ariel Shamir.
\newblock Clipascene: Scene sketching with different types and levels of abstraction.
\newblock 2022{\natexlab{a}}.

\bibitem[Vinker et~al.(2022{\natexlab{b}})Vinker, Pajouheshgar, Bo, Bachmann, Bermano, Cohen-Or, Zamir, and Shamir]{vinker2022clipasso}
Yael Vinker, Ehsan Pajouheshgar, Jessica~Y. Bo, Roman~Christian Bachmann, Amit~Haim Bermano, Daniel Cohen-Or, Amir Zamir, and Ariel Shamir.
\newblock Clipasso: Semantically-aware object sketching.
\newblock \emph{ACM Trans. Graph.}, 41\penalty0 (4), 2022{\natexlab{b}}.

\bibitem[Wang et~al.(2004)Wang, Xu, Shum, and Cohen]{WangVideoTooning2004}
Jue Wang, Yingqing Xu, Heung-Yeung Shum, and Michael~F. Cohen.
\newblock Video tooning.
\newblock In \emph{ACM SIGGRAPH 2004 Papers}, page 574–583, New York, NY, USA, 2004. Association for Computing Machinery.

\bibitem[Wang et~al.(2023{\natexlab{a}})Wang, Yuan, Chen, Zhang, Wang, and Zhang]{wang2023modelscope}
Jiuniu Wang, Hangjie Yuan, Dayou Chen, Yingya Zhang, Xiang Wang, and Shiwei Zhang.
\newblock Modelscope text-to-video technical report.
\newblock \emph{arXiv preprint arXiv:2308.06571}, 2023{\natexlab{a}}.

\bibitem[Wang et~al.(2023{\natexlab{b}})Wang, Yuan, Zhang, Chen, Wang, Zhang, Shen, Zhao, and Zhou]{videocomposer2023}
Xiang* Wang, Hangjie* Yuan, Shiwei* Zhang, Dayou* Chen, Jiuniu Wang, Yingya Zhang, Yujun Shen, Deli Zhao, and Jingren Zhou.
\newblock Videocomposer: Compositional video synthesis with motion controllability.
\newblock \emph{arXiv preprint arXiv:2306.02018}, 2023{\natexlab{b}}.

\bibitem[Wang et~al.(2023{\natexlab{c}})Wang, Chen, Ma, Zhou, Huang, Wang, Yang, He, Yu, Yang, et~al.]{wang2023lavie}
Yaohui Wang, Xinyuan Chen, Xin Ma, Shangchen Zhou, Ziqi Huang, Yi Wang, Ceyuan Yang, Yinan He, Jiashuo Yu, Peiqing Yang, et~al.
\newblock Lavie: High-quality video generation with cascaded latent diffusion models.
\newblock \emph{arXiv preprint arXiv:2309.15103}, 2023{\natexlab{c}}.

\bibitem[Weissenborn et~al.(2019)Weissenborn, T{\"a}ckstr{\"o}m, and Uszkoreit]{weissenborn2019scaling}
Dirk Weissenborn, Oscar T{\"a}ckstr{\"o}m, and Jakob Uszkoreit.
\newblock Scaling autoregressive video models.
\newblock \emph{arXiv preprint arXiv:1906.02634}, 2019.

\bibitem[Weng et~al.(2018)Weng, Curless, and Kemelmacher-Shlizerman]{Weng2018PhotoW3}
Chung-Yi Weng, Brian Curless, and Ira Kemelmacher-Shlizerman.
\newblock Photo wake-up: 3d character animation from a single photo.
\newblock \emph{2019 IEEE/CVF Conference on Computer Vision and Pattern Recognition (CVPR)}, pages 5901--5910, 2018.

\bibitem[Weng et~al.(2019)Weng, Curless, and Kemelmacher-Shlizerman]{weng2019photo}
Chung-Yi Weng, Brian Curless, and Ira Kemelmacher-Shlizerman.
\newblock Photo wake-up: 3d character animation from a single photo.
\newblock In \emph{Proceedings of the IEEE/CVF conference on computer vision and pattern recognition}, pages 5908--5917, 2019.

\bibitem[Wu et~al.(2021)Wu, Huang, Zhang, Li, Ji, Yang, Sapiro, and Duan]{wu2021godiva}
Chenfei Wu, Lun Huang, Qianxi Zhang, Binyang Li, Lei Ji, Fan Yang, Guillermo Sapiro, and Nan Duan.
\newblock Godiva: Generating open-domain videos from natural descriptions.
\newblock \emph{arXiv preprint arXiv:2104.14806}, 2021.

\bibitem[Wu et~al.(2022)Wu, Liang, Ji, Yang, Fang, Jiang, and Duan]{wu2022nuwa}
Chenfei Wu, Jian Liang, Lei Ji, Fan Yang, Yuejian Fang, Daxin Jiang, and Nan Duan.
\newblock N{\"u}wa: Visual synthesis pre-training for neural visual world creation.
\newblock In \emph{European conference on computer vision}, pages 720--736. Springer, 2022.

\bibitem[Xie and Tu(2015)]{xie2015holistically}
Saining Xie and Zhuowen Tu.
\newblock Holistically-nested edge detection.
\newblock In \emph{Proceedings of the IEEE international conference on computer vision}, pages 1395--1403, 2015.

\bibitem[Xing et~al.(2015)Xing, Wei, Shiratori, and Yatani]{Xing2015}
Jun Xing, Li-Yi Wei, Takaaki Shiratori, and Koji Yatani.
\newblock Autocomplete hand-drawn animations.
\newblock \emph{ACM Trans. Graph.}, 34\penalty0 (6), 2015.

\bibitem[Xing et~al.(2016)Xing, Kazi, Grossman, Wei, Stam, and Fitzmaurice]{XingEnergyBrushes2016}
Jun Xing, Rubaiat Kazi, Tovi Grossman, Li-Yi Wei, Jos Stam, and George Fitzmaurice.
\newblock Energy-brushes: Interactive tools for illustrating stylized elemental dynamics.
\newblock pages 755--766, 2016.

\bibitem[Xing et~al.(2023)Xing, Xia, Zhang, Chen, Wang, Wong, and Shan]{xing2023dynamicrafter}
Jinbo Xing, Menghan Xia, Yong Zhang, Haoxin Chen, Xintao Wang, Tien-Tsin Wong, and Ying Shan.
\newblock Dynamicrafter: Animating open-domain images with video diffusion priors.
\newblock \emph{arXiv preprint arXiv:2310.12190}, 2023.

\bibitem[Xu et~al.(2020)Xu, Hospedales, Yin, Song, Xiang, and Wang]{xu2020deep}
Peng Xu, Timothy~M. Hospedales, Qiyue Yin, Yi-Zhe Song, Tao Xiang, and Liang Wang.
\newblock Deep learning for free-hand sketch: A survey and a toolbox, 2020.

\bibitem[Yan et~al.(2021)Yan, Zhang, Abbeel, and Srinivas]{yan2021videogpt}
Wilson Yan, Yunzhi Zhang, Pieter Abbeel, and Aravind Srinivas.
\newblock Videogpt: Video generation using vq-vae and transformers.
\newblock \emph{arXiv preprint arXiv:2104.10157}, 2021.

\bibitem[Yi et~al.(2020)Yi, Liu, Lai, and Rosin]{yi2020unpaired}
Ran Yi, Yong-Jin Liu, Yu-Kun Lai, and Paul~L Rosin.
\newblock Unpaired portrait drawing generation via asymmetric cycle mapping.
\newblock In \emph{Proceedings of the IEEE/CVF Conference on Computer Vision and Pattern Recognition}, pages 8217--8225, 2020.

\bibitem[Yu et~al.(2022)Yu, Wang, Vasudevan, Yeung, Seyedhosseini, and Wu]{Yu2022CoCaCC}
Jiahui Yu, Zirui Wang, Vijay Vasudevan, Legg Yeung, Mojtaba Seyedhosseini, and Yonghui Wu.
\newblock Coca: Contrastive captioners are image-text foundation models.
\newblock 2022.

\bibitem[Zhou et~al.(2023)Zhou, Wang, Yan, Lv, Zhu, and Feng]{zhou2023magicvideo}
Daquan Zhou, Weimin Wang, Hanshu Yan, Weiwei Lv, Yizhe Zhu, and Jiashi Feng.
\newblock Magicvideo: Efficient video generation with latent diffusion models, 2023.

\bibitem[Zhu et~al.(2023)Zhu, Ma, Chen, and Yuan]{zhu2023motionvideogan}
Jingyuan Zhu, Huimin Ma, Jiansheng Chen, and Jian Yuan.
\newblock Motionvideogan: A novel video generator based on the motion space learned from image pairs.
\newblock \emph{IEEE Transactions on Multimedia}, 2023.

\end{thebibliography}
}

\clearpage
\appendix
\setcounter{page}{1}

\newpage
\twocolumn[
\centering
\Large
\textbf{\thetitle}\\
\vspace{2em}Supplementary Material \\
\vspace{1.0em}
] %

\part{}
\vspace{-20pt}
\parttoc

\section{Additional results and videos}
All videos and a large number of additional results are available in our \href{https://livesketch.github.io/}{supplementary website}. These include an array of subjects animated with our method, along with additional comparisons, ablation experiments and visualizations of limitations. Please note that all comparisons and ablation baseline results use our default parameters, while the large \href{https://livesketch.github.io/#gallery}{video gallery} includes results with different parameter settings, chosen according to our aesthetic preferences.

\section{Analysis and ablation}

In this section we present an array of experiments that explore the sensitivity of our method to different hyperparameters of the approach. These include technical changes (such as learning rate adjustments), but also conceptual explorations such as the effect of sketch abstraction on the generated videos.

\subsection{Text prompt effect}
\begin{figure}
    \centering
    \includegraphics[width=1\linewidth]{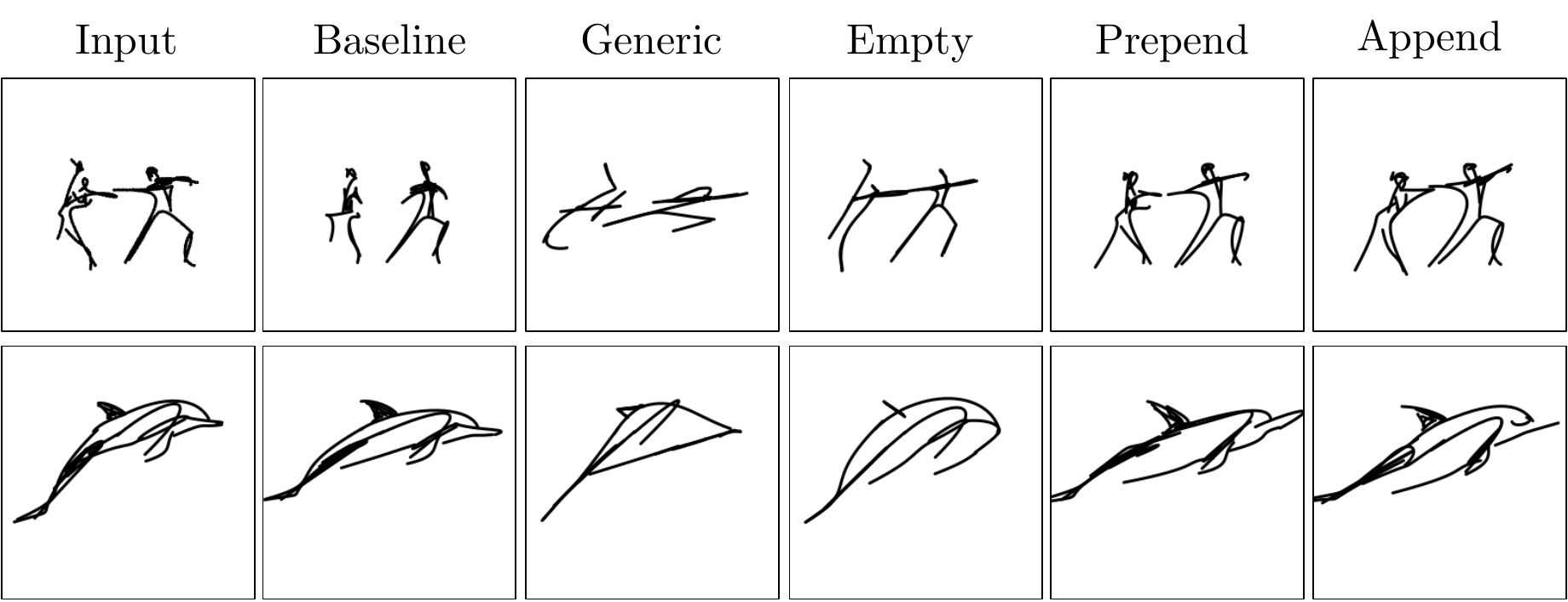}
    \caption{Text prompt effect. We investigate the effects of using a generic prompt (\ap{The object is moving}) for all sketches, the effect of using an empty prompt, or prepending and appending strings that compel the diffusion model go generate sketches. Additional video results are shown in the website.}
    \label{fig:text_prompt_effect}
\end{figure}

Our animation process is guided by a user-provided text, based on the prior of a pretrained text-to-video model.
This section further examines how the specified prompt affects the animation.
We first verify that the text itself influences the results in a meaningful way. To do so, we apply our method to several example sketches, using two alternatives:  A \ap{generic} prompt (\ap{the object is moving}), and the empty prompt (``"). 
The results are shown in \cref{fig:text_prompt_effect} and in the \href{https://livesketch.github.io/#text_prompt_effect}{\ap{Text Prompt Effect} section} of the website.
Using the generic prompt leads to irrelevant animations in which both the motion and the sketch appearance exhibit significant artifacts. Using an empty prompt leads to results with no visible motion, and large shape deviations. 
We can thus conclude that using prompts tailored for the input sketch is crucial, both to preserve its characteristics and for the ability to generate meaningful motion. 

We further examine the impact of modifying the prompt in a way that would motivate the text-to-video to create a sketch. Specifically, we either prepend the string \ap{A sketch of} or append the string \ap{Abstract sketch. Line drawing} to the prompts. 

In general, explicitly prompting for a sketch works comparably well to the original prompts. In some cases we observe slight differences in the extent of the motion or in the adherence to slight details in the input sketch (\eg the penguin's left fin is filled out when using the sketch prompts). However, these can likely be accounted for with learning rate tuning. We thus conclude that the model can reasonably infer the semantics of the object even when the prompt does not directly convey its sketch-based nature.

Finally, we show additional results for applying different prompts to the same input sketch (see \href{https://livesketch.github.io/#varying}{\ap{Varying the Prompt}} in the provided website).
For example, observe how the boxer changes his motion in accordance with the texts provided, demonstrating the actions of jumping, running, and punching. Similarly, a cat can be made to change its pose, or walk towards the camera. 
However, in some cases the method is not sensitive enough to the changes in the provided text prompt. This is particularly apparent when the prompt requests large changes in the shape of the subject, or when the diffusion model struggles to generate the described motion even in it's basic text-to-video setup.
In the video website, we demonstrate this on the ballerina sketch, where the specifics of the prompts are largely ignored, leading to similar dancing motions. However, notice that supplying the base diffusion model with those same prompts, also creates videos with dancing that is unrelated to the motion described in the prompt. We hope that this limitation could be overcome as better, more expressive text-to-video models become available.

\subsection{Different levels of abstraction}
We also demonstrate the effect of altering the abstraction level of the input sketches. We show results for three objects with three levels of abstraction. The sketches were generated using 16, 8, and 4 strokes. An example is provided in \cref{fig:abstraction_levels}, and more examples and the full videos are provided in the supplementary website's \href{https://livesketch.github.io/#abstraction}{\ap{Abstraction Level} section}. 
As can be seen, even for the extreme case of very abstract sketches with only four strokes, our method still manages to produce animations that fit the given prompt.
Yet, the abstract animations may appear less smooth, leaving room for future work to tackle such challenging cases.  

\begin{figure}
    \centering
    \includegraphics[width=1\linewidth]{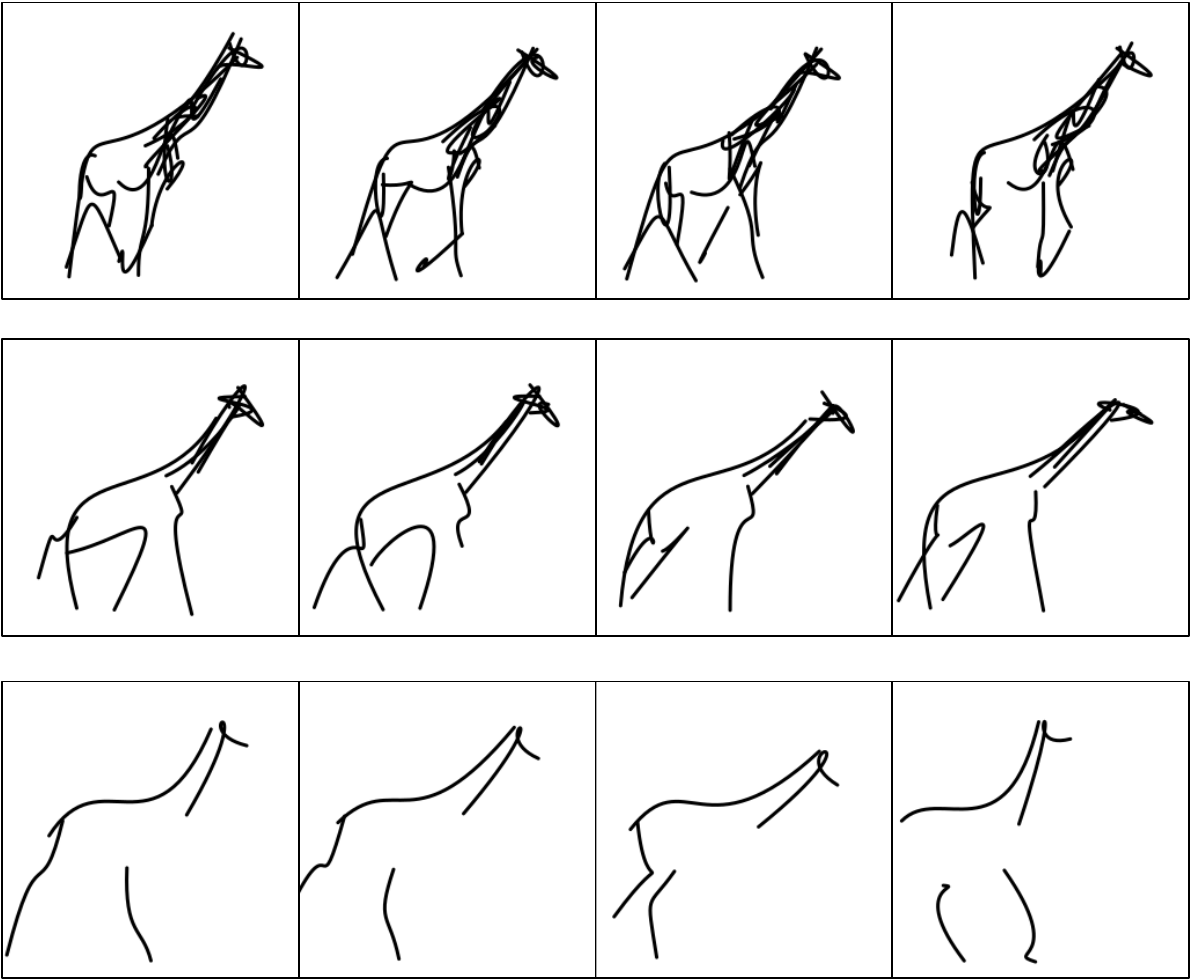}
    \caption{Different levels of abstraction. We show four selected frames for each level of abstraction. The model can successfully synthesize movement even for very abstract representations. }
    \label{fig:abstraction_levels}
\end{figure}

\subsection{Sketch representation}
\begin{figure}
    \centering
    \includegraphics[width=1\linewidth]{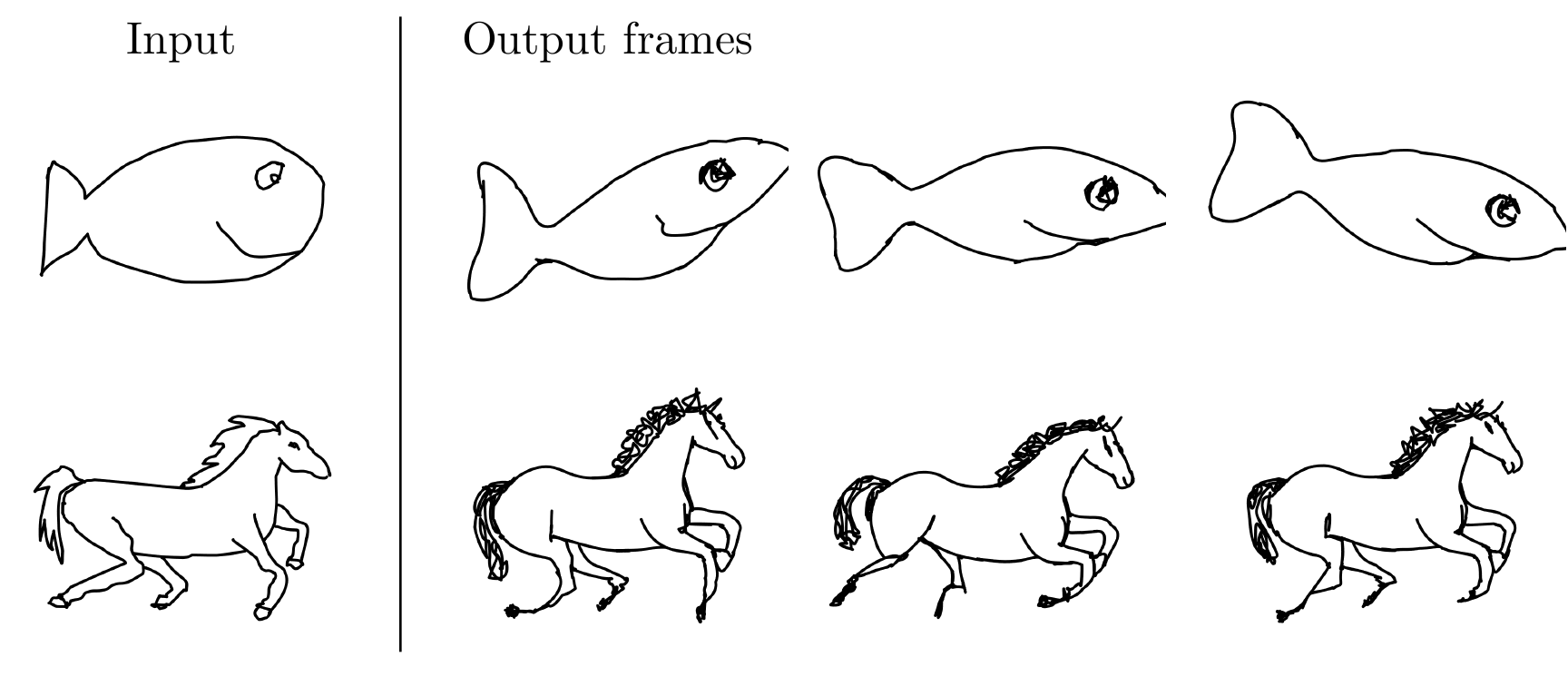}
    \caption{Human-drawn sketches. We applied our method to sketches from the TU-Berlin dataset. With our default parameters, these create reasonable motion but fail to preserve the exact sketch appearance. By tuning the parameters for this input style, shape preservation can be improved. See the website for examples.}
    \label{fig:sketch_rep}
\end{figure}

As described in the main paper, we represent a sketch as a set of black cubic Bezier curves, and use CLIPasso \cite{vinker2022clipasso} to automatically generate the sketches shown in the paper.
However, our approach can be applied to alternative sketch representations.
As highlighted in the limitations section of the main paper, employing different sketch representations may require additional hyperparameter tuning. To illustrate the impact of changing the sketch representation, we applied our method to sketches from the TU-Berlin sketch dataset \cite{eitz2012hdhso}, a human-drawn class-based sketch dataset.
We showcase the results of four representative sketches. Our method was directly applied to the provided SVG files.
\cref{fig:sketch_rep} shows a few representative frames from the videos produced for two sketches. More results are shown in the \href{https://livesketch.github.io/#sketch_representation}{supplementary website}. As can be seen, our method successfully animated the sketches, however their appearance is not fully preserved when using the default hyperparameters. This can be improved by using lower learning rates for the local path.

\subsection{Learning rate scaling and tradeoffs}
As discussed in the main paper, there exists a trade-off between the quality of generated motion and the capacity to retain the appearance of the initial sketch. To illustrate this trade-off, we conducted an experiment wherein we randomly selected three sketches from each class in our evaluation set (9 sketches in total). We then tested the impact of scaling the local learning rate within the range of $0.01$ to $0.0001$, keeping all parameters constant except for the local learning rate.
Qualitative results are shown in the website, under the \href{https://livesketch.github.io/#trade-off}{"Trade-off" section}.
Observe that as we move from the left ($0.0001$) to right ($0.1$), the motion in the animations increases, better aligning with the text prompt. However, this comes at the cost of preserving the original sketch's appearance. For example, observe how the fish and the crab undergo complete transformations when using a learning rate greater than $0.001$. This trade-off introduces additional control for the user, who may prioritize stronger motion over sketch fidelity.

\begin{figure}
    \centering
    \includegraphics[width=\linewidth]{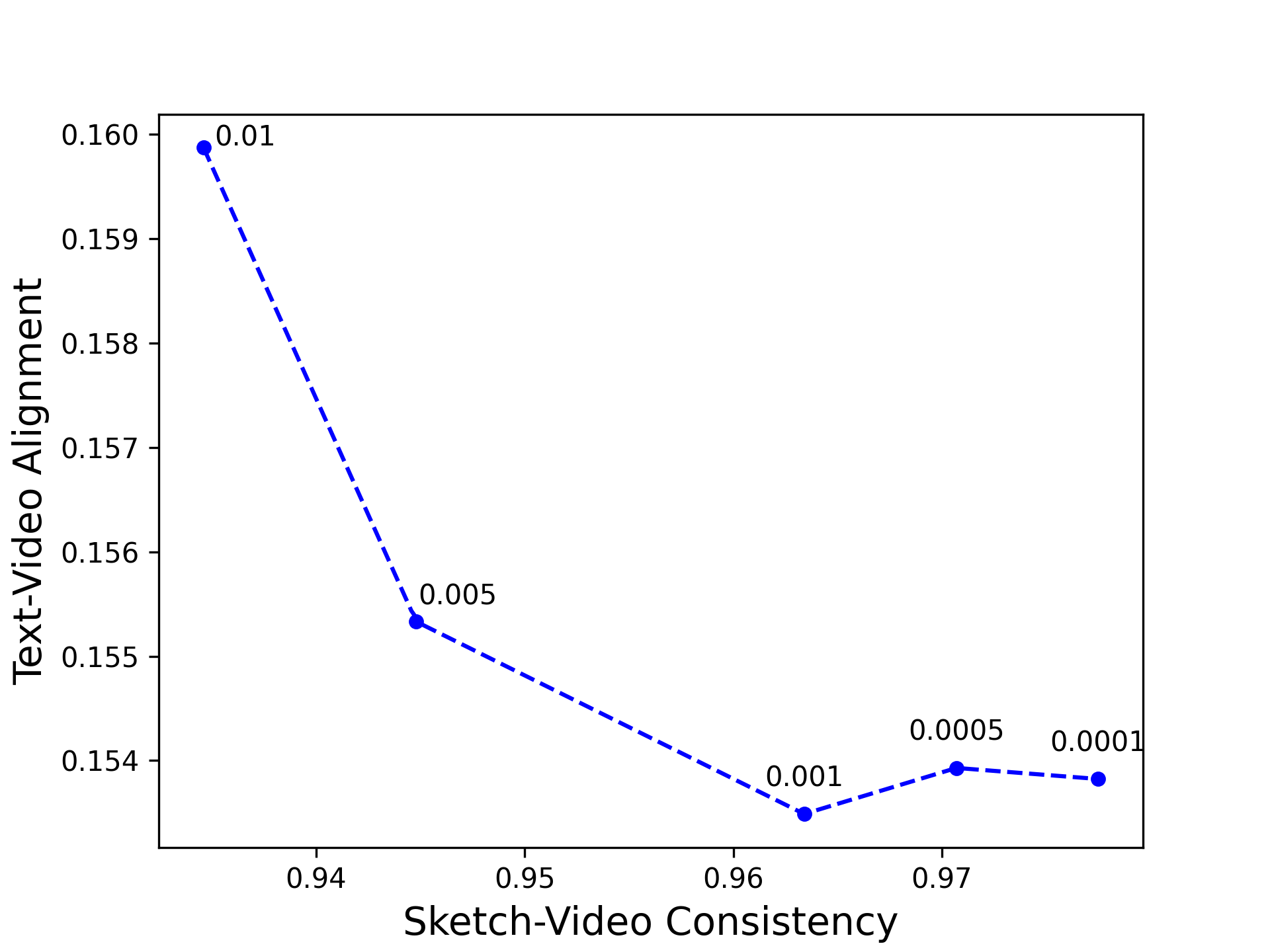}
    \caption{Investigation of the tradeoff between motion quality and sketch preservation. Increasing the local learning rates trades one aspect for another.}
    \label{fig:lr_tradeoff}
\end{figure}

Furthermore, we assess the results using CLIP-based metrics (\cref{fig:lr_tradeoff}). As can be observed, increasing the learning rate leads to a smooth tradeoff between motion quality and sketch preservation. Working with learning rates $\in [0.001, 0.005]$ generally leads to a good compromise between the two aspects - though a user can choose a different working point according to their preferences.

\subsection{Hyperparameter effects}
We demonstrate how changing different hyperparameters in our method can provide the user with additional control (see \href{https://livesketch.github.io/#hyperparameters}{\ap{Hyperparameter Effects}} in the website).
We observe different effects across various sketches, which may be attributed to the video model's prior or the initial sketch quality. 
Specifically, in the third column ("+lr local"), we showcase the impact of increasing the learning rate of the local path. As evident, in some cases (biking and butterfly), this improved the generated motion without significantly harming the sketch's appearance. However, in other cases (cobra and boat), increasing the local path's learning rate leads to a complete alteration of the original sketch. In the fourth and fifth columns we show the effect of increasing the translation and scale prediction weights. As observed, this indeed causes the objects to move more across the frame or change their scale.

\begin{figure}
    \centering
    \includegraphics[width=1\linewidth]{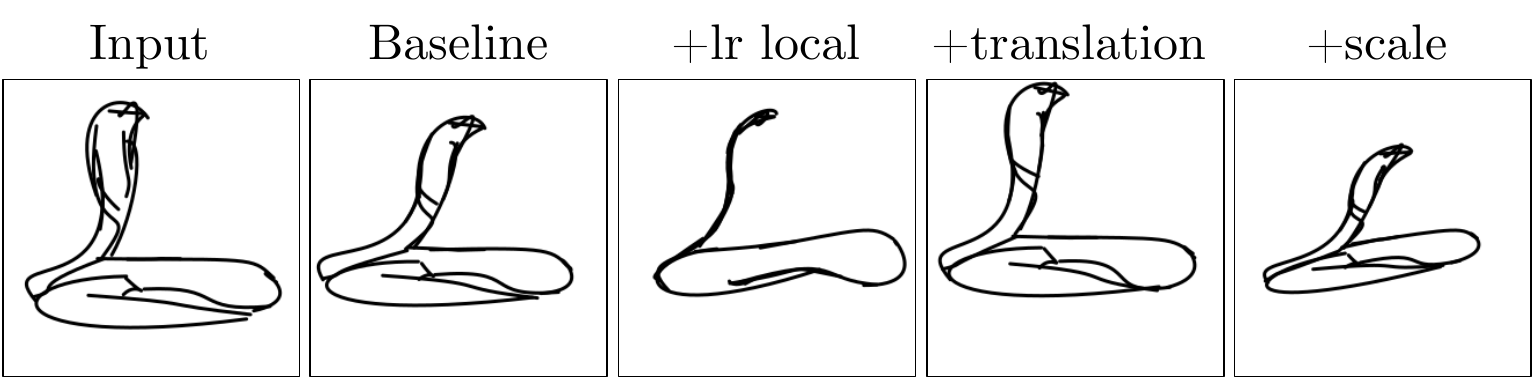}
    \caption{Hyperparameter effect. We show one representative frame from each video (the full videos and additional examples are provided in the website).}
    \label{fig:hyperparameter}
\end{figure}

\begin{figure*}
    \centering
    \includegraphics[width=1\linewidth]{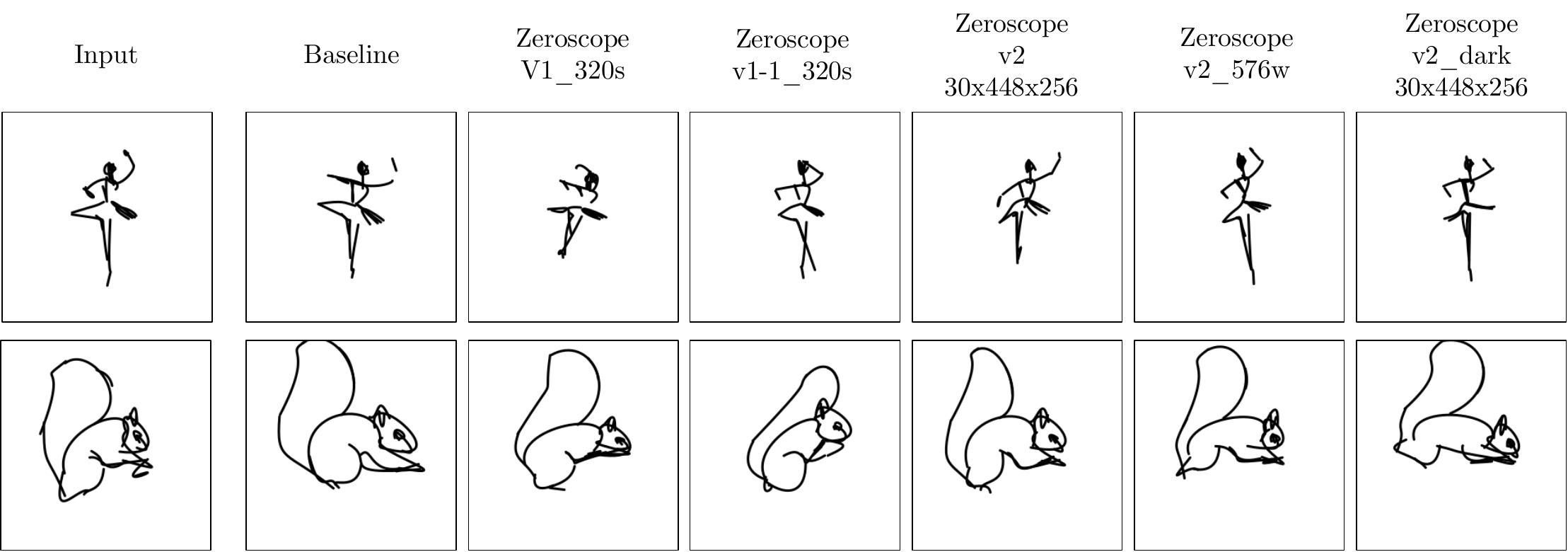}
    \caption{Other text-to-video backbones. We show the first frame from the results of five alternative text-to-video models. The full videos and additional examples are provided in the website. Observe that the choice of backbone model affects the output video in terms of both the sketch's appearance and the type of generated motion.}
    \label{fig:vidmodels}
\end{figure*}

\subsection{Other text-to-video backbones}
We investigate the performance of the model when we swap one text-to-video prior for another.
In the main paper, we use ModelScope~\cite{wang2023modelscope} as our text-to-video diffusion backbone. Here, we qualitatively evaluate the effect of replacing it with other text-to-video models. In particular, we look at a set of ZeroScope models, tuned across a range of resolutions and framerates. The results are shown in the supplementary videos (website section \href{https://livesketch.github.io/#vidmodel}{\ap{Comparing Video Models}}). Two representative examples are provided in \cref{fig:vidmodels}. Our method generalizes to these models with no additional changes. However, note that different models do lead to different motion patterns, and some of them may result in different tradeoffs between the level of motion and the ability to preserve the sketch. For example, observe the cat (second row) which either wags their tail, raises its front legs, or does both, depending on the model. For some models (\eg zeroscope v1-1 320s) the cat appears more deformed, and a user may prefer to use another working point on the local learning-rate axis in order to restore the shape.

\section{Implementation and technical details}
Here we outline additional details required to reproduce our work and experiments.
We will release all code and image sets used for evaluations to facilitate further research and comparisons.

\subsection{Sketch generation}
Unless otherwise noted, all sketches presented in the main paper and the supplementary material were generated using CLIPasso \cite{vinker2022clipasso}. CLIPasso is a method for automatically generating object sketches represented with cubic Bezier curves. In the majority of examples, we applied CLIPasso with the default settings, using 16 strokes. The sketch's canvas size is $256\times256$, and the strokes width is $1.5$. 
It is important to note that our method can be employed with vector sketches created through alternative approaches, such as \cite{clipascene, CLIPDraw, SketchRNN, jain2022vectorfusion}, or even sketched by hand. For optimal performances, we recommend to represent the input sketch with cubic Bezier curves.

\subsection{Additional training details}
To improve stability in early training steps, we initialize $\mathcal{M}$ so that the predicted local displacements are small and the global transformations~$\gtransform^j$ are close to the identity matrix.

When sampling timesteps for the SDS loss, we follow DreamFusion~\cite{poole2022dreamfusion} and avoid sampling very early or very late steps. In practice we sample the steps uniformly in the range $[50, 950]$.

When rendering the video frames for training we use a canvas size of $256\times256$, even when using text-to-video models trained with different aspect ratios. This limitation is primarily due to memory constraints. Lifting this restriction may aid in improving visual fidelity at the cost of higher VRAM requirements. We similarly restrict ourselves to 24 frames. Increasing this value can improve smoothness at the cost of additional memory. Our baseline method requires roughly 23GB of VRAM.

\subsection{Evaluation details}

\subsubsection{Baseline implementations}

When comparing to alternative methods, we used the following implementations: 

\begin{itemize}
    \item ModelScope: \url{https://huggingface.co/spaces/damo-vilab/MS-Image2Video-demo/tree/main}
    \item ZeroScope: \url{https://huggingface.co/spaces/fffiloni/zeroscope-img-to-video/tree/main}
    \item VideoCrafter: \url{https://huggingface.co/spaces/VideoCrafter/VideoCrafter/tree/main}
    \item Animated Drawings: \url{https://sketch.metademolab.com/canvas}
    \item Gen-2: \url{https://research.runwayml.com/gen2}
\end{itemize}

Note that Gen-2 is actively updated. We obtained our results on October 19th, 2023. 

\subsubsection{Evaluation metrics} 
For our sketch-to-video consistency metric we use OpenAI's CLIP ViT-B/32. For the text-to-video alignment metric we use Microsoft's xclip-large-patch14. This X-CLIP model expects 8 input frames, which are sampled uniformly from the generated video.

\subsubsection{Evaluation data}

In \cref{tab:eval_inputs_animals,tab:eval_inputs_humans,tab:eval_inputs_objects} we provide the list of sketches used for our quantitative evaluations, along with their associated prompt.

\subsubsection{User Study}
\begin{figure}
    \centering
    \includegraphics[width=\linewidth]{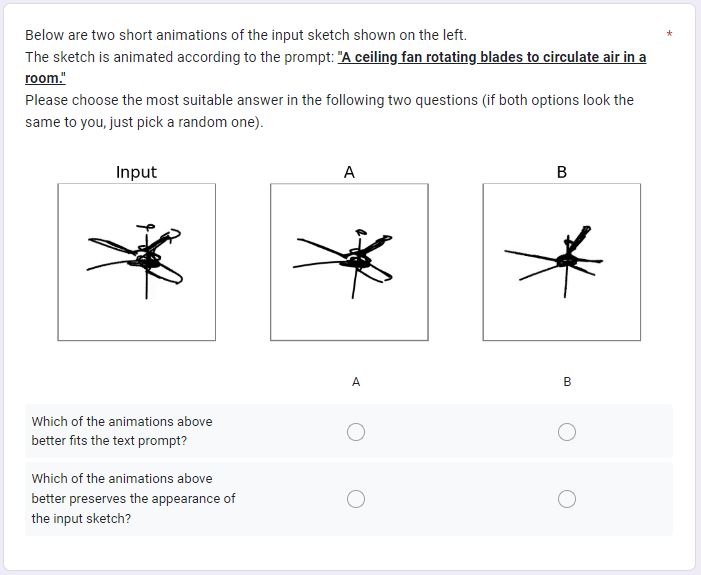}
    \caption{User study example question.}
    \label{fig:ustudy_example}
\end{figure}

As discussed in section 5.2 of the main paper, we conduct a user study to validate our suggested components.
The user study examines the sketch-to-video consistency and text-to-video alignment of the animations produced when disabling different components of our method.
An example question is shown in \cref{fig:ustudy_example}. These questions were repeated for all the targets in the evaluation set, each time comparing our full method to a random choice of the ablation scenarios.

\begin{table*}[hbt]\setlength{\tabcolsep}{3pt}
\vspace{-3pt}
\caption{Sketches, and prompts used for our quantitative evaluations for the "animal" class. }\label{tab:eval_inputs_animals}

\centering 
\begin{tabular}{cl} 

\begin{tabular}{c}\includegraphics[width=0.1\linewidth]{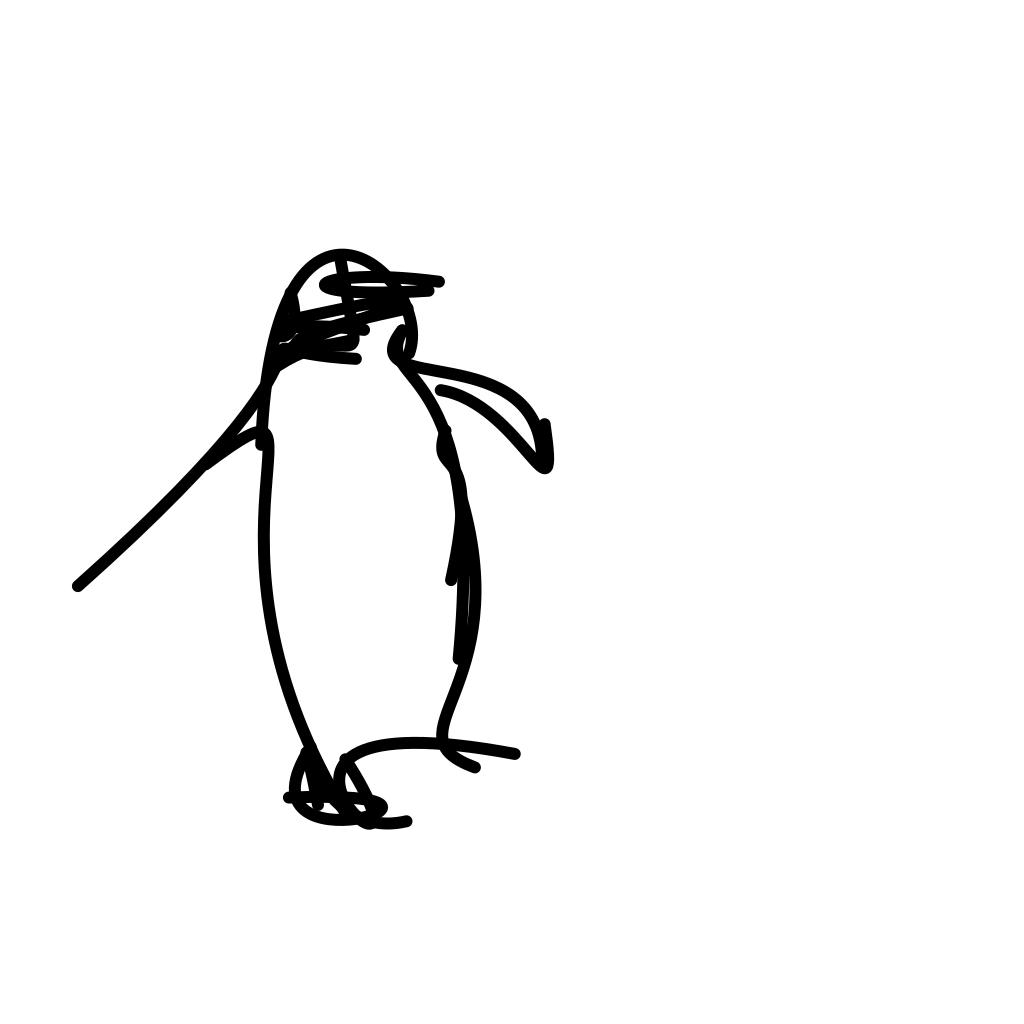}\end{tabular} & \begin{tabular}{l}The penguin is shuffling along the ice terrain, taking deliberate and cautious step with its flippers \\  outstretched to maintain balance.\end{tabular} \\
\begin{tabular}{c}\includegraphics[width=0.1\linewidth]{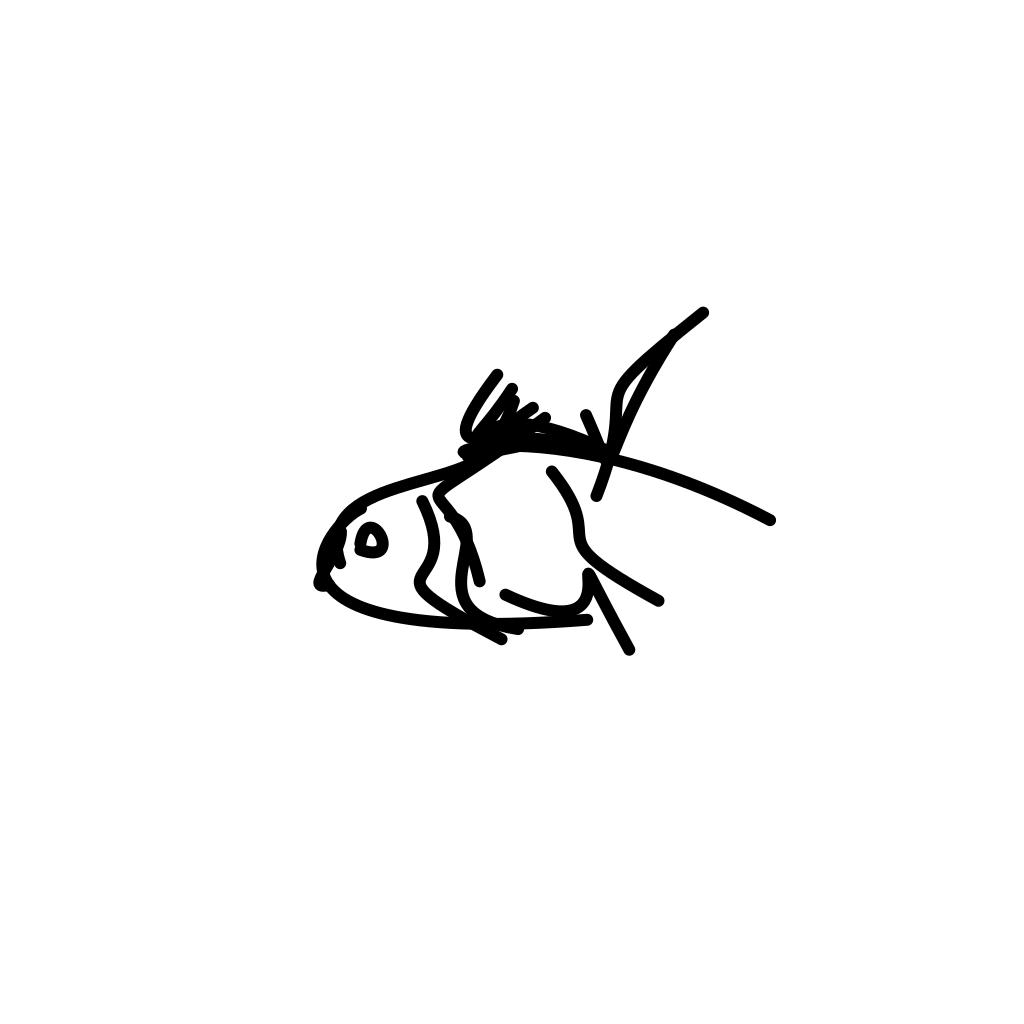}\end{tabular} & \begin{tabular}{l}The goldenfish is gracefully moving through the water, its fins and tail fin gently propelling it \\  forward with effortless agility.\end{tabular} \\
\begin{tabular}{c}\includegraphics[width=0.1\linewidth]{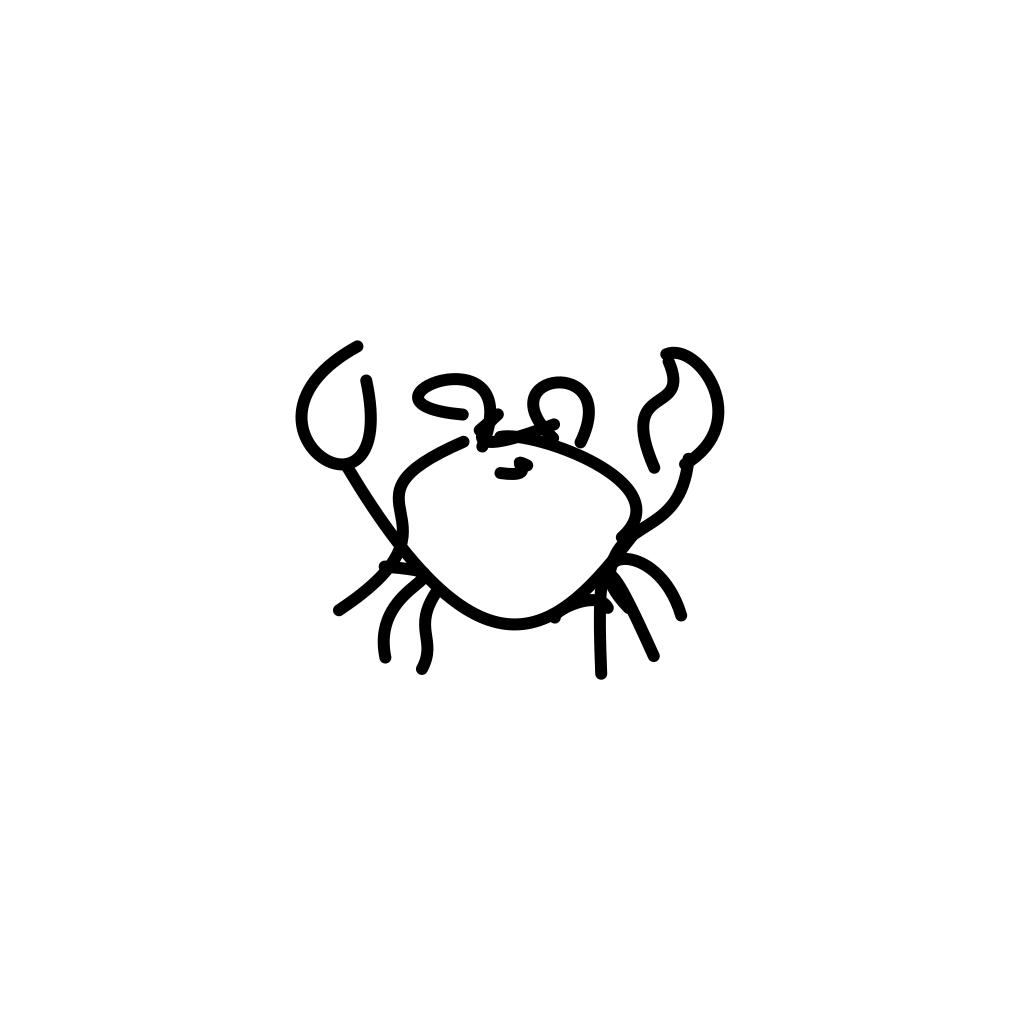}\end{tabular} & \begin{tabular}{l}The crab scuttled sideways along the sandy beach, its pincers raised in a defensive stance.\end{tabular} \\
\begin{tabular}{c}\includegraphics[width=0.1\linewidth]{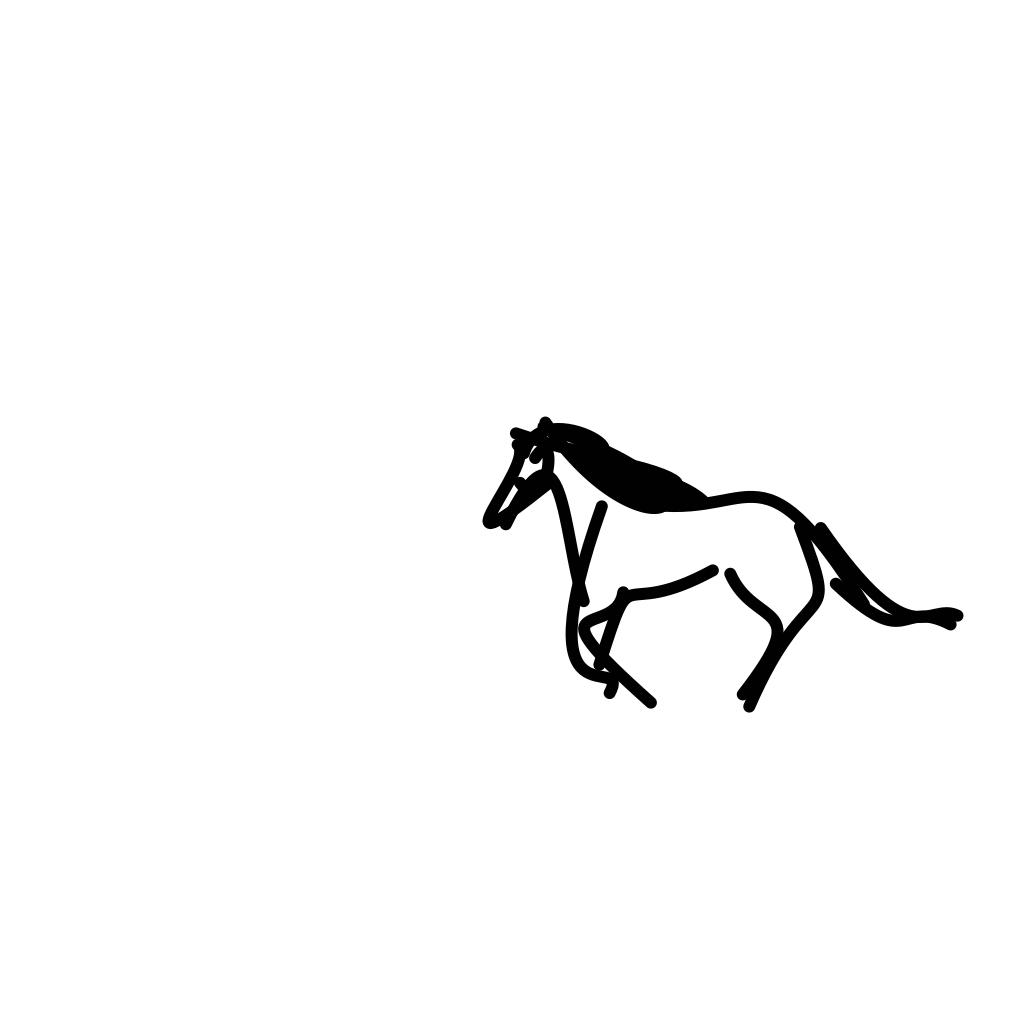}\end{tabular} & \begin{tabular}{l}A galloping horse.\end{tabular} \\
\begin{tabular}{c}\includegraphics[width=0.1\linewidth]{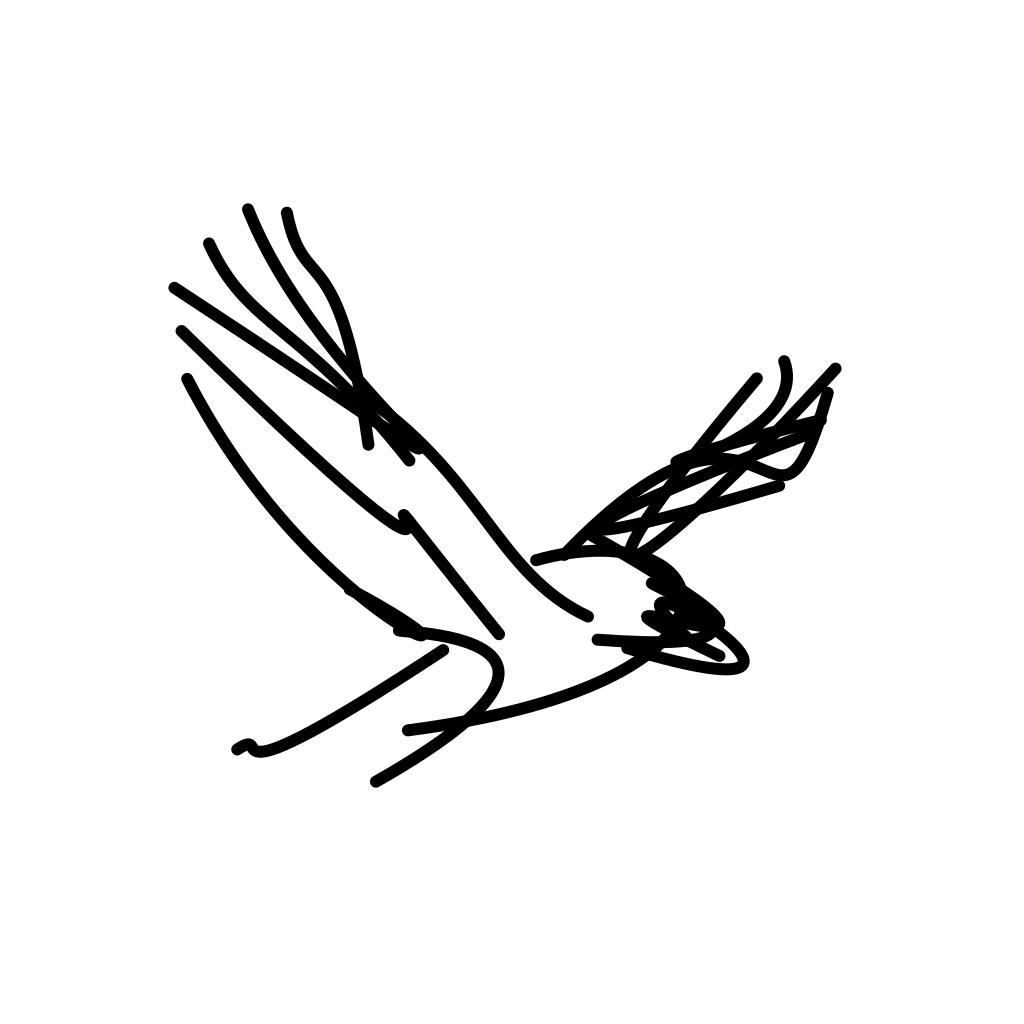}\end{tabular} & \begin{tabular}{l}The eagle soars majestically, with powerful wing beats and effortless glides, displaying precise \\  control and keen vision as it maneuvers gracefully through the sky.\end{tabular} \\
\begin{tabular}{c}\includegraphics[width=0.1\linewidth]{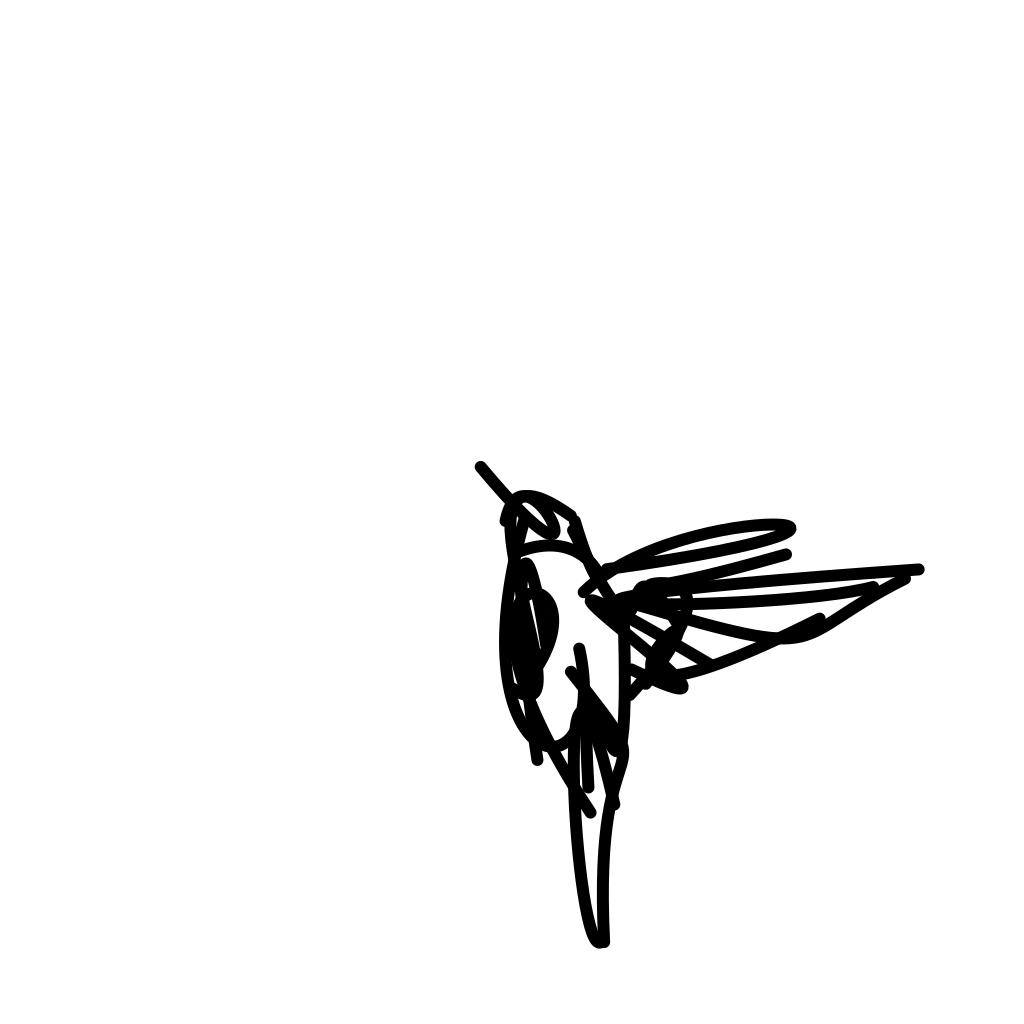}\end{tabular} & \begin{tabular}{l}A hummingbird hovers in mid-air and sucks nectar from a flower.\end{tabular} \\
\begin{tabular}{c}\includegraphics[width=0.1\linewidth]{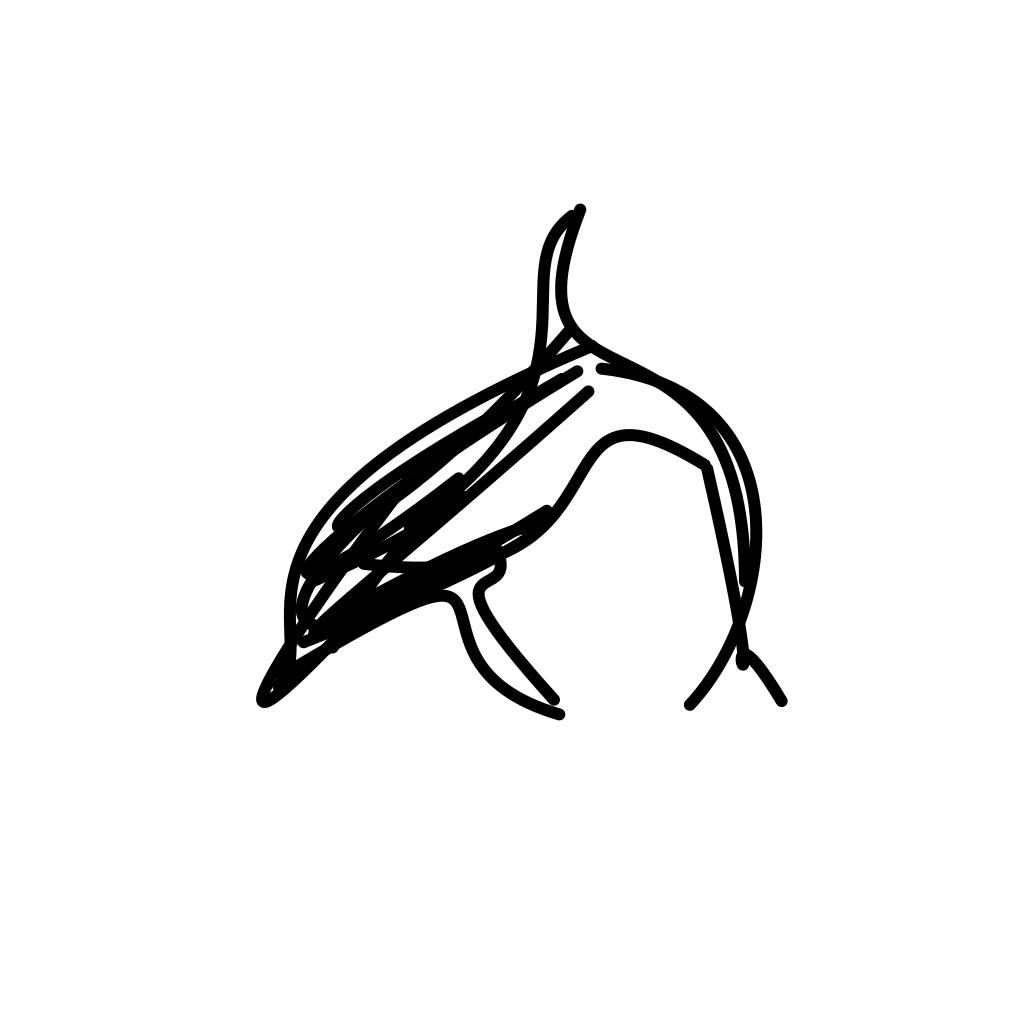}\end{tabular} & \begin{tabular}{l}A dolphin swimming and leaping out of the water.\end{tabular} \\
\begin{tabular}{c}\includegraphics[width=0.1\linewidth]{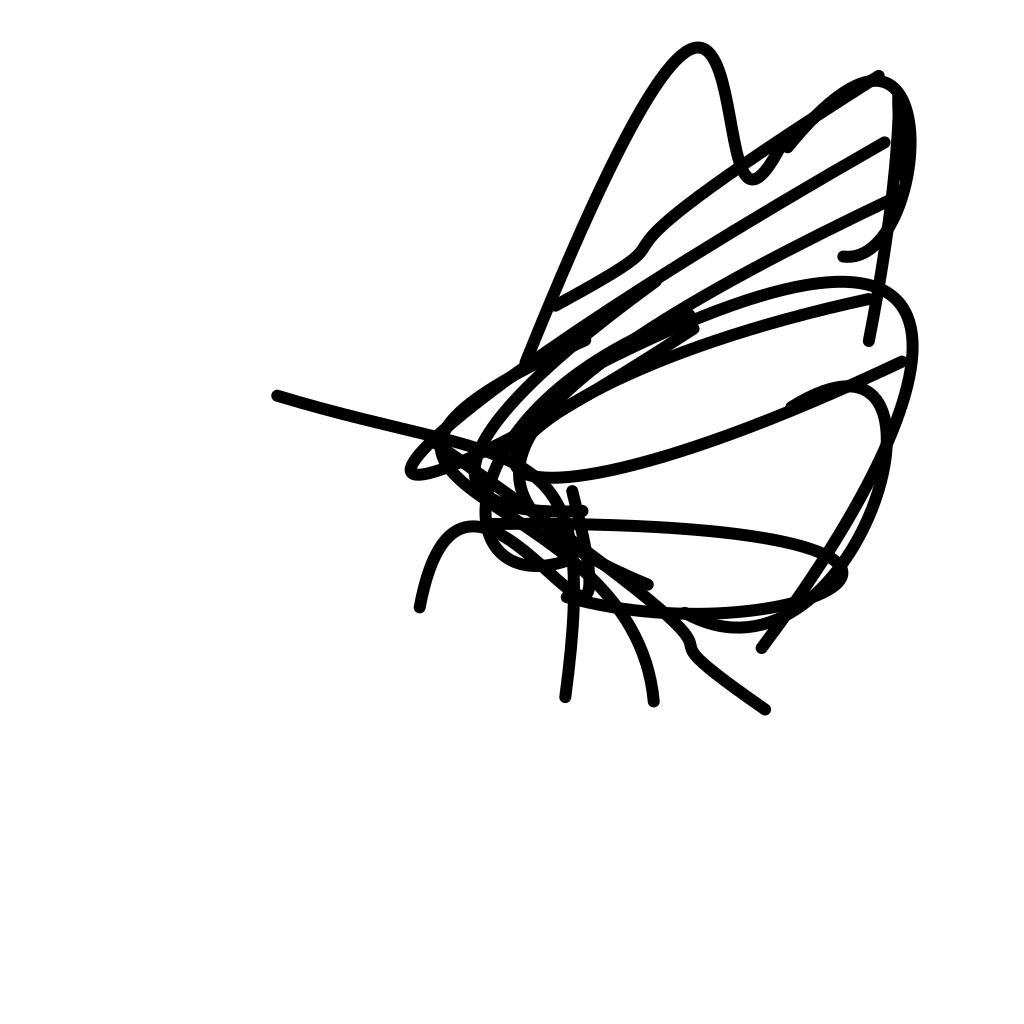}\end{tabular} & \begin{tabular}{l}A butterfly fluttering its wings and flying gracefully.\end{tabular} \\
\begin{tabular}{c}\includegraphics[width=0.1\linewidth]{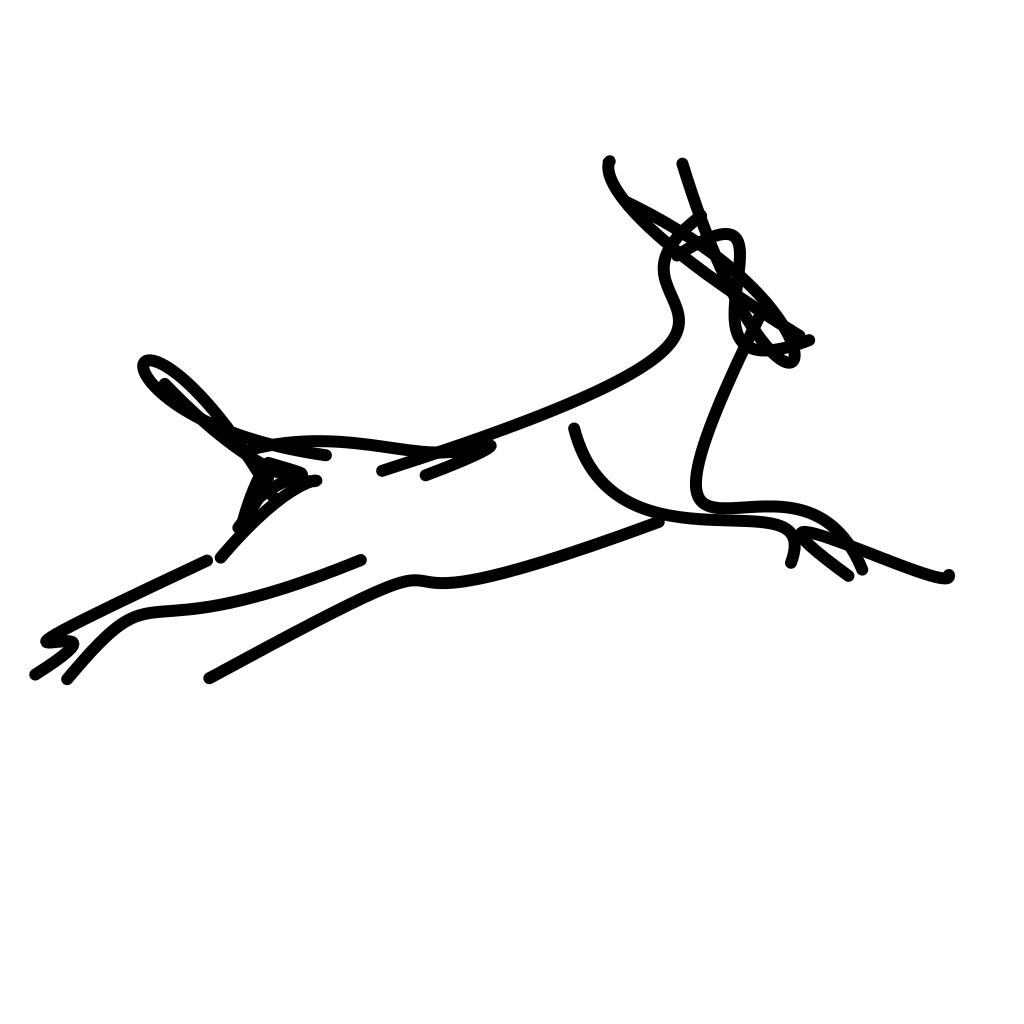}\end{tabular} & \begin{tabular}{l}A gazelle galloping and jumping to escape predators.\end{tabular} \\
\begin{tabular}{c}\includegraphics[width=0.1\linewidth]{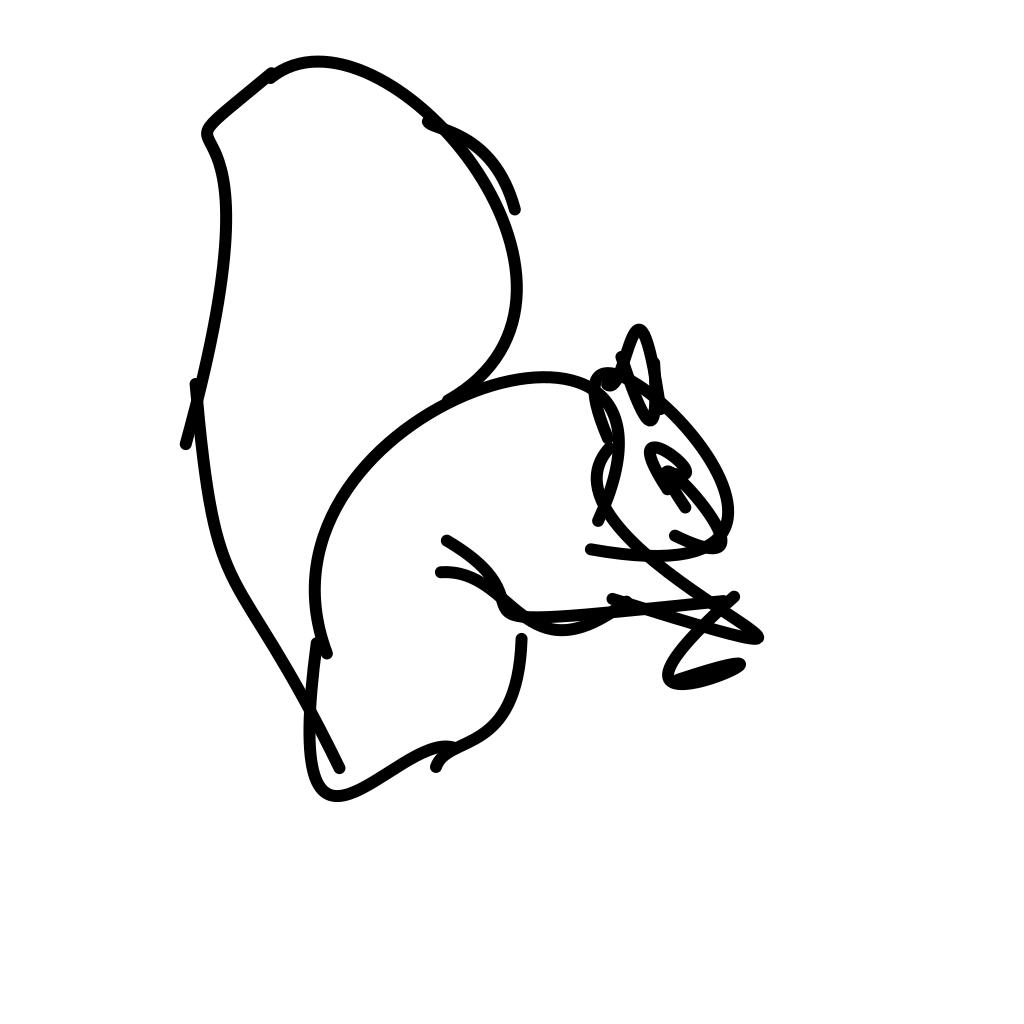}\end{tabular} & \begin{tabular}{l}The squirrel uses its dexterous front paws to hold and manipulate nuts, displaying meticulous and \\  deliberate motions while eating.\end{tabular} \\

\end{tabular}
\vspace{-8pt}
\end{table*}

\begin{table*}[hbt]\setlength{\tabcolsep}{3pt}
\vspace{-3pt}
\caption{Sketches, and prompts used for our quantitative evaluations for the "human" class. }\label{tab:eval_inputs_humans}

\centering 
\begin{tabular}{cl} 

\begin{tabular}{c}\includegraphics[width=0.1\linewidth]{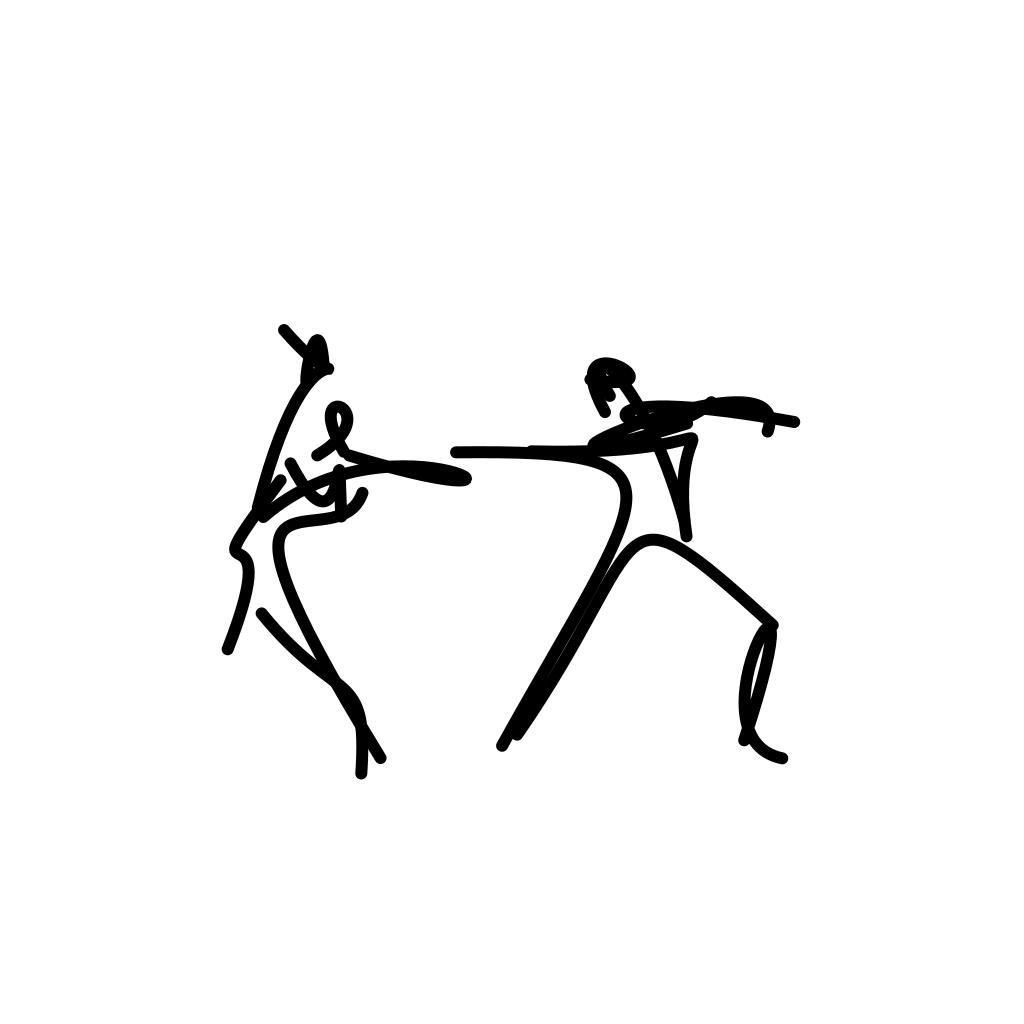}\end{tabular} & \begin{tabular}{l}The two dancers are passionately dancing the Cha-Cha, their bodies moving in sync with the infectious \\  Latin rhythm.\end{tabular} \\
\begin{tabular}{c}\includegraphics[width=0.1\linewidth]{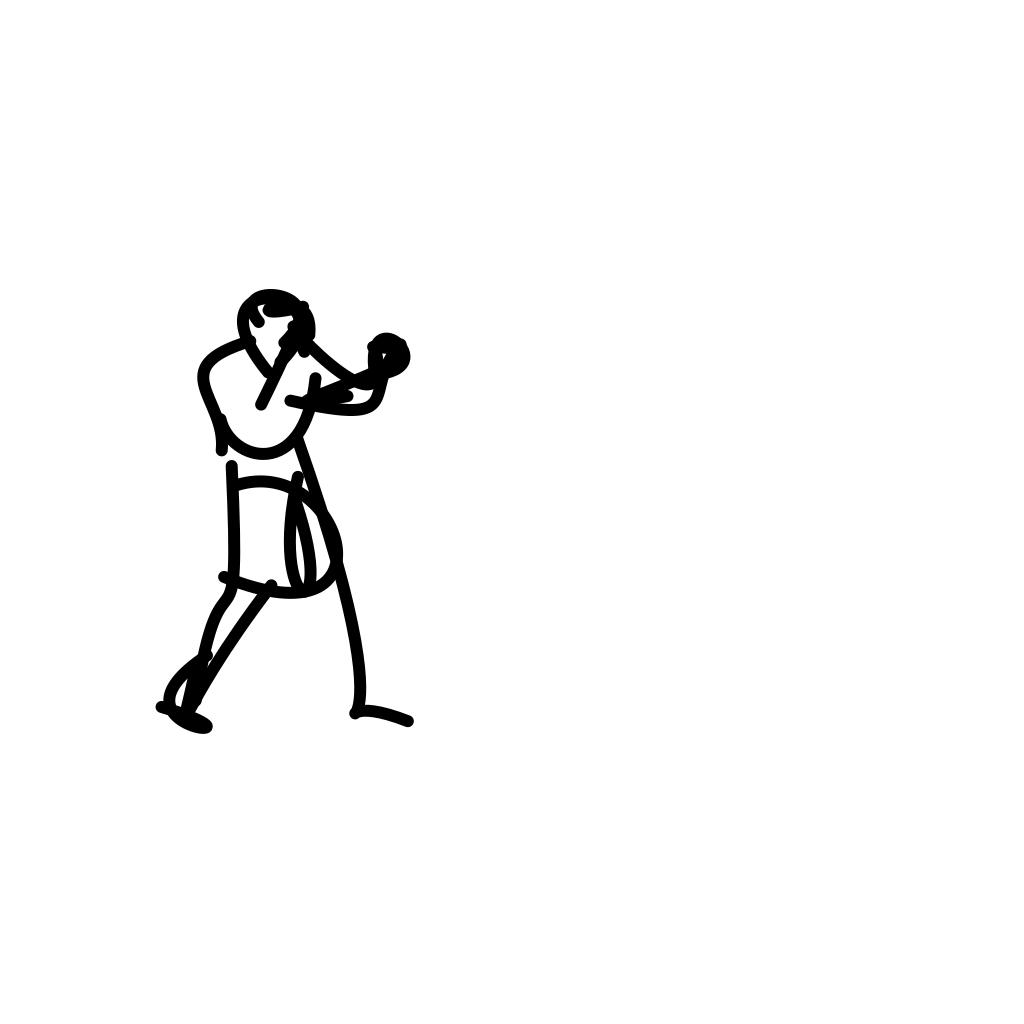}\end{tabular} & \begin{tabular}{l}The boxer ducking and weaving to avoid his opponent's punches, and to punch him back.\end{tabular} \\
\begin{tabular}{c}\includegraphics[width=0.1\linewidth]{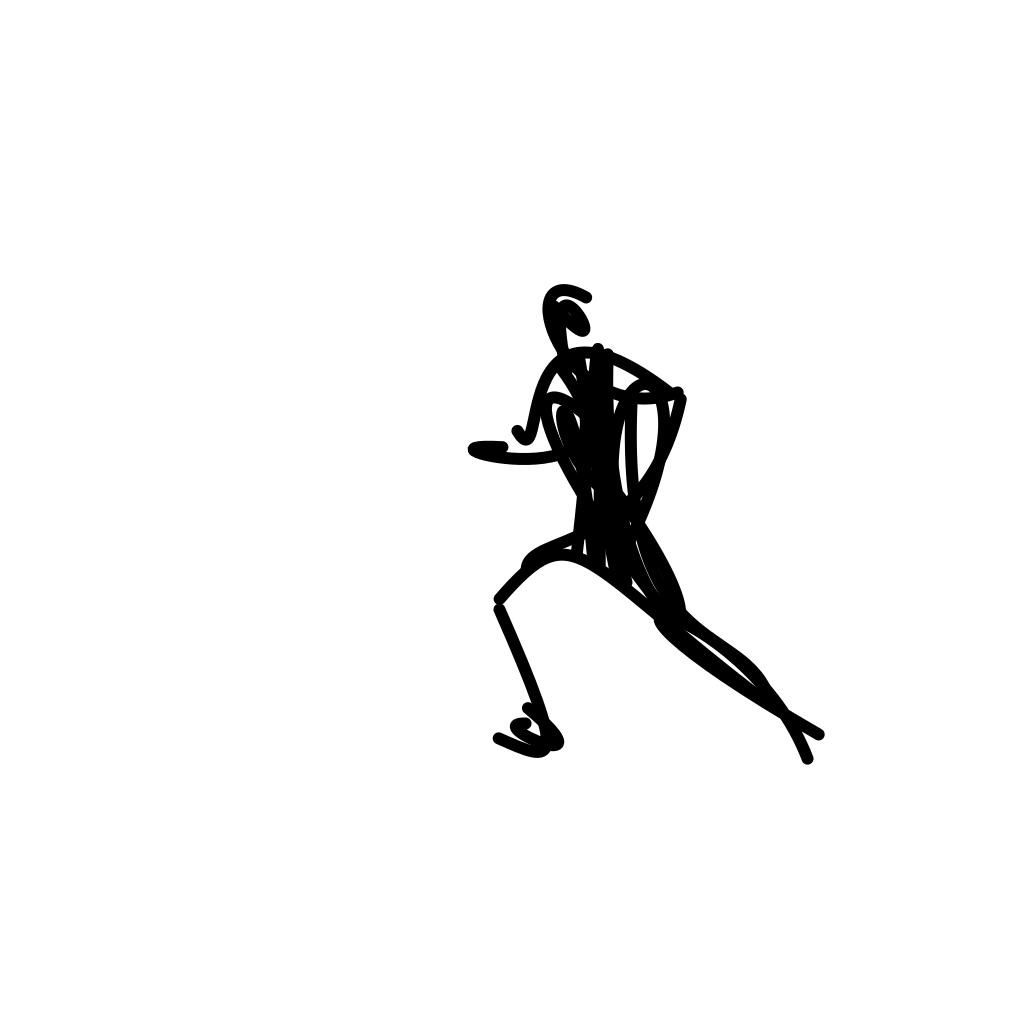}\end{tabular} & \begin{tabular}{l}The runner runs with rhythmic leg strides and synchronized arm swing propelling them forward while \\  maintaining balance.\end{tabular} \\
\begin{tabular}{c}\includegraphics[width=0.1\linewidth]{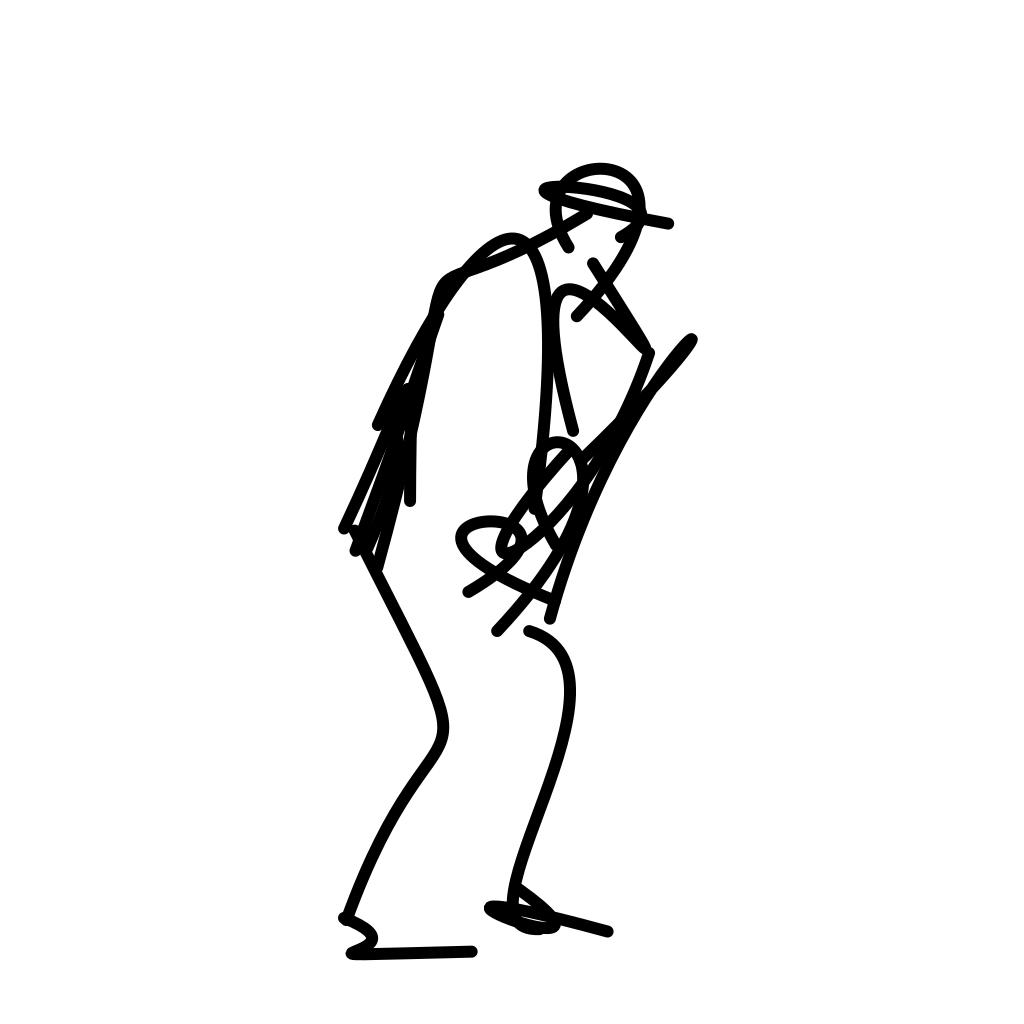}\end{tabular} & \begin{tabular}{l}The jazz saxophonist performs on stage with a rhythmic sway, his upper body sways subtly to the \\  rhythm of the music.\end{tabular} \\
\begin{tabular}{c}\includegraphics[width=0.1\linewidth]{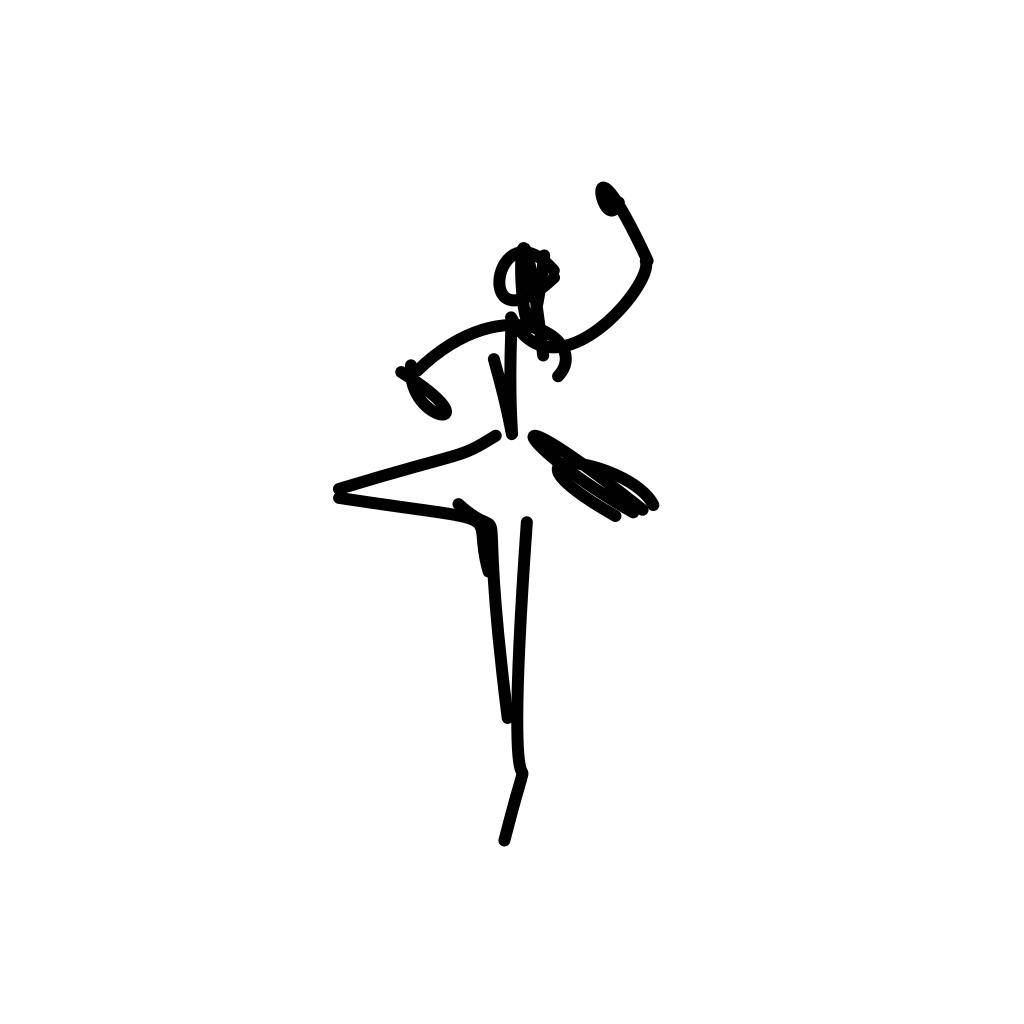}\end{tabular} & \begin{tabular}{l}The ballerina is dancing.\end{tabular} \\
\begin{tabular}{c}\includegraphics[width=0.1\linewidth]{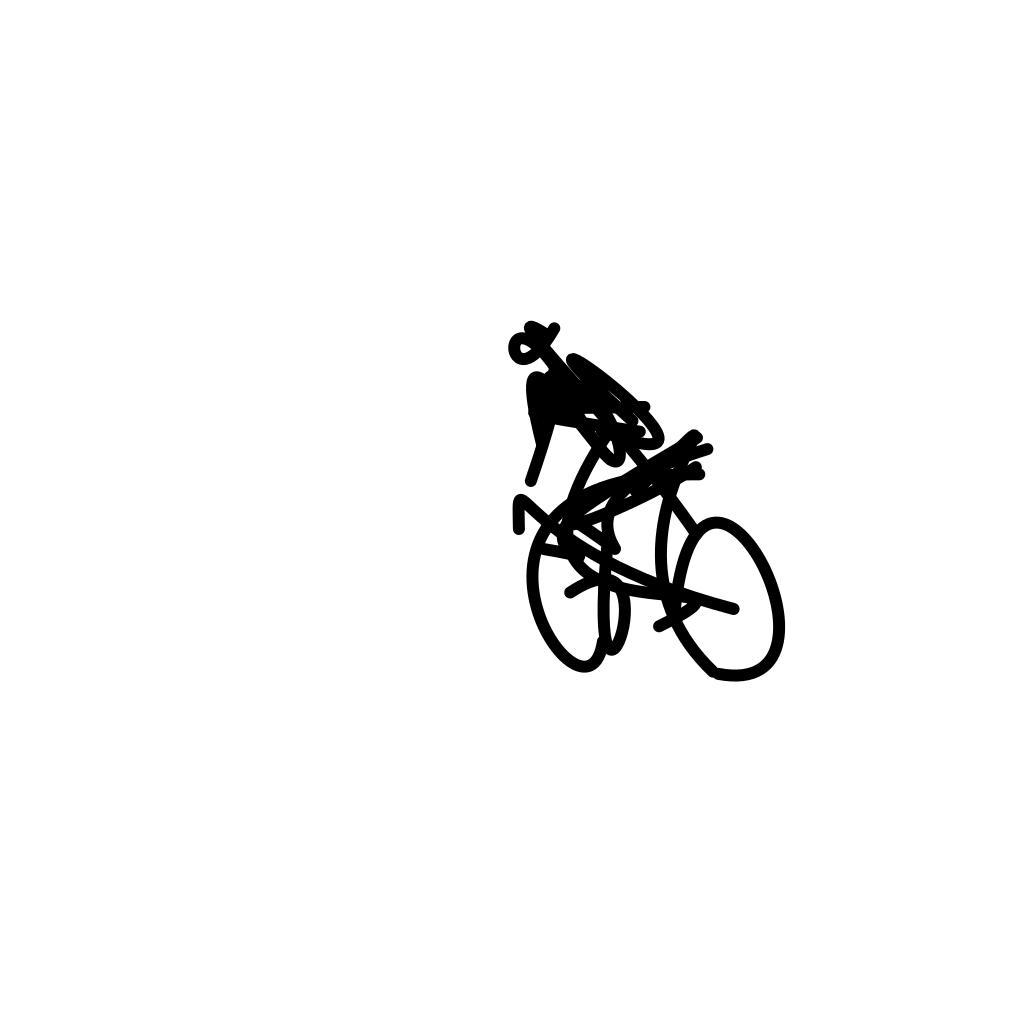}\end{tabular} & \begin{tabular}{l}The biker is pedaling, each leg pumping up and down as the wheels of the bicycle spin rapidly, propelling \\  them forward.\end{tabular} \\
\begin{tabular}{c}\includegraphics[width=0.1\linewidth]{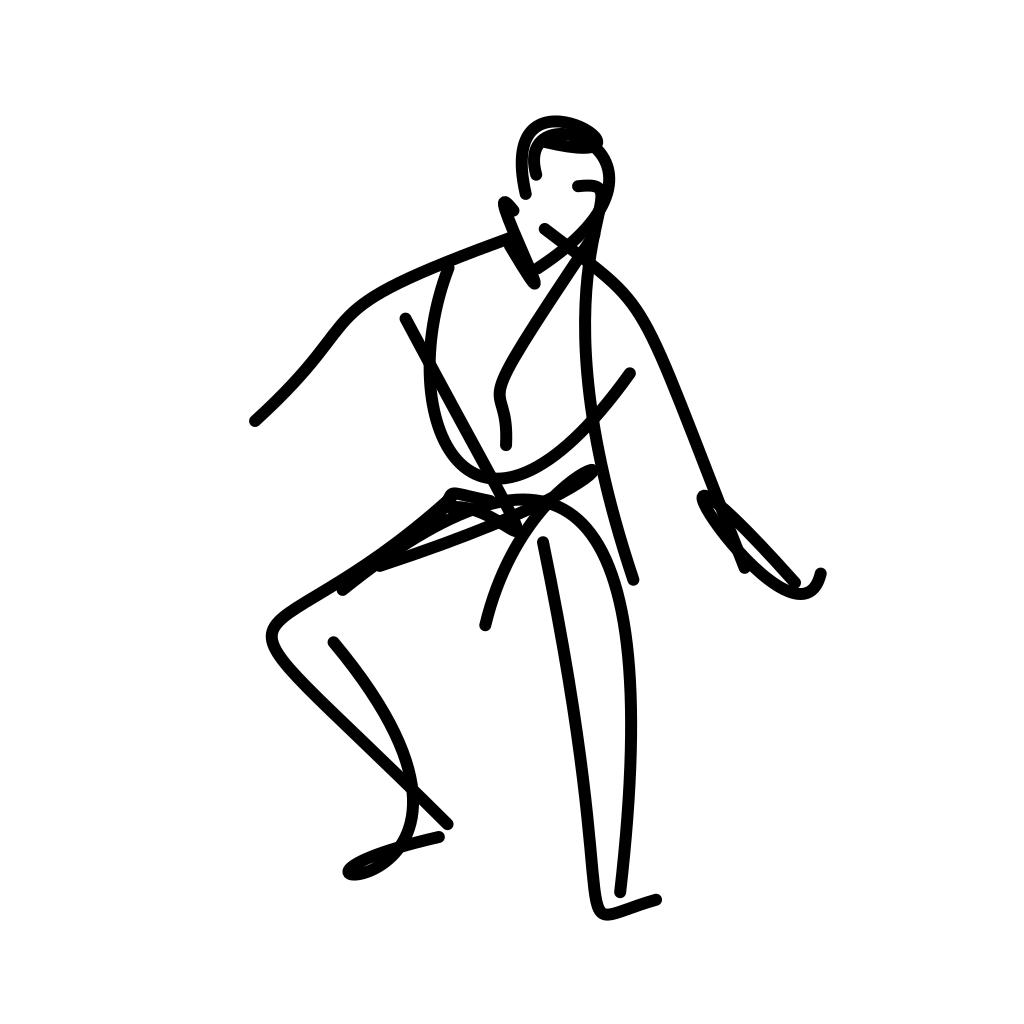}\end{tabular} & \begin{tabular}{l}A martial artist executing precise and controlled movements in different forms of martial arts.\end{tabular} \\
\begin{tabular}{c}\includegraphics[width=0.1\linewidth]{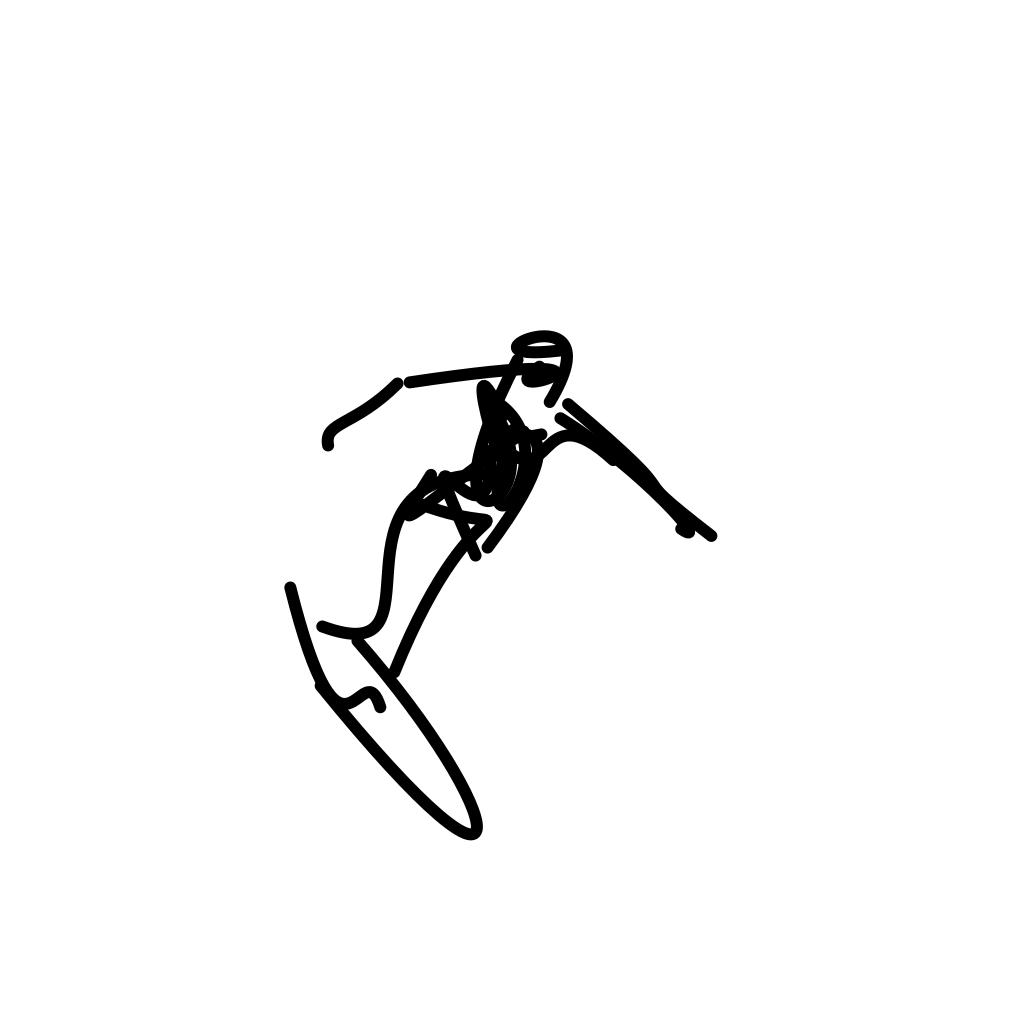}\end{tabular} & \begin{tabular}{l}A surfer riding and maneuvering on waves on a surfboard.\end{tabular} \\
\begin{tabular}{c}\includegraphics[width=0.1\linewidth]{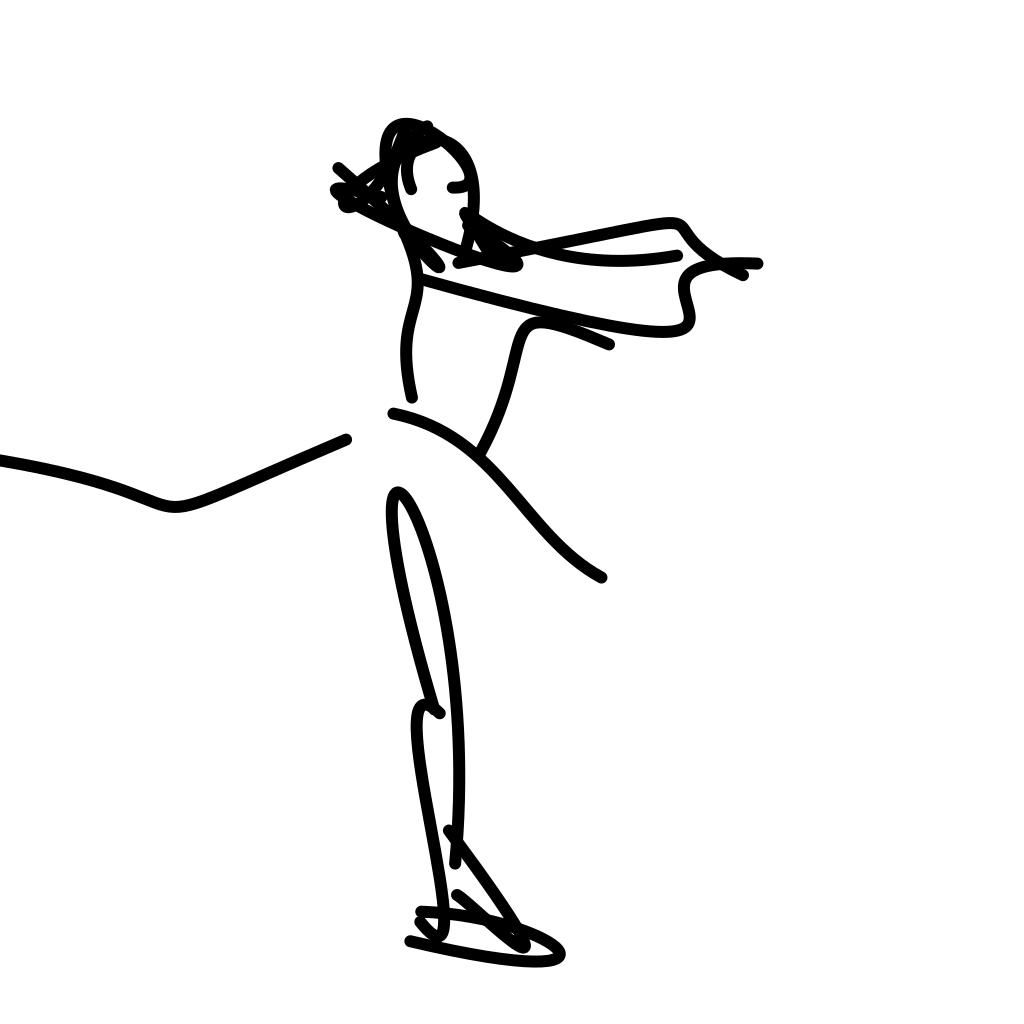}\end{tabular} & \begin{tabular}{l}A figure skater gliding, spinning, and performing jumps on ice skates.\end{tabular} \\
\begin{tabular}{c}\includegraphics[width=0.1\linewidth]{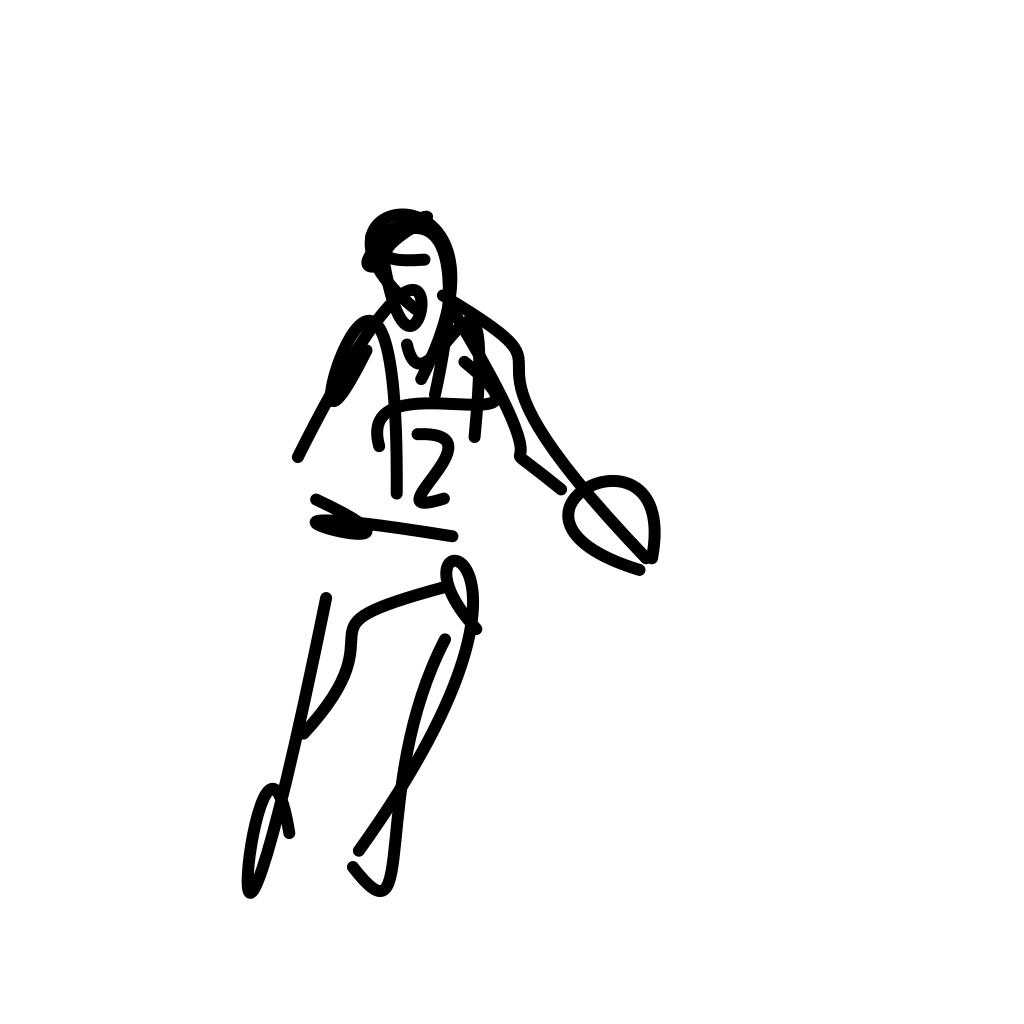}\end{tabular} & \begin{tabular}{l}A basketball player dribbling and passing while playing basketball.\end{tabular} \\

\end{tabular}
\vspace{-8pt}
\end{table*}

\begin{table*}[hbt]\setlength{\tabcolsep}{3pt}
\vspace{-3pt}
\caption{Sketches, and prompts used for our quantitative evaluations for the "object" class. }\label{tab:eval_inputs_objects}

\centering 
\begin{tabular}{cl} 

\begin{tabular}{c}\includegraphics[width=0.1\linewidth]{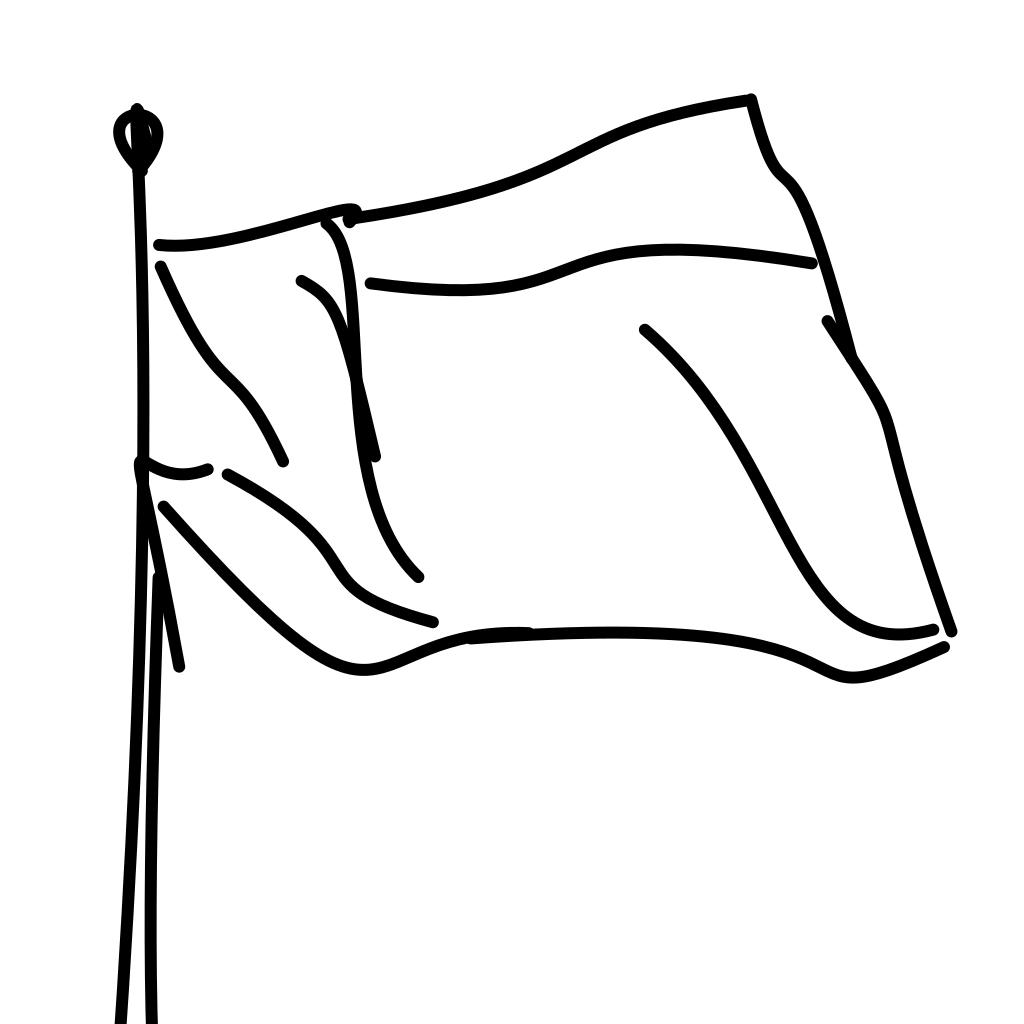}\end{tabular} & \begin{tabular}{l}A waving flag fluttering and rippling in the wind.\end{tabular} \\
\begin{tabular}{c}\includegraphics[width=0.1\linewidth]{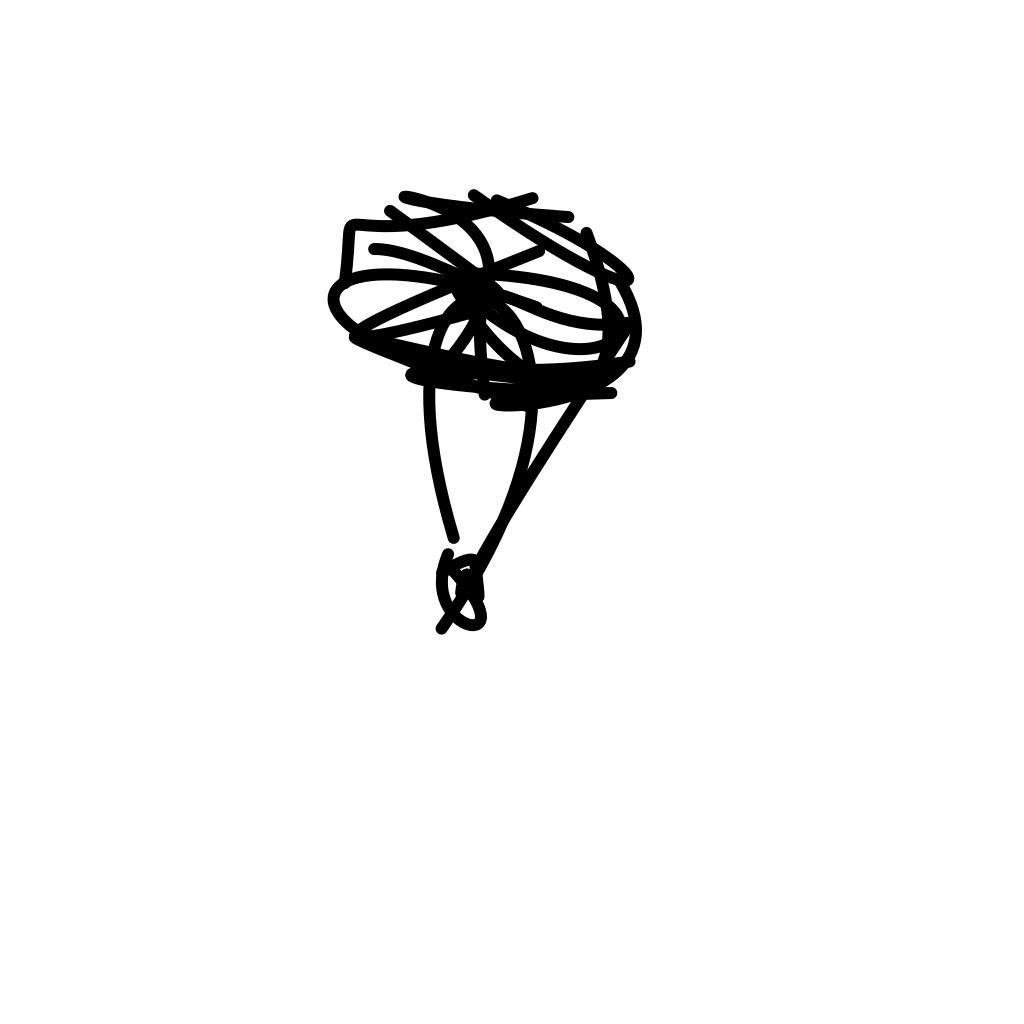}\end{tabular} & \begin{tabular}{l}A parachute descending slowly and gracefully after being deployed.\end{tabular} \\
\begin{tabular}{c}\includegraphics[width=0.1\linewidth]{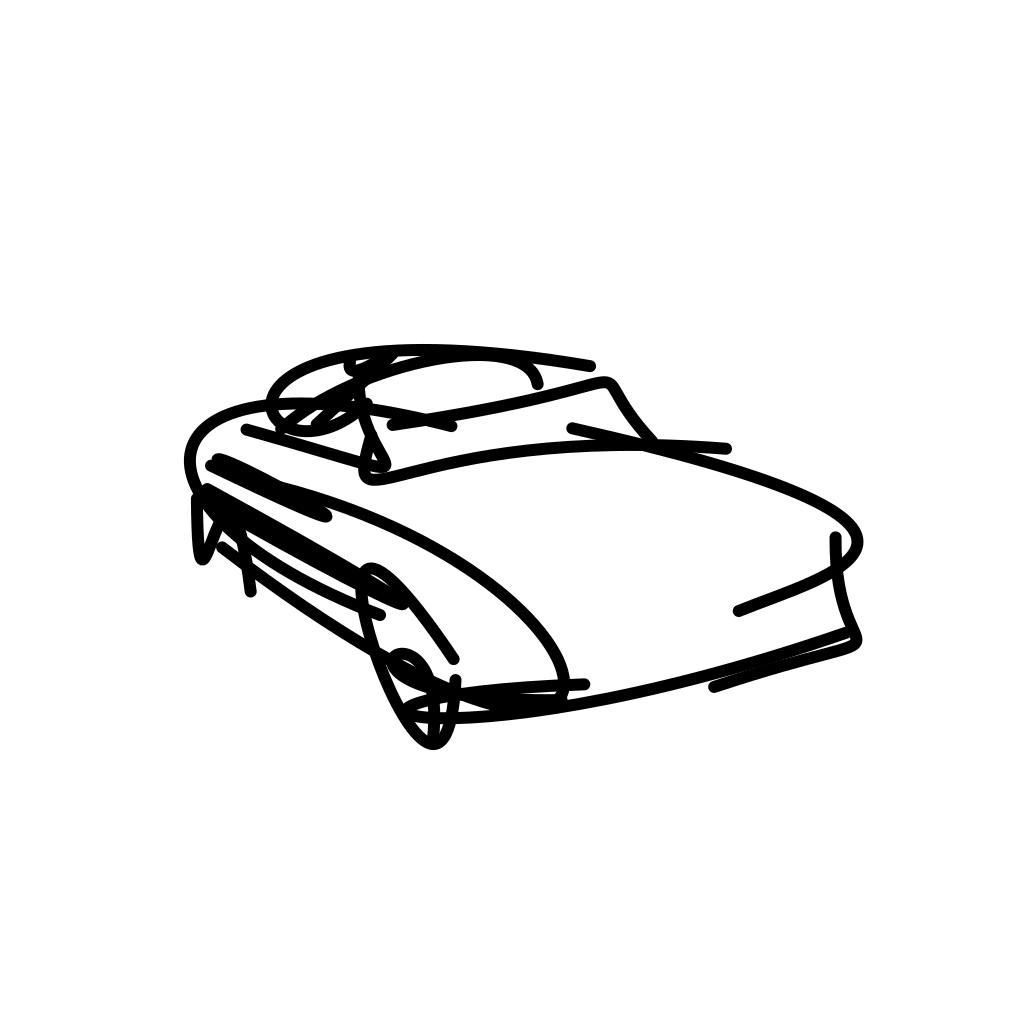}\end{tabular} & \begin{tabular}{l}A wind-up toy car, moving forward or backward when wound up and released.\end{tabular} \\
\begin{tabular}{c}\includegraphics[width=0.1\linewidth]{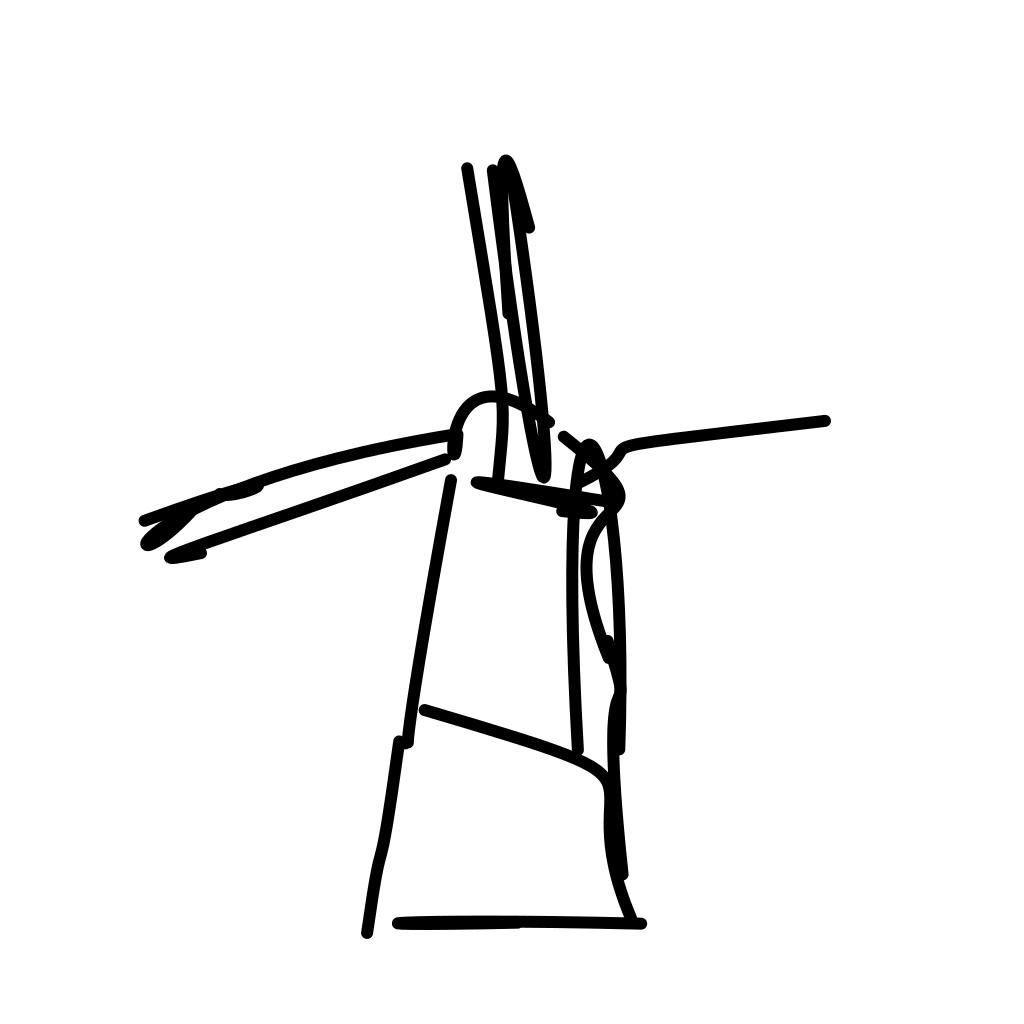}\end{tabular} & \begin{tabular}{l}A windmill spinning its blades in the wind to generate energy.\end{tabular} \\
\begin{tabular}{c}\includegraphics[width=0.1\linewidth]{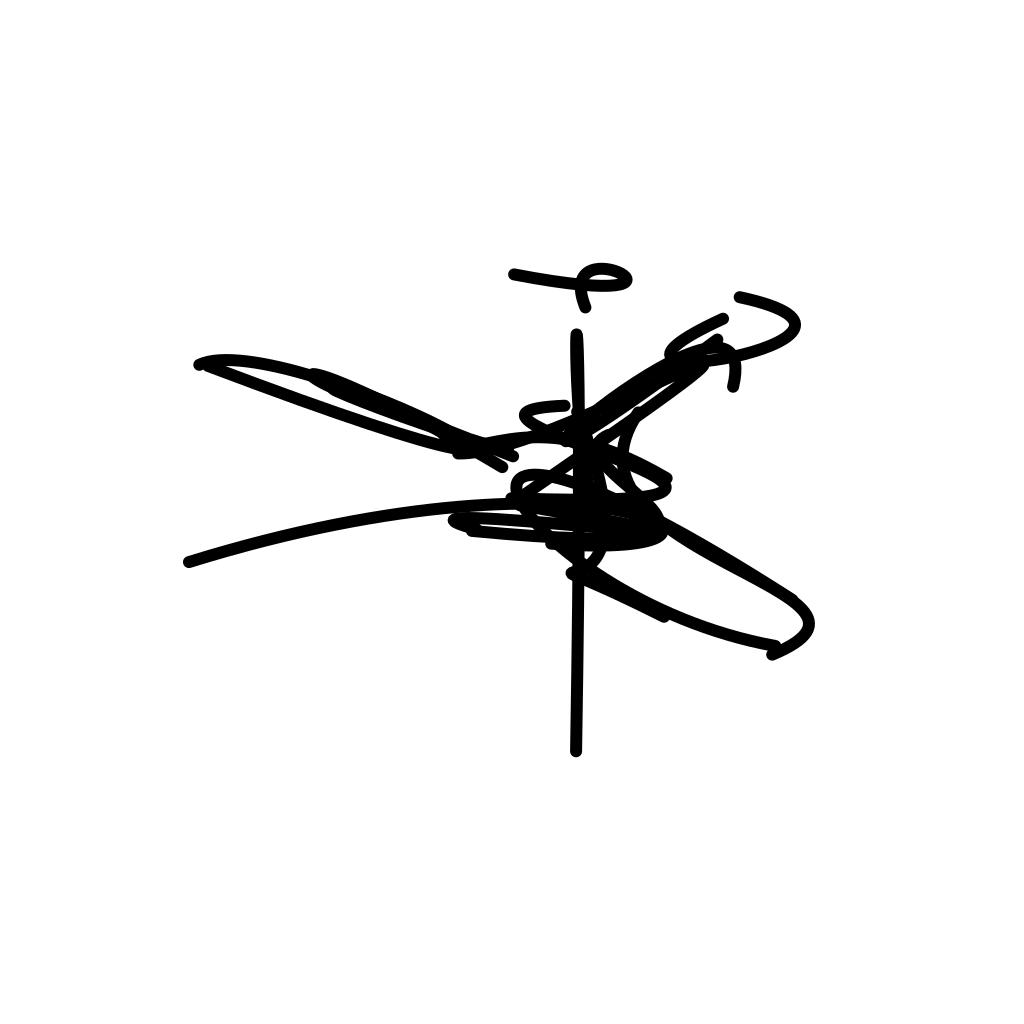}\end{tabular} & \begin{tabular}{l}A ceiling fan rotating blades to circulate air in a room.\end{tabular} \\
\begin{tabular}{c}\includegraphics[width=0.1\linewidth]{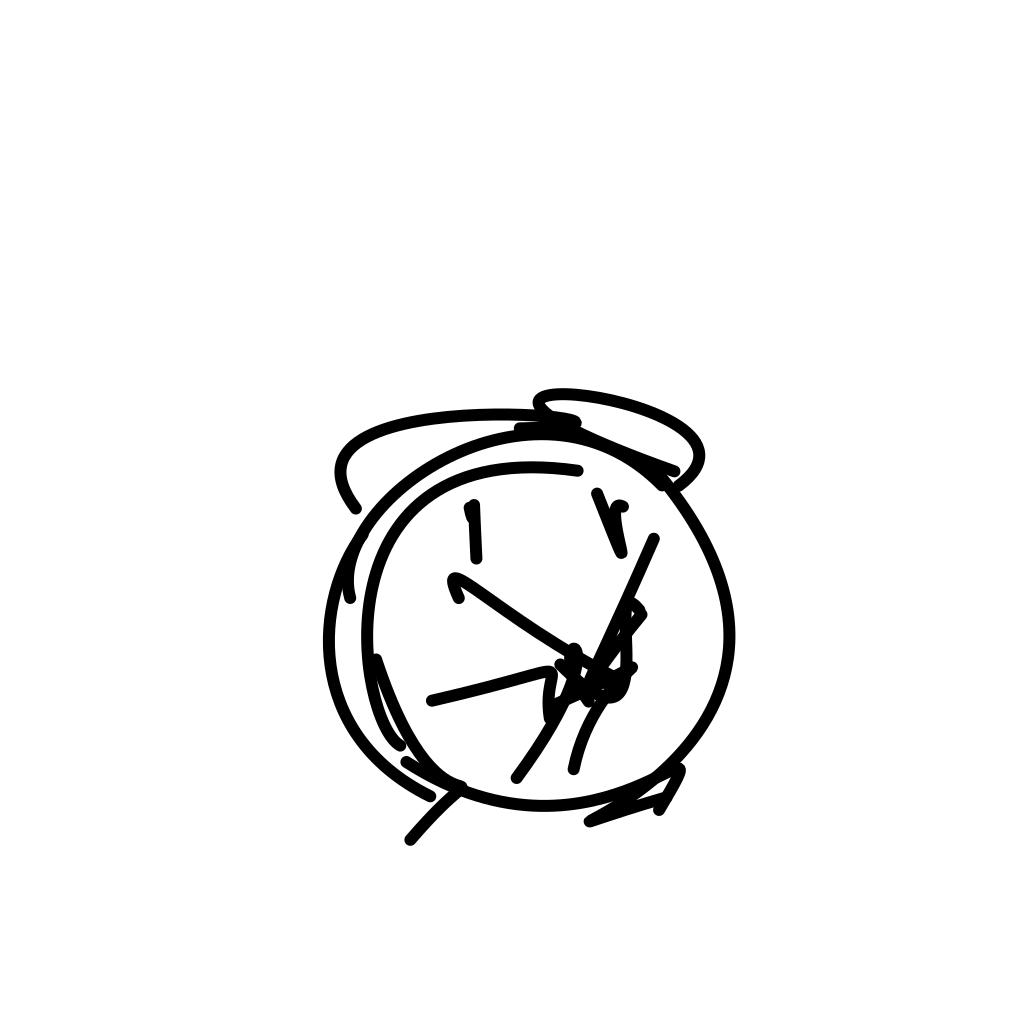}\end{tabular} & \begin{tabular}{l}A clock hands ticking and rotating to indicate time on a clock face.\end{tabular} \\
\begin{tabular}{c}\includegraphics[width=0.1\linewidth]{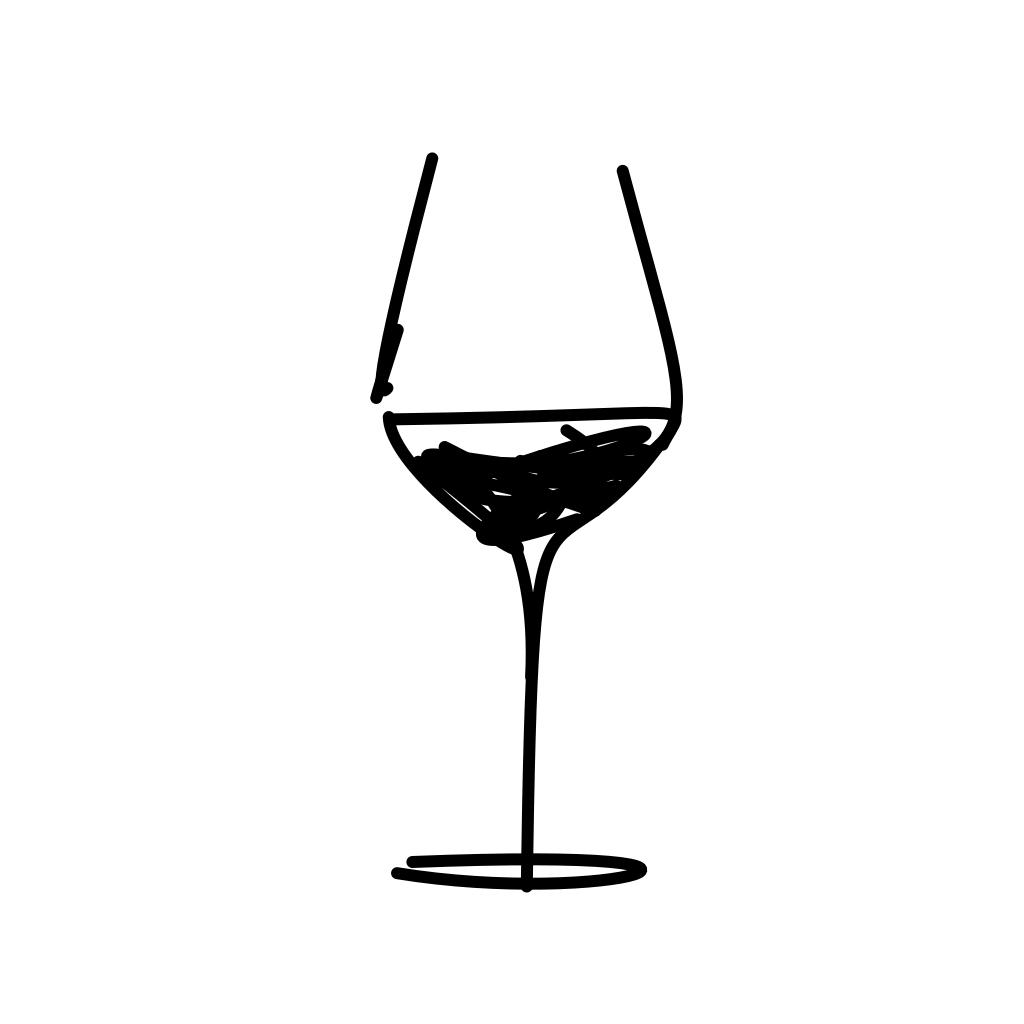}\end{tabular} & \begin{tabular}{l}The wine in the wine glass sways from side to side.\end{tabular} \\
\begin{tabular}{c}\includegraphics[width=0.1\linewidth]{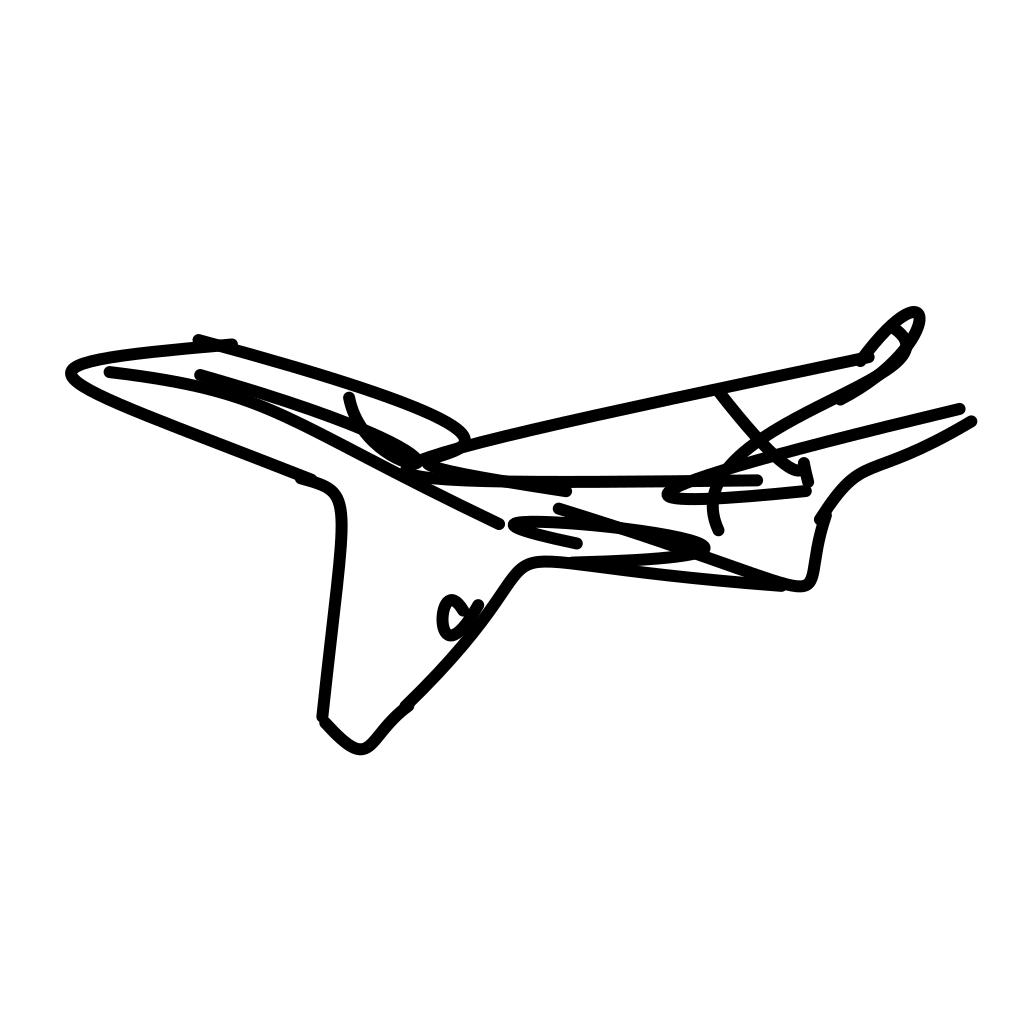}\end{tabular} & \begin{tabular}{l}The airplane moves swiftly and steadily through the air.\end{tabular} \\
\begin{tabular}{c}\includegraphics[width=0.1\linewidth]{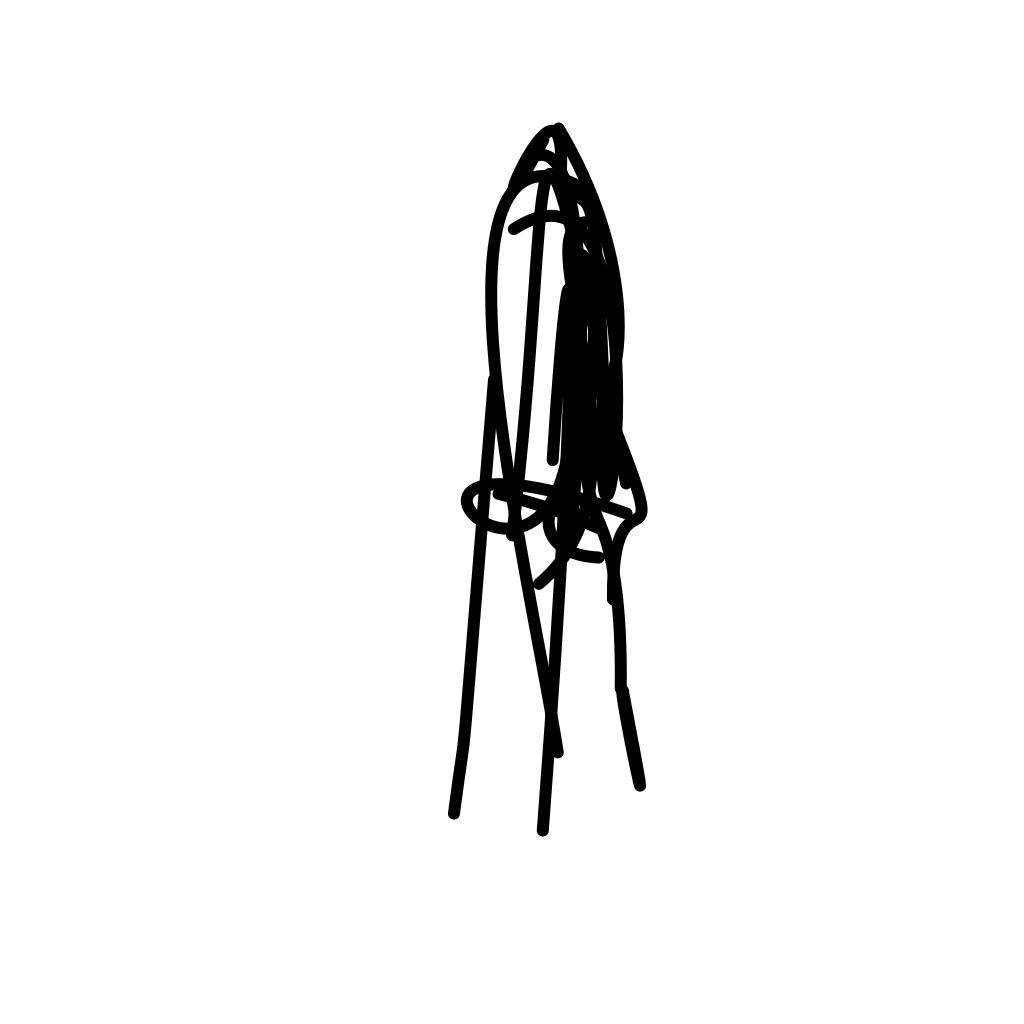}\end{tabular} & \begin{tabular}{l}The spaceship accelerates rapidly during takeoff, utilizing powerful rocket engines.\end{tabular} \\
\begin{tabular}{c}\includegraphics[width=0.1\linewidth]{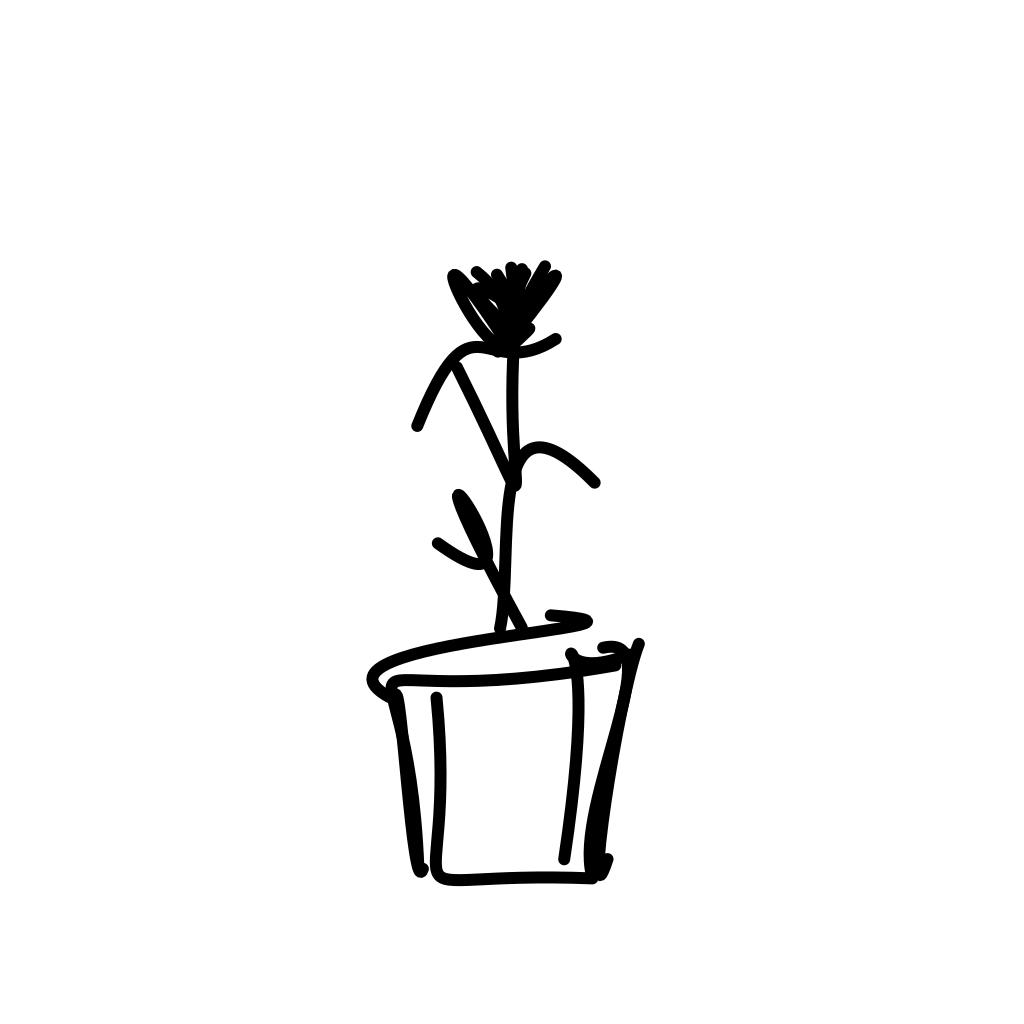}\end{tabular} & \begin{tabular}{l}The flower is moving and growing, swaying gently from side to side.\end{tabular} \\

\end{tabular}
\vspace{-8pt}
\end{table*}

\end{document}